\documentclass{article} %
\usepackage{iclr2027_conference,times}

\usepackage{amsmath,amsfonts,bm}

\def\eqref#1{equation~\ref{#1}}

\def\1{\bm{1}}

\DeclareMathAlphabet{\mathsfit}{\encodingdefault}{\sfdefault}{m}{sl}
\SetMathAlphabet{\mathsfit}{bold}{\encodingdefault}{\sfdefault}{bx}{n}

\newcommand{\ourtax}{\textsc{Habit}}
\newcommand{\ourbench}{\textsc{AgentHabit}}

\usepackage{booktabs}
\usepackage{array}
\usepackage{tabularx}
\usepackage{caption}  
\usepackage{amsmath}
\usepackage[breakable]{tcolorbox}

\newlength{\ptcolA}
\newlength{\ptcolB}
\newlength{\ptcolAB}

\newcommand{\ptboth}[1]{%
  \multicolumn{2}{>{\raggedright\arraybackslash}p{\ptcolAB}@{}}{#1}}

\newcommand{\ptgroup}[1]{%
  \addlinespace[2pt]%
  \multicolumn{3}{@{}l@{}}{\textbf{#1}}\\
  \addlinespace[3pt]}

\newcommand{\na}{\textrm{--}}
\newcommand{\sci}[2]{\ensuremath{#1\times 10^{#2}}}%
\usepackage[utf8]{inputenc} %
\usepackage[T1]{fontenc}    %
\usepackage{hyperref}       %
\usepackage{url}            %
\usepackage{booktabs}       %
\usepackage{amsfonts}       %
\usepackage{amsmath}        %
\usepackage{nicefrac}       %
\usepackage{microtype}      %
\usepackage{xcolor}         %
\usepackage{graphicx}       %
\usepackage{subcaption}     %
\usepackage{wrapfig}
\usepackage{longtable}      %
\usepackage{array}
\usepackage{lineno}
\usepackage{placeins}       %
\usepackage{newunicodechar}
\newunicodechar{’}{\textquoteright}

\title{\ourbench{}: Characterizing Distinct \\Behaviors of Agents on Everyday Tasks}

\author{%
  Woojung Song\thanks{Equal contribution; \textsuperscript{$\dag$}Corresponding author.} \quad
  Hoyeol Yang$^{*}$ \quad
  Jeonghoon Shim \quad
  Sungjib Lim \\
  \textbf{Jonggeun Lee} \quad
  \textbf{Yunho Choi} \quad
  \textbf{Yohan Jo}$^{\dag}$
  \\
  Seoul National University, Seoul, Republic of Korea\\
  \texttt{\{opusdeisong, hoyeol, yohan.jo\}@snu.ac.kr}
}

\iclrfinalcopy %
\begin{document}

\maketitle
\lhead{Preprint}

\begin{abstract}
Large language model (LLM) agents assist users with everyday tasks that can be completed in many reasonable ways. Even when their answers are useful, how agents carry out these tasks may not match users' preferences and needs. For example, agents differ in whether they ask clarifying questions or search the web. We introduce \ourtax{}, a taxonomy of 23 behavioral axes in five categories, which three authors and three LLMs derive bottom-up from 408 agent trajectories across 17 domains. On held-out tasks, \ourtax{} distinguishes models more clearly than existing taxonomies of human values and agent actions while supporting comparably consistent annotation. Building on \ourtax{}, we construct \ourbench{}, a benchmark that profiles each agent's behavioral tendencies from its trajectories on 86 everyday tasks. Profiling 18 models with \ourbench{} reveals a range of distinctive tendencies. For example, most GPT and Claude models state their assumptions and offer alternatives when requirements conflict, whereas Qwen and Google's models more often leave assumptions or changes to requirements unstated. These profiles remain recognizable even when built from entirely different sets of tasks, indicating that they reflect general tendencies rather than task-specific behavior. Prompting agents to adopt specific behaviors shifts some axes readily but barely changes others, while fine-tuning on another model's trajectories changes only part of a model's profile and leaves much of it intact. Overall, \ourtax{} and \ourbench{} provide a systematic framework for characterizing how agents carry out everyday tasks beyond task success, offering insights to guide the development of agents whose behavior better fits users' needs.
\end{abstract}

\section{Introduction}
\label{sec:intro}

\begin{figure}[t]
\centering
  \includegraphics[width=\textwidth]{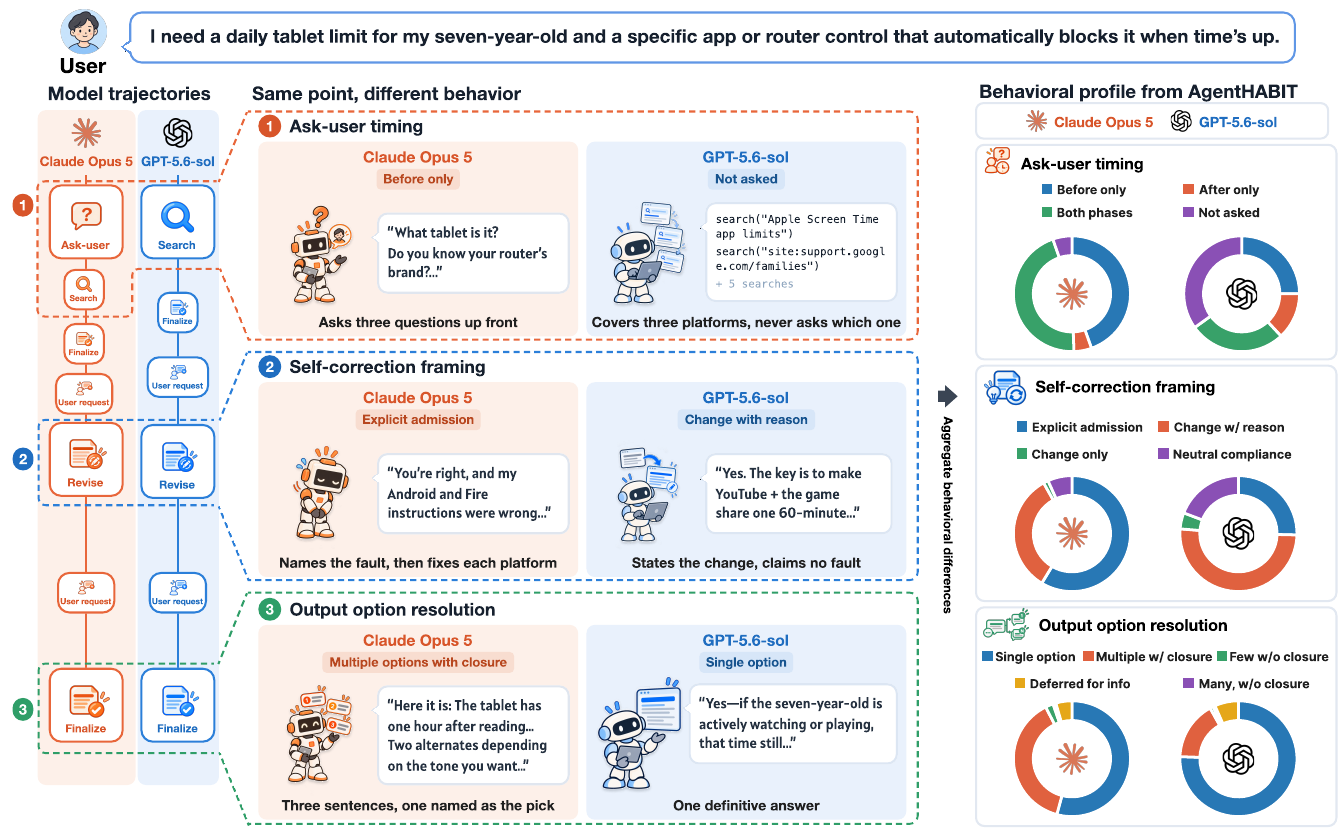}
\caption{\textbf{From everyday interactions to behavioral profiles.}
How two agents respond to the same opening request (left) and their behavioral profiles (right).}
\label{fig:example}
\end{figure}

AI agents assist users with everyday tasks such as drafting emails, learning new concepts, and seeking advice~\citep{Chatterji2025HowPU,handa2025economic}. Agents can successfully complete these tasks in a variety of ways~\citep{zhao2024wildchat,jiang2026artificial}. For example, when asked how to automatically limit a child's tablet use, Claude asks about the device, while GPT gives instructions for several devices without asking (Figure~\ref{fig:example}). Parents seeking personalized advice may welcome these questions, while those wanting a quick answer may find them an unnecessary delay~\citep{amershi2019guidelines}. Even when the final answer is useful, users may already have spent time and money on unnecessary search and analysis. Without understanding these tendencies, users risk relying on agents whose behavior conflicts with their preferences and task requirements. They may also struggle to anticipate how agents will proceed and what form of output to expect. Profiling these tendencies can make agent behavior more transparent, helping users assess where agent behavior falls short of user needs or goes beyond task requirements and choose models accordingly.

Comparing these behavioral tendencies across models requires a taxonomy that captures differences in agents' choices when multiple reasonable approaches are available. We introduce \ourtax{}, a taxonomy grounded in 408 trajectories from eight models performing open-ended tasks across 17 domains derived from the Anthropic Economic Index~\citep{handa2025economic}. The taxonomy includes 23 behavioral axes, such as \emph{Ask-user timing}, \emph{Self-correction framing}, and \emph{Output option resolution} (Figure~\ref{fig:example}). For example, \emph{Ask-user timing} distinguishes whether an agent asks the user questions only before giving its initial answer (\emph{Before only}), only afterward (\emph{After only}), both before and afterward (\emph{Both phases}), or never (\emph{Not asked}). Each axis has a definition, categorical labels, and a \emph{decide-by} rule specifying which evidence to use when assigning a label. To test whether \ourtax{} generalizes beyond the tasks used to construct it, we apply it to 408 trajectories from 17 held-out tasks. We compare it with Schwartz's human values used in psychometric profiling~\citep{han2025valueportrait,dong2026agentvaluebench} and agent action categories~\citep{gao2026actonomy}. \ourtax{} more clearly distinguishes models' behavioral tendencies and yields more varied label distributions, while supporting consistent annotation (Section~\ref{sec:axis-evaluation}).

Building on \ourtax{}, we introduce \ourbench{}, a benchmark profiling agent behavior on everyday tasks. We aggregate behavior observed across tasks into a profile of each model's characteristic tendencies on the 23 behavioral axes (Figure~\ref{fig:example}). The resulting profiles show that most GPT and Claude models state their assumptions and offer alternatives when requirements conflict, while Qwen and Google's models more often leave assumptions or changes to requirements unstated (Figure~\ref{fig:basic_profile}). Such omissions can make it difficult for users to judge whether the advice fits their circumstances. Since these tendencies could be specific to the tasks evaluated, we compare profiles constructed from non-overlapping task subsets. The profiles continue to distinguish models and model families. This stability supports using \ourbench{} to compare models' behavioral tendencies across the 23 axes.

We examine whether these profiles can be obtained at lower cost by asking models to report how they typically behave. Self-reports match models' most common observed behaviors on roughly half of the axes, limiting their use as a substitute for trajectory-based profiling. We also examine whether agents can be prompted to adopt specific behaviors or trained to shift toward another model's behavioral profile. 
We found that some behavioral axes are much easier to steer through prompting than others. Moreover, the same axes tended to be easy or difficult to steer, suggesting that steering results from one model may carry over to related models. In contrast, fine-tuning a model on another model's trajectories with SFT or DPO produced little shift toward the target model, leaving much of the original behavioral profile intact.

In summary, our work provides a framework for describing and comparing how agents carry out everyday tasks. By combining a taxonomy of observable behaviors with profiles derived from task trajectories, \ourtax{} and \ourbench{} capture differences in how models interact with users, gather information, and deliver results. These resources complement task-performance evaluations by helping users choose agents that fit their needs and helping agent builders adapt agent behavior to users' preferences.

\section{The \ourtax{} Taxonomy}
\label{sec:taxonomy}

To characterize how agents carry out everyday tasks, we construct \ourtax{} through bottom-up human and LLM analysis of 408 task trajectories. The taxonomy organizes observable differences in how agents work into behavioral axes, each with a rubric for annotating trajectories. We first introduce the taxonomy's categories and axes (Section~\ref{sec:taxonomy-overview}), then describe the construction process from task design to axis proposal and merging (Section~\ref{sec:taxonomy-construction}). Finally, we validate \ourtax{} on held-out tasks, comparing it with existing approaches in terms of model discriminability, annotation consistency, and label diversity (Section~\ref{sec:axis-evaluation}).

\subsection{Taxonomy Overview}
\label{sec:taxonomy-overview}

\begin{wrapfigure}[17]{r}{0.40\textwidth}
\centering
\includegraphics[width=\linewidth]{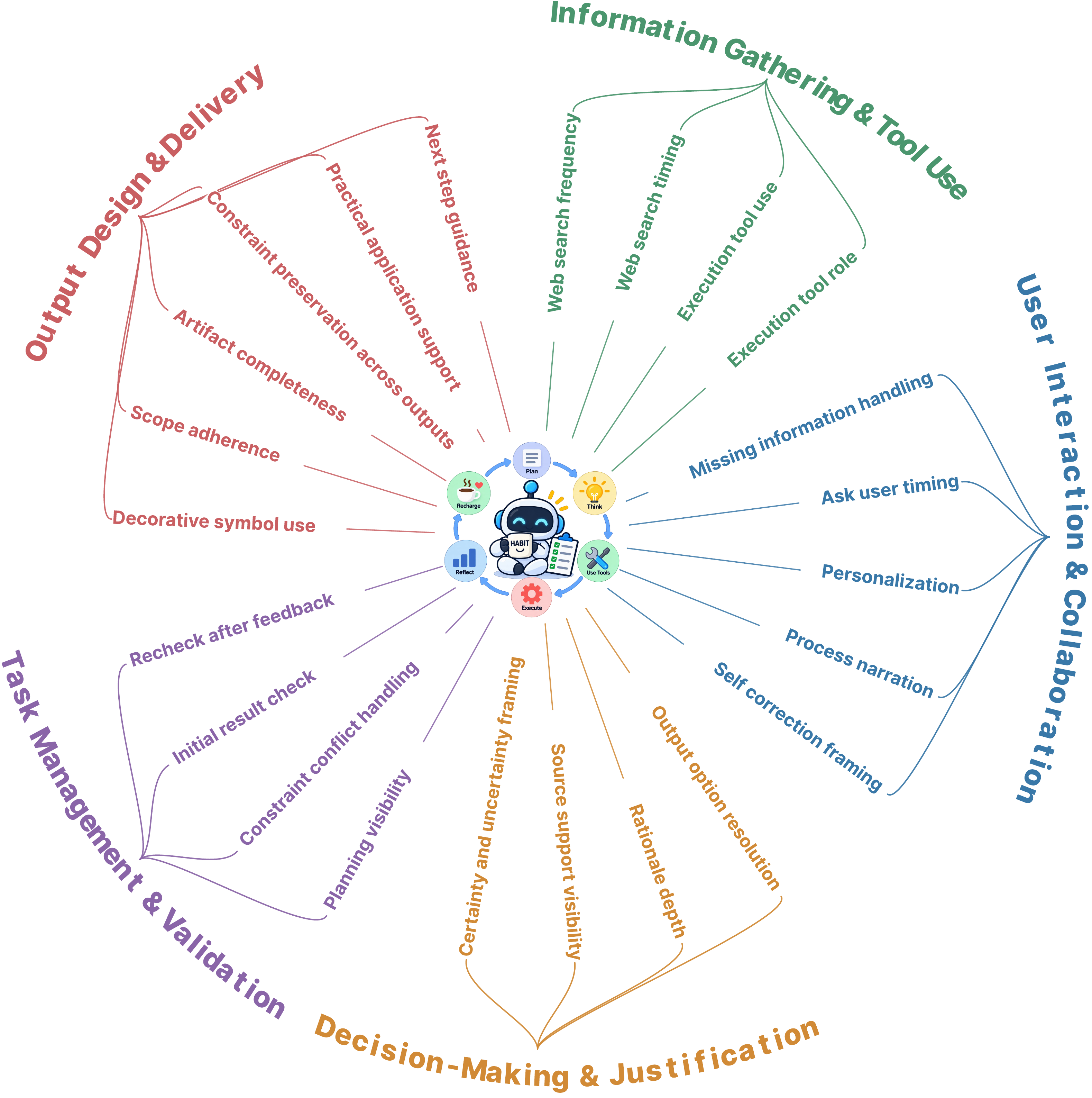}
\caption{Overview of \ourtax{}}
\label{fig:taxonomy}
\end{wrapfigure}

\ourtax{} organizes 23 behavioral axes into five broad categories (Figure~\ref{fig:taxonomy}). \textbf{User Interaction \& Collaboration} describes how agents interact with users, including when they seek clarification and how they acknowledge errors. \textbf{Information Gathering \& Tool Use} covers when and how agents use search and computational tools. \textbf{Decision-Making \& Justification} describes how agents present and prioritize options and support their conclusions. \textbf{Task Management \& Validation} covers how agents present their plans and check their work. \textbf{Output Design \& Delivery} describes what agents deliver and how they guide users after delivering an answer. Within these categories, each of the 23 axes is defined by its own rubric, which contains a behavioral definition, categorical labels that distinguish different forms of the behavior, and a decide-by rule. The decide-by rule specifies which evidence to use when assigning a label, along with any timing or precedence rules. Appendix~\ref{app:inventory} provides the full inventory.

\subsection{Constructing \ourtax{}}
\label{sec:taxonomy-construction}

\begin{figure}[t]
\centering
\includegraphics[width=\textwidth]{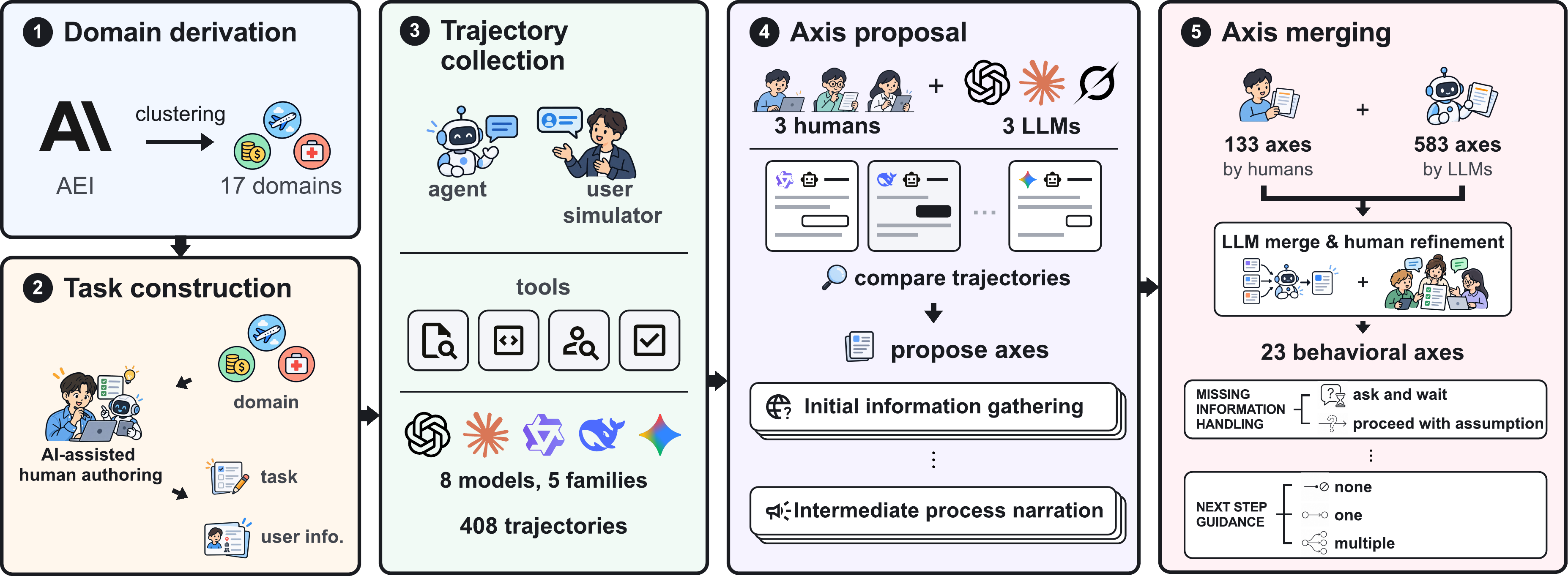}
\caption{Five-stage construction of \ourtax{}. }
\label{fig:overview}
\end{figure}

We construct \ourtax{} in five stages: domain derivation, task construction, trajectory collection, axis proposal, and axis merging (Figure~\ref{fig:overview}).

\textbf{Domain derivation.}
We derive 17 domains for open-ended tasks from the aggregate request categories of the Anthropic Economic Index (AEI)~\citep{handa2025economic} (Figure~\ref{fig:overview}, Stage~1). These categories summarize large-scale real-world Claude usage on claude.ai and the first-party API. We retain categories covering tasks that agents can approach in a variety of reasonable ways. Within each platform, we group related categories into broader domains. For example, the \emph{Home Vehicle Devices} domain brings together categories covering home heating, household maintenance, and related requests. The full derivation procedure is described in Appendix~\ref{app:domains}.

\textbf{Task construction.}
\label{sec:task-construction}
For each of the 17 domains, the authors develop one construction task with AI assistance (Figure~\ref{fig:overview}, Stage~2). Each task consists of the user's opening request and private user information that a user simulator draws on when responding to the agent. To build each task, we first specify the user's goal, relevant facts, and practical constraints. Based on this information, we use LLMs to draft requirements for the answer and possible follow-up requests. We then manually review and revise each task. For example, consider the screen-time task in Figure~\ref{fig:example}. If the agent asks which apps the child uses, the simulator can answer that the child mostly watches YouTube (Appendix~\ref{app:tasks}).

\textbf{Trajectory collection.}
\label{sec:trajectory-collection}
We run eight models on each of the 17 construction tasks three times, yielding 408 trajectories (Figure~\ref{fig:overview}, Stage~3). We select these models from five model families to capture behavioral diversity (Appendix~\ref{app:roster}). Each trajectory includes agent messages, tool calls and results, and user responses. Agents share a system prompt and have access to four tools: web search (via the \href{https://docs.tavily.com/documentation/api-reference/endpoint/search}{Tavily Search API}), Python executor, ask user, and finalize answers.

Several differences in how agents carry out everyday tasks, such as when they ask the user questions and how they revise after feedback, emerge only through interaction with a user. LLM-simulated users are widely used to evaluate agents in such multi-turn settings~\citep{yao2025taubench}. However, agents behave quite differently when users withhold information or grow impatient~\citep{shim2026noncollaborative}, so a simulator that is more cooperative than real users could misrepresent how agents behave in practice. We therefore design a user simulator based on GPT-5.6 Sol that mirrors how real users write and give feedback. We sample persona writing styles from distributions estimated from real user turns in WildChat-1M~\citep{zhao2024wildchat} and assign feedback dispositions according to their observed frequencies (Appendix~\ref{app:persona}). We use this user simulator to generate user responses during trajectory collection. It follows its assigned persona and uses the task's user information and conversation history to answer the agent's questions (Figure~\ref{fig:example}). Once the agent provides an answer to the task, the user simulator can accept it, request revisions or additional work, or end the interaction without accepting it. We cap revisions at two per run so that trajectories remain short enough for reliable LLM judging. To assess the realism of the simulator, three human annotators independently rate 74 of its utterances. They judge the utterances as highly natural, contextually appropriate, and consistent with the assigned role (mean ratings of 4.55, 4.68, and 4.57 out of 5), with substantial agreement across annotators (Gwet's AC1 of 0.62--0.71). Appendix~\ref{app:simulator-validation} provides the evaluation protocol and detailed statistical results.

\textbf{Axis proposal.}
Following qualitative content analysis~\citep{hsieh2005three}, three authors each spend more than 20 hours independently reviewing all 408 construction trajectories to identify how agents differ in carrying out tasks (Figure~\ref{fig:overview}, Stage~4). Because subtle differences are easy to overlook across this many trajectories, three LLMs from different model families independently review the same trajectories to complement the human review. Each proposer works task by task, comparing the 24 trajectories collected for a task (eight models, three runs each) and drafting candidate axes that capture how they differ. This process yields 716 candidate axes, each with a definition, categorical labels, a decide-by rule, and an explicit \texttt{N/A} condition (Appendix~\ref{app:annotation}).

\textbf{Axis merging.}
We merge and refine the candidate axes into 23 behavioral axes (Figure~\ref{fig:overview}, Stage~5). Each axis should describe a behavioral difference observed across tasks, and its rubric should apply to unseen trajectories without changes. An LLM first groups candidate axes that describe overlapping behaviors and merges each group into a single axis, rewriting its rubric so that it is not specific to any one task. Three authors then jointly review the resulting axes against these criteria. They merge axes that still capture the same behavior, remove axes whose differences do not recur across tasks, and revise definitions, categorical labels, and decide-by rules so that each rubric can be applied to new trajectories as written (Appendices~\ref{app:annotation} and~\ref{app:inventory}).

\subsection{Validating \ourtax{}}
\label{sec:axis-evaluation}

Prior work evaluates behavioral descriptions by whether they distinguish models and can be judged consistently~\citep{dunlap2025vibecheck}. We examine whether \ourtax{} meets these criteria on 408 new trajectories from 17 held-out tasks not used in taxonomy construction. As reference points, we compare \ourtax{} with two existing approaches to describing model behavior: Schwartz's ten human values used in psychometric profiling~\citep{han2025valueportrait}, which we apply through value ratings, and Act\textperiodcentered{}ONOMY's 46 agent subactions used for trace interpretation~\citep{gao2026actonomy}, which we apply through passage-level subaction annotations.

\begin{table}[htbp]
\centering
\small
\caption{Taxonomy comparison on held-out tasks. Bold indicates the highest value in each column.}
\label{tab:taxonomy-comparison}
\begin{tabular}{@{}lccccc@{}}
\toprule
Taxonomy & Matching (\%) & Gap & Entropy & Intra-AC1 & Inter-AC1 \\
\midrule
\ourtax{} & \textbf{92.2} & \textbf{0.114} & \textbf{0.653} & \textbf{0.851} & 0.718 \\
Schwartz human values & 72.8 & 0.047 & 0.491 & 0.650 & 0.383 \\
Act\textperiodcentered{}ONOMY & 87.2 & 0.044 & 0.387 & 0.830 & \textbf{0.755} \\
\bottomrule
\end{tabular}
\end{table}

To find out whether \ourtax{} can distinguish models by their behavior, we train a linear classifier to predict model identity from the behavioral labels assigned to each trajectory. We use leave-one-task-out cross-validation to evaluate matching on tasks not seen during classifier training. Across 28 model pairs, \ourtax{} achieves 92.2\% model-matching accuracy, compared with 72.8\% for Schwartz values, 87.2\% for Act\textperiodcentered{}ONOMY, and 50\% under random matching (Table~\ref{tab:taxonomy-comparison}). As a complementary measure, we calculate how much higher label agreement is between repeated runs of the same model than between different models on the same task. This difference (Gap) is 0.114 for \ourtax{}, more than twice that of either reference taxonomy. These results suggest that \ourtax{} captures behavioral tendencies that distinguish models on new tasks.

To assess whether the behavioral descriptions can be judged consistently, we compare annotations from three judge models, each annotating every trajectory three times. We measure agreement across repeated annotations and between judges using intra- and inter-judge Gwet's AC1. \ourtax{} has an intra-judge AC1 of 0.851 and an inter-judge AC1 of 0.718, compared with 0.650 and 0.383 for Schwartz values and 0.830 and 0.755 for Act\textperiodcentered{}ONOMY. The high intra-judge agreement supports consistent rubric application across repeated annotations. Inter-judge agreement is significantly higher than for Schwartz values and close to that of the binary subaction-presence summaries used for Act\textperiodcentered{}ONOMY, even though \ourtax{} distinguishes multiple forms of behavior on each axis.

Beyond these criteria, we examine how much variation the labels capture across trajectories. Even when judges agree, an axis provides little information about behavioral differences in the evaluated trajectories if nearly all receive the same label. We measure label diversity using normalized entropy, which increases as labels are used more evenly. \ourtax{} has an entropy of 0.653, compared with 0.491 for Schwartz values and 0.387 for Act\textperiodcentered{}ONOMY, indicating more varied and balanced label use. Together, these results show that \ourtax{} captures a range of behavioral differences that can be consistently annotated and used to distinguish models on new tasks. Annotation procedures and statistical comparisons are provided in Appendix~\ref{app:axis-evaluation-protocol}.

\section{The \ourbench{} Benchmark}
\label{sec:benchmark}
To profile model behavior using \ourtax{}, we construct \ourbench{}, a benchmark of 86 open-ended tasks. We first describe how we collect trajectories and build behavioral profiles from them (Section~\ref{sec:benchmark-setup}). We then present the main results (Section~\ref{sec:benchmark-results}) and examine profile stability across task subsets (Section~\ref{sec:profile-stability}).
\subsection{Benchmark Setup and Profiling}
\label{sec:benchmark-setup}

\textbf{Task construction.}
We construct 86 new open-ended tasks, one per subdomain across the 17 domains derived from AEI (Appendix~\ref{app:domains}). These tasks are distinct from the 34 tasks used to construct and evaluate \ourtax{}. Following the procedure in Section~\ref{sec:taxonomy-construction}, an LLM drafts each task for a given domain and subdomain. The draft specifies the user's goal, relevant facts, practical constraints, and requirements for the answer, along with possible follow-up requests. The authors review and revise every task for realism and overlap with existing tasks.

\textbf{Models and trajectory collection.}
We collect 4,644 trajectories by having 18 models perform each of the 86 tasks three times. We evaluate GPT-OSS 120B and GPT-5.6 Sol, Terra, and Luna; Claude Opus 5, Sonnet 5, and Haiku 4.5; Qwen 3.7 Max, Plus, and Flash; DeepSeek V4 Flash-0731; Gemini 3.7 Flash; Gemma 4 31B, 26B-A4B, and 12B; Nemotron 3.5 Lightning 30B-A3B; Muse Spark 1.3; and Solar Pro 4. We follow the same trajectory collection procedure described in Section~\ref{sec:trajectory-collection}.

\textbf{Behavioral profiling.}
We construct behavioral profiles by applying the 23 \ourtax{} rubrics to task trajectories. An LLM judge reads each trajectory, without seeing which model produced it, and assigns a categorical label on one axis at a time, citing supporting evidence from the trajectory. For judge selection, three human annotators independently apply the same rubrics to 17 trajectories on all 23 axes. We compare annotations from four LLM judges with the resulting human-majority labels and select Qwen3.8-27B. It achieves 83.52\% agreement on the 364 cases with a human-majority label (Cohen's $\kappa=0.8282$), the highest observed agreement among the four judges. The judge comparison and selection procedure are detailed in Appendix~\ref{app:human-validation}. Using the selected judge, we annotate each trajectory three times to reduce the effect of judge stochasticity. Aggregating these labels across tasks and runs yields each model's label distribution on each axis, and the 23 distributions together form its behavioral profile.

\subsection{Main Results}
\label{sec:benchmark-results}
\begin{figure}[t!]
\begin{center}
  \includegraphics[width=\textwidth]{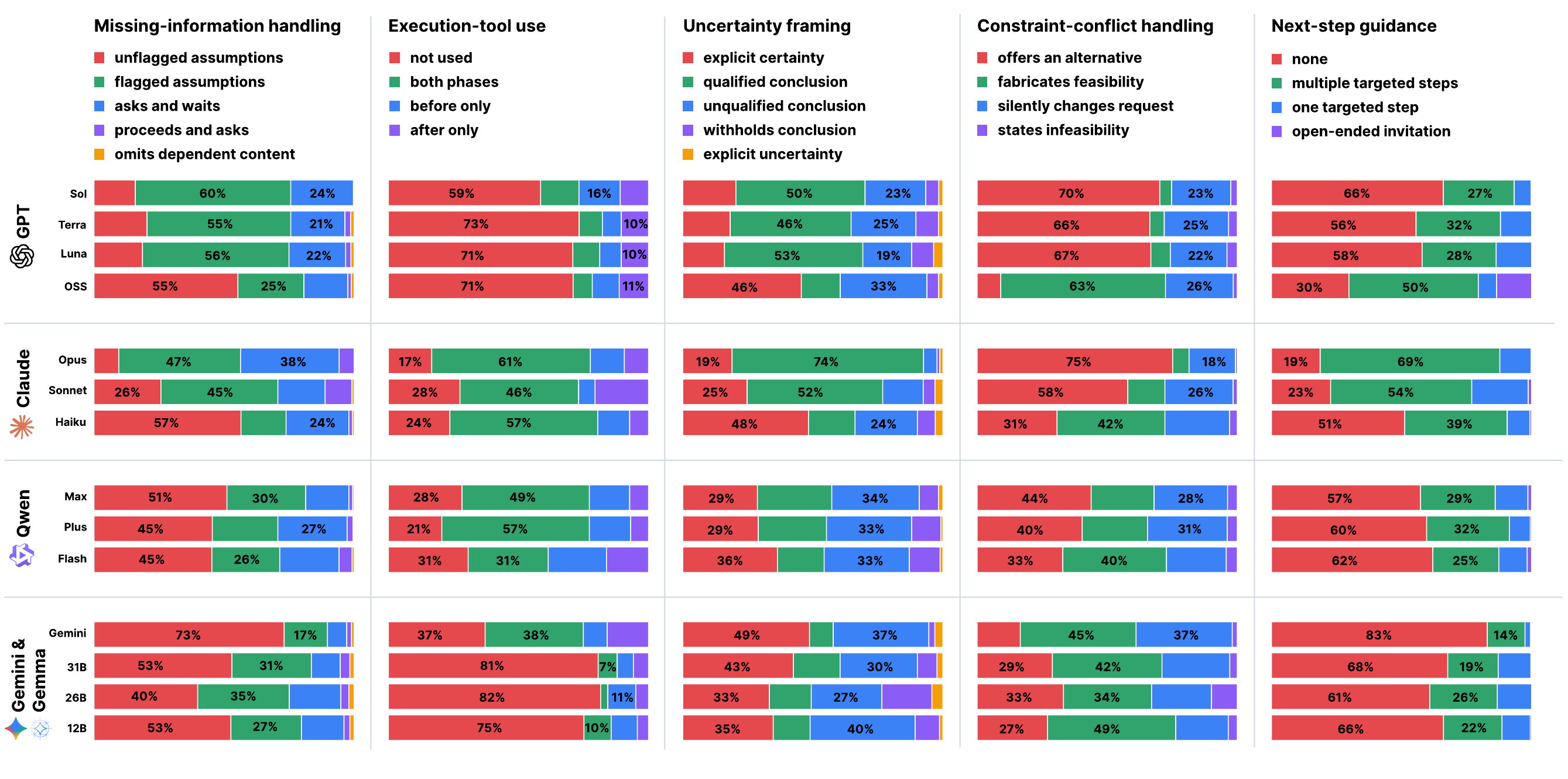}
\end{center}
\caption{Behavioral profiles on five representative HABIT axes, renormalized after excluding N/A.}
\label{fig:basic_profile}
\end{figure}

Figure~\ref{fig:basic_profile} shows behavioral profiles on one axis from each of the five broad \ourtax{} categories. Columns show behavioral axes, and rows group models by family. Each bar chart shows one model's label distribution on an axis, with colored segments indicating how often it exhibits each behavior.

\textbf{Assumptions and conflicting requirements.}
In the leftmost column, Qwen and Google's models often fill in missing information without asking the user or stating their assumptions (red). Apart from their smaller variants, GPT and Claude models tend to state their assumptions (green). Leaving such assumptions unstated can cost users time and effort. For example, a user who relies on public transport might book accommodation based on a travel plan before discovering that it assumes access to a car. Models show a similar pattern when the user's requirements cannot all be satisfied (\emph{constraint-conflict handling}, fourth column). GPT and Claude models generally explain the conflict and offer a feasible alternative (red), with exceptions among their smaller variants. Qwen and especially Google's models more often provide answers that violate or change the requirements without making this clear (green).

\textbf{Tool use and user preferences.}
The second column shows how models differ in their use of code-execution tools. GPT and Gemma models often complete tasks without these tools (red), while Claude and Qwen models often use them both before and after the first substantive deliverable (green). A similar but less uniform pattern appears in web search: Qwen models and Claude Opus 5 search more often than GPT and Gemma models (Appendix Figure~\ref{fig:appendix-profile-information-gathering}). However, which level of tool use is preferable depends on what user wants from the task. For example, when planning a trip one user may want a few quick suggestions without extensive searching, while another may want a detailed report grounded in external sources. Behavioral profiles can thus help users choose models whose tool-use tendencies match their needs.

\textbf{Guidance on interpreting and using results.}
In the third column, GPT-5.6 models tend to acknowledge uncertainty or limitations when presenting conclusions (green), while Qwen models more often leave them unstated (blue). The rightmost column shows whether models offer guidance on what to do next after delivering an answer. Claude models tend to suggest multiple concrete next steps (green), while Qwen models more often provide none (red). Some users may value guidance on when an answer applies and what to do next, while others may prefer a concise answer with fewer follow-up suggestions. Appendix~\ref{app:full-profiles} provides complete profiles for all 18 models across all 23 behavioral axes.

\begin{wrapfigure}[20]{h}{0.45\textwidth}
\centering
\includegraphics[width=\linewidth]{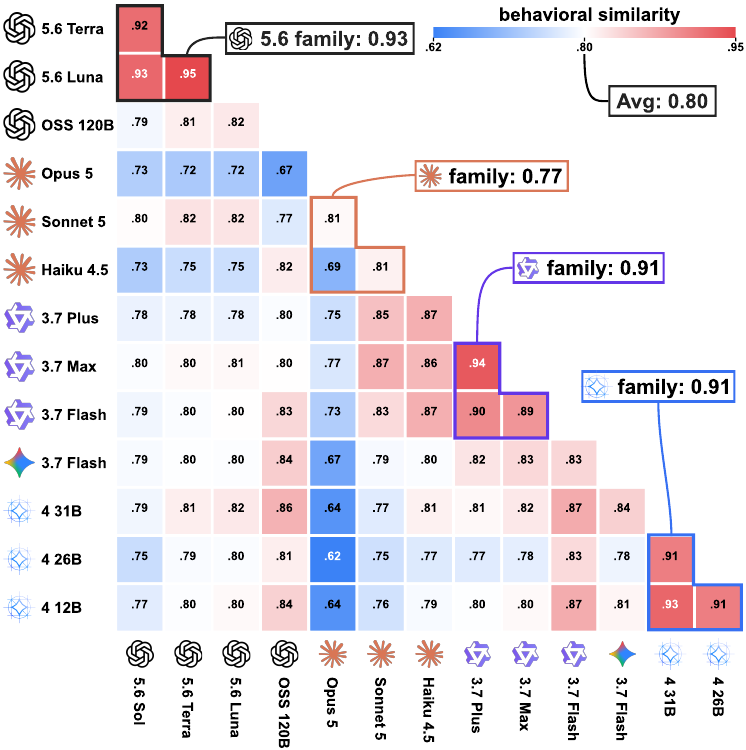}
\caption{Pairwise similarity of behavioral profiles across models.}
\label{fig:family-similarity}
\end{wrapfigure}

\textbf{Similarity within model families.}
Figure~\ref{fig:family-similarity} shows pairwise similarities between model profiles. We define \emph{profile similarity} as one minus the mean total variation (TV) distance between label distributions across the 23 behavioral axes. Higher values indicate more similar behavioral profiles. Mean within-family similarity is 0.93 for GPT-5.6, 0.91 for Qwen, and 0.91 for Gemma. All three exceed the overall mean of 0.80 across the model pairs in the figure (all $p<.001$, Appendix~\ref{app:family-similarity-tests}). Claude has a lower mean similarity of 0.77, with a particularly large difference between Opus and Haiku. Thus, while models from the same family tend to behave similarly, this does not always hold, and family alone is not a reliable guide when choosing an agent.

\subsection{Profile Stability Across Task Subsets}
\label{sec:profile-stability}

To assess whether \ourbench{} profiles are stable across tasks, we compared profiles constructed from disjoint task subsets. We sampled subsets from the 86 tasks in \ourbench{} at 1\%, from 10\% to 90\% in increments of 10 percentage points, and at 99\%. For each model, we constructed separate profiles using the sampled tasks and the remaining, non-sampled tasks. Using each subset profile as a query, we identified the nearest profile constructed from the remaining tasks and measured whether the matched profile belonged to the same model or to a model from the same family. We repeated this process 1,000 times for each sampling ratio and report the average results. With only 10\% of the tasks, 80.21\% of the subset profiles were matched to the same model, while 98.90\% were matched to a model from the same family (Table~\ref{tab:full-profile-stability}). When 50\% of the tasks were used, these rates increased to 89.86\% and 100.00\%, respectively. These results indicate that \ourbench{} captures task-general model profiles rather than profiles specific to particular tasks. Full results are provided in Appendix~\ref{app:profile-stability}.

\section{Analysis}

\subsection{Can Self-Reports Recover Behavioral Profiles?}
\label{sec:self-report}

\begin{table*}[ht]
\centering
\caption{Self-report--behavior correspondence. \textit{Mode} denotes Mode Agreement (\%).}
\label{tab:self-report-behavior}
\small
\setlength{\tabcolsep}{4pt}
\resizebox{\textwidth}{!}{%
\begin{tabular}{l ccc c ccc ccc c ccc cc}
\toprule
 & \multicolumn{3}{c}{GPT-5.6} & GPT-OSS & \multicolumn{3}{c}{Claude} & \multicolumn{3}{c}{Qwen 3.7} & Gemini 3.7 & \multicolumn{3}{c}{Gemma 4} & & \\
\cmidrule(lr){2-4} \cmidrule(lr){5-5} \cmidrule(lr){6-8} \cmidrule(lr){9-11} \cmidrule(lr){12-12} \cmidrule(lr){13-15}
Metric & Sol & Terra & Luna & 120B
       & Opus 5 & Sonnet 5 & Haiku 4.5
       & Max & Plus & Flash
       & Flash
       & 31B & 26B & 12B
       & Mean & Random \\
\midrule
Mode (\%) & 65.2 & 69.6 & 47.8 & 47.8 & 56.5 & 65.2 & 34.8 & 43.5 & 47.8 & 56.5 & 43.5 & 43.5 & 39.1 & 52.2 & 50.9 & 33.3 \\
\bottomrule
\end{tabular}%
}
\end{table*}

Our profiles are based on behavior observed during task execution, but obtaining them requires collecting and annotating full trajectories. Self-reports could offer a less costly alternative. Prior work has used self-reports to profile LLMs' personalities, values, and behavioral tendencies~\citep{huang2024on, lim2026psychometric, jiang2023evaluating}. However, LLMs' self-reported values do not always align with their response tendencies in everyday scenarios~\citep{song2026humanpsychometricquestionnairesmischaracterize, han-etal-2026-quantifying}. We therefore test whether models can accurately report their typical behavior in \ourbench{} tasks. For each of the 23 axes, models answered a multiple-choice question (e.g., ``How does the assistant respond when the requested constraints cannot all be satisfied?'') by selecting the option that best described their typical behavior. We repeated each question ten times for each of the 14 models, shuffling the options each time, and aggregated the responses into self-reported profiles for comparison with the corresponding \ourbench{} profiles. Appendix~\ref{app:self-report} provides the full questionnaire and protocol.

Table~\ref{tab:self-report-behavior} shows that self-reports only partly reflect observed behavior. \textit{Mode Agreement} measures whether the behavior a model selects most often in the questionnaire is also its most common behavior. Mode Agreement averages 50.9\% across axes and models, above the uniform random baseline of 33.3\% but still leaving roughly half the axes with a mismatch. Thus, self-reports alone do not reliably identify models' most common behaviors.

\subsection{Dependence Across Behavioral Axes}
\label{sec:axis-dependence}
If several axes capture the same behavioral pattern, they may carry redundant information in a model's profile. We therefore examine how much information the 23 axes of \ourtax{} share across task trajectories. For each axis pair, we compute Cram\'er's $V$ from how often their labels occur together. $V$ measures the strength of association between the two categorical axes, with higher values indicating stronger association. We also examine how much knowing the label on one axis tells us about the label on another. For this, we use the uncertainty coefficient $U$, which measures the fraction of uncertainty about one axis's label reduced by knowing the other. We compute $U$ in both directions because one axis may explain more of the other than vice versa.

Although some behavioral axes are associated, knowing the label on one axis generally provides limited information about the label on another. Under Cohen's effect-size conventions for Cram\'er's $V$, 53 of the 253 axis pairs (20.9\%) show medium or large associations~\citep{cohen1988statistical}. However, across all pairs and both directions, the mean reduction in uncertainty measured by $U$ is 3.40\% ($U=0.0340$). For example, knowing how clearly an agent identifies sources for its claims reduces uncertainty about how it handles missing information by only 1.40\% ($U=0.0140$). These results suggest that most of the axes capture different aspects of agent behavior. Details are in Appendix~\ref{app:axis-dependence}.

\subsection{Steering Agent Behavior}
\label{sec:behavior-steering}

The profiles in Section~\ref{sec:benchmark} show how models tend to behave when carrying out tasks. Can we also change these tendencies to better match our preferences? We examine two approaches: prompting agents to exhibit specific behaviors and fine-tuning them on another model's trajectories. We use Gemma 4 12B for both approaches because it is the only model in our evaluation that we can fine-tune with our available computing resources. We also apply prompt-based steering to Gemma 4 31B to examine whether similar effects occur in a larger model from the same family.

\begin{wraptable}{r}{0.6\textwidth}
    \vspace{-12pt}
    \centering
    \caption{Prompt-steering target-label rates (\%) for six representative shared-target axes. $\Delta$ denotes the absolute change.}
    \label{tab:prompt-steering}
    \setlength{\tabcolsep}{3pt}
    \renewcommand{\arraystretch}{1.1}
    \resizebox{\linewidth}{!}{%
    \begin{tabular}{@{}lrrrrrr@{}}
    \toprule
    & \multicolumn{3}{c}{Gemma 4 31B}
    & \multicolumn{3}{c}{Gemma 4 12B} \\
    \cmidrule(lr){2-4}\cmidrule(l){5-7}
    Axis & Base & Steered & $\Delta$
    & Base & Steered & $\Delta$ \\
    \midrule
    Decorative Symbol Use
    & 16.5 & 98.4 & +81.9 & 3.2 & 96.5 & +93.3 \\
    Next Step Guidance
    & 0.3 & 45.0 & +44.7 & 0.5 & 22.9 & +22.4 \\
    Web Search Timing
    & 4.3 & 37.2 & +32.9 & 4.7 & 28.3 & +23.6 \\
    Missing Information Handling
    & 1.6 & 11.6 & +10.1 & 1.6 & 1.6 & 0.0 \\
    Source Support Visibility
    & 4.1 & 6.6 & +2.5 & 3.2 & 8.5 & +5.3 \\
    Process Narration
    & 0.6 & 2.7 & +2.1 & 7.6 & 17.4 & +9.8 \\
    \midrule
    Macro-avg.\ (23 axes)
    & 3.6 & 23.0 & +19.5 & 3.3 & 21.5 & +18.2 \\
    \bottomrule
    \end{tabular}}
    \vspace{-10pt}
\end{wraptable}

\textbf{Prompt-based steering.}
To make changes easier to observe, we prompted each model to adopt its least frequent behavior on each of the 23 axes. We compared how often these behaviors occurred with and without prompting. As shown in Table~\ref{tab:prompt-steering}, the frequency of the target behavior, averaged across the 23 axes, rose from 3.6\% to 23.0\% for 31B and from 3.3\% to 21.5\% for 12B. In both models, instructions to use decorative symbols extensively had large effects. Instructions to search only after the first substantive deliverable led to more moderate increases in the requested behavior, while instructions to cite sources throughout the answer had much smaller effects. More broadly, behaviors that changed more in one model also tended to change more in the other. Across the 22 axes with the same target behavior and instruction, the changes were strongly correlated (Pearson $r=0.909$). Results from one model may therefore help guide attempts to steer another model from the same family. Details are in Appendix~\ref{app:prompt-steering}.

\textbf{Training-based steering.}
We tested whether training on another model's trajectories could bring their behavioral profiles closer. We chose Claude Opus 5, whose profile was the most distant from Gemma 4 12B's in our evaluation (Figure~\ref{fig:family-similarity}). We trained separate models using SFT and DPO on \ourbench{} tasks and evaluated them on the held-out \ourtax{} tasks used in Section~\ref{sec:axis-evaluation}. SFT used Opus responses as training targets, while DPO favored Opus trajectories over Gemma trajectories from the same task. We measured these changes using the profile similarity defined in Section~\ref{sec:benchmark-results}. Similarity to Opus changed only slightly, from approximately 0.604 before training to 0.601 after SFT and 0.607 after DPO. Neither change was statistically significant. The trained models also remained close to the original Gemma profile, with similarities of 0.908 for SFT and 0.932 for DPO. These results suggest that training on another model's trajectories alone may not be enough to produce the intended behavioral changes. Details are in Appendix~\ref{app:profile-transfer-training}.

\section{Related Work}
\label{sec:related}

\paragraph{Evaluating agent outcomes and processes.}
Agent evaluations primarily measure task success in domains such as coding, computer use, and web navigation~\citep{jimenez2024swebench,zhou2024webarena,xie2024osworld,liu2024agentbench}. Other benchmarks provide finer-grained assessments by measuring intermediate progress~\citep{ma2024agentboard} or task-completion efficiency~\citep{xu2025crab}. Together, these benchmarks provide systematic measures of agent performance. While these benchmarks reveal how well agents complete tasks, they offer limited insight into agents' behavior during task execution, such as how thoroughly they verify their work or how they handle ambiguous instructions. We instead characterize agents by the behavioral tendencies that recur across tasks.

\paragraph{Profiling large language models.}
Research on model profiling broadly follows two lines: psychometric profiling and purpose-specific behavioral analysis. Psychometric constructs, such as Schwartz's theory of basic human values~\citep{schwartz1992universals}, have been widely used to profile models by scoring their responses to questionnaire-style items~\citep{ren2024valuebench,han2025valueportrait,zhao2024worldvaluesbench,jiang2024personallm,lee2025trait}. Human values have also been used to profile agents based on their task-execution trajectories~\citep{dong2026agentvaluebench}. Rather than applying predefined inventories, bottom-up methods inductively derive evaluation criteria or value profiles from model outputs~\citep{perez2023modelwritten} and real-world interactions~\citep{huang2025valueswild}. These approaches characterize models in terms of values and traits.  However, they do not establish links between value or trait scores and agents' actions during task execution. A high openness score does not indicate whether an agent will explore multiple alternatives, seek additional information, or commit early to a single approach.

Some studies analyze agent behavior for specific analytical purposes. These include hierarchical action taxonomies derived through grounded theory for trace interpretation and oversight~\citep{gao2026actonomy}, and categorizations of control decisions and failure modes for benchmark auditing and failure diagnosis~\citep{mazaheri2026agentatlas}. Cross-task behavioral consistency has also been measured using predefined features weighted by contributions to task success~\citep{banerjee2026crosstask}. While these approaches support trace interpretation, diagnostic evaluation, and reliability assessment, \ourtax{} provides broader, more interpretable profiles of agents' behavioral tendencies across tasks.

\section{Conclusion}
We introduced \ourtax{}, a taxonomy of 23 behavioral axes, and \ourbench{}, a benchmark designed to profile how LLM agents carry out everyday tasks. Through behavioral profiles of 18 models, we find that agents differ systematically in how they seek information, handle user requirements, and present results, with these profiles distinguishing individual models, recurring within several model families, and remaining stable across disjoint task subsets. Our steering experiments further show that these tendencies are not easily changed, as prompting shifts some behaviors far more than others and fine-tuning on another model's trajectories leaves much of the original profile intact. By grounding behavioral descriptions in what agents actually do during task execution, \ourtax{} and \ourbench{} provide a foundation for comparing agents beyond task success and for developing methods to shape their behavior. We hope these resources help users choose agents that suit their ways of working and guide the development of agents that better fit users' needs.

\subsection*{AI use statement}
In this work, we used generative AI tools to generate synthetic datasets, implement methods, assist with translation, clean and reformat datasets, and support qualitative and thematic data analysis. We have not used generative AI tools to design or provide feedback on research methodology or experiments, or to interpret results. Developing theoretical models or conceptual frameworks, formulating mathematical claims, providing critical ingredients for proving mathematical claims, assisting in the writing of proofs, and proposing or refining hypotheses are not applicable to this work. Additionally, we used generative AI tools to create and edit software code and to edit the paper for improved readability, including grammar correction. We have reviewed all AI-assisted work. All AI-generated code was reviewed by the authors, and all AI-assisted translations were also reviewed by the authors. We take responsibility for the final content of this work, including text, claims, or artifacts produced with the aid of generative AI.

\subsection*{Reproducibility statement}
We plan to publicly release our code and data, including task definitions, user personas, prompts, trajectories, annotations, and analysis scripts. For Tavily search results, we release only the source URLs and exclude the retrieved content, as it consists of third-party web content that may be subject to copyright. We also provide detailed experimental settings in the appendices of this paper. Taken together, these resources will support the verification of our findings and future research.

Several considerations may limit exact reproducibility. Results may vary due to changes to or retirement of API-based models, stochastic language model outputs, and changes in search results over time. To mitigate the effects of these factors, we collect three trajectories for each model–task pair and annotate each trajectory three times. While these repetitions help reduce statistical uncertainty, some degree of uncertainty may still remain.

\bibliography{iclr2027_conference}
\bibliographystyle{iclr2027_conference}

\clearpage

\appendix

\section{Domains}
\label{app:domains}

\begingroup
\footnotesize
\renewcommand{\arraystretch}{1.08}
\setlength{\tabcolsep}{4pt}
\setlength{\LTleft}{0pt}
\setlength{\LTright}{\fill}
\setlength{\LTcapwidth}{\textwidth}
\begin{longtable}{@{}r >{\raggedright\arraybackslash}p{0.25\textwidth} >{\raggedright\arraybackslash}p{0.66\textwidth}@{}}
\caption{The 17 task domains and 86 subdomains derived from AEI request categories, grouped by platform.}
\label{tab:domains}\\
\toprule
\textbf{\#} & \textbf{Domain} & \textbf{Subdomains} \\
\midrule
\endfirsthead
\multicolumn{3}{l}{\tablename~\thetable\ (continued)}\\
\toprule
\textbf{\#} & \textbf{Domain} & \textbf{Subdomains} \\
\midrule
\endhead
\midrule
\multicolumn{3}{r}{Continued on next page}\\
\endfoot
\bottomrule
\endlastfoot
\multicolumn{3}{@{}l}{\textit{claude.ai-derived domains (56 subdomains)}}\\*
\addlinespace[2pt]
1 & Business Marketing Operations & Business Presentations \& Process Diagrams\newline Startup \& Business Strategy\newline Digital Marketing \& Social Media\newline Financial Analysis \& Spreadsheets\newline Company Research \& Business Intelligence\newline Business Operations \& Logistics \\
\addlinespace[3pt]
2 & Academic Education Research & Educational Curriculum \& Teaching Materials\newline Social Sciences \& Humanities\newline Academic Research Methods \& Statistical Analysis\newline Academic Coursework \& Research Writing\newline Books \& Academic Literature\newline Education \& Healthcare Training \\
\addlinespace[3pt]
3 & Health Wellness Relationships & Personal Development \& Organizational Leadership\newline Recipes \& Nutrition Planning\newline Beauty Fashion \& Wellness Products\newline Fitness Training \& Weight Management\newline Psychology \& Neuroscience\newline Family Life \& Social Benefits\newline Medical \& Health \\
\addlinespace[3pt]
4 & STEM Sciences Engineering & Math \& Physics Problems\newline Earth \& Space Sciences\newline Multidisciplinary Engineering\newline Chemistry \& Life Sciences\newline AI \& Machine Learning\newline Optimization Algorithms \& System Performance \\
\addlinespace[3pt]
5 & Creative Media Design & UI and Brand Design\newline Architecture \& Visual Arts\newline Fiction \& Public Affairs\newline Video \& Podcast Production\newline Music \& Film/TV\newline Video Games \& Tabletop Roleplay \\
\addlinespace[3pt]
6 & Legal Finance Careers & Personal Finance \& Investing\newline Legal Documents \& Law\newline Job Applications \& Career Development\newline Immigration Visas \& Real Estate \\
\addlinespace[3pt]
7 & Software Data Systems & Network \& Domain Administration\newline File Systems \& Storage\newline Software \& Device Configuration\newline Software Project Management \& Mobile Apps\newline Blockchain \& Software Development\newline Geospatial, Cloud \& Data Engineering\newline Conversation History \& Mixed Queries\newline Document \& Image Data Extraction \\
\addlinespace[3pt]
8 & Places People Planning & Product Shopping Recommendations\newline Travel Planning \& Local Places\newline Geographic Demographics \& Regions\newline Scheduling \& Time Management\newline Celebrations, Gifts \& Weddings\newline Names, Genealogy \& Biographies \\
\addlinespace[3pt]
9 & Home Vehicle Devices & Home Energy \& Utility Systems\newline Audio Equipment \& Device Troubleshooting\newline Vehicle Ownership \& Mechanics \\
\addlinespace[3pt]
10 & Security Codes Safety & Cybersecurity \& Workplace Safety\newline Codes, Puzzles \& Lottery \\
\addlinespace[3pt]
11 & Sports Performance Management & Sports Data \& Events\newline Sports Management \& Coaching \\
\addlinespace[3pt]
\midrule
\multicolumn{3}{@{}l}{\textit{First-party API-derived domains (30 subdomains)}}\\*
\addlinespace[2pt]
12 & Business Intelligence Operations & Business Intelligence \& Financial Markets\newline Business Communications \& Quality Evaluation\newline B2B Sales Enablement \& Intelligence\newline Business Operations \& Project Management\newline Business Strategy \& Startup Fundraising\newline Business Intelligence \& Professional Profiles\newline Recruitment \& Job Matching\newline Business Documents \& Spreadsheets \\
\addlinespace[3pt]
13 & Consumer Commerce Services & SEO \& Digital Marketing Copy\newline Automotive Sales \& Logistics\newline Appointments \& Travel\newline Real Estate \& Digital Advertising\newline Consumer Products \& E-commerce\newline Payments \& Consumer Finance \\
\addlinespace[3pt]
14 & Education Health Research & Education \& Academic Tutoring\newline Clinical Data \& Documentation\newline Academic Research Data \& Writing\newline Mental Health \& Crisis Intervention \\
\addlinespace[3pt]
15 & Regulated Document Analysis & Document Intelligence\newline Legal Documents \& Research\newline Industrial Product Data \& Trade Compliance \\
\addlinespace[3pt]
16 & Technical Systems Security & Technical Documentation Systems\newline AI System Evaluation \& Security\newline Cybersecurity Operations \& Threat Analysis\newline IT Systems \& Network Devices\newline AI Agents \& Game Interaction \\
\addlinespace[3pt]
17 & Visual Geospatial Media & Digital Design \& Data Visualization\newline AI Video Production Workflows\newline Image \& Screenshot Analysis\newline Energy \& Environmental Geospatial Data \\
\addlinespace[3pt]
\end{longtable}
\endgroup

\paragraph{Source inventory and filtering.}
We collect request-cluster names from four public AEI releases (March 27, 2025; September 15, 2025; January 15, 2026; and March 24, 2026) and deduplicate them, yielding 2{,}821 unique request categories. A fixed LLM rubric retains categories describing agentic, open-ended tasks that admit several reasonable approaches or outputs and excludes categories primarily concerned with software engineering. This leaves 1{,}058 unique categories (876 claude.ai records and 266 first-party API records, with 84 appearing on both platforms). Categories may occur in both platform inventories.

\paragraph{Domain derivation.}
We process the platform inventories separately. We use Qwen3-Embedding-8B to embed the retained category names and apply adaptive Ward clustering, yielding 56 claude.ai subdomains and 30 first-party API subdomains. An LLM names and groups these subdomains into 11 claude.ai domains and 6 API domains. Table~\ref{tab:domains} lists all 17 domains and their 86 subdomains. For \ourtax{}, each construction and held-out split contains one task per domain; \ourbench{} contains one task per subdomain.

\FloatBarrier
\section{Task Construction and User Context}
\label{app:tasks}

This section describes the 34 tasks used to construct and evaluate \ourtax{} and the private user context shown as \emph{user information} in Figure~\ref{fig:overview}.

Each domain contains a construction task and a held-out task, with different user situations, requested outputs, and private contexts. Authors develop the tasks from aggregate AEI request categories. GPT-5.6 Sol adds output requirements and optional follow-up requests to the user contexts. Authors refine these additions for consistency, coverage of user needs, and feasible information-disclosure conditions. Automated checks validate the schema, and authors review and revise every task.

Table~\ref{tab:task-openings} lists the opening requests for all 34 tasks.

\begin{nolinenumbers}
\begingroup
\footnotesize
\renewcommand{\arraystretch}{1.05}
\setlength{\tabcolsep}{3pt}
\setlength{\LTleft}{0pt}
\setlength{\LTright}{0pt}
\setlength{\LTcapwidth}{\linewidth}
\captionsetup{singlelinecheck=false}
\begin{longtable}{@{}>{\raggedright\arraybackslash}p{0.23\linewidth} >{\raggedright\arraybackslash}p{0.15\linewidth} >{\raggedright\arraybackslash}p{0.56\linewidth}@{}}
\caption{Verbatim opening requests for the 17 construction and 17 held-out tasks.}
\label{tab:task-openings}\\
\toprule
\textbf{Domain} & \textbf{Task} & \textbf{Opening request} \\
\midrule
\endfirsthead
\multicolumn{3}{c}{\tablename\ \thetable\ (continued)}\\
\toprule
\textbf{Domain} & \textbf{Task} & \textbf{Opening request} \\
\midrule
\endhead
\midrule
\multicolumn{3}{r}{Continued on next page}\\
\endfoot
\bottomrule
\endlastfoot
\multicolumn{3}{@{}l}{\textbf{Construction tasks}}\\
\addlinespace[1pt]
Business Marketing Operations & Marketing budget & How should I split our \$12,000 monthly ad budget next quarter across Google Search, Instagram, and paid influencers? \\
Academic Education Research & Science unit plan & Can you help me plan eight 50-minute lessons on energy transfer for a class of 28? \\
Health Wellness Relationships & Skin products & I need an affordable skincare routine to help with ongoing breakouts along my jawline. I’d like to keep it around \$60 a month. \\
STEM Sciences Engineering & Home solar & I own a house in New Jersey and spend about \$2,400 a year on electricity. Are rooftop solar panels worth the money, and should I buy them? \\
Creative Media Design & Signup flow & Could you recommend a better order for our seven-screen mobile web signup flow based on last month’s funnel? Signup drop-off is bad, and I want an actionable sequencing change. \\
Legal Finance Careers & Going freelance & Can you give me an actionable checklist for leaving my salaried graphic design job and going freelance? \\
Software Data Systems & DB backfill & I need a script to normalize the country column in our production Postgres users table in one pass. \\
Places People Planning & Neighborhood pick & Rank the top three Boston neighbourhoods for finding a flat within my firm \$3,200 monthly budget and commuting to Kendall Square in under an hour. \\
Home Vehicle Devices & Heating bill & how can i get my UK energy bill down without letting the flat drop below 20c, ive got a 3 month old so thats not negotiable \\
Security Codes Safety & Email breach & Someone got into my Gmail, and there are messages in my sent folder that I did not write. Give me a numbered list of what to do, in order. \\
Sports Performance Management & League schedule & I need a fixture list for our eight-team Saturday league where everyone plays everyone once. We have ten Saturdays and two pitches. \\
Business Intelligence Operations & Survey priorities & I need to turn our internal survey on workplace tools into a one-slide recommendation naming the single priority to fix next quarter. \\
Consumer Commerce Services & Blog SEO & Our organic traffic is down about 70\% in three months. Help me figure out how to recover it without just writing more blog posts. \\
Education Health Research & Kid math & My eleven-year-old’s maths marks are slipping, and I’m considering increasing tutoring from one hour to three hours a week. Is that advisable, and how should I structure the three hours? \\
Regulated Document Analysis & Loan compare & which loan should i go for on £40,000 over five years, offer A fixed 5.2\% or offer B variable 4.6\%? \\
Technical Systems Security & API docs & Could you recommend one tool for updating and maintaining our internal REST API documentation? Please give me a concrete choice rather than a comparison. \\
Visual Geospatial Media & Sales dashboard & Can you help me design a sales dashboard for leadership that makes current-month revenue immediately visible? \\
\addlinespace[2pt]
\midrule
\multicolumn{3}{@{}l}{\textbf{Held-out tasks}}\\
\addlinespace[1pt]
Business Marketing Operations & Agency switch & Should we keep, renegotiate or leave our marketing agency? We’re six months in on £6,500 a month and inbound leads are down. \\
Academic Education Research & Lit review & I need help planning a 6,000-word literature review on whether school uniform policy affects attendance. I have six weeks until submission. \\
Health Wellness Relationships & Sleep fix & I keep waking around 3 a.m. and can’t get back to sleep. I need a plan that actually targets that. \\
STEM Sciences Engineering & Well water & I bought a property with a private well and need help understanding the water test. Can you compare each result with the relevant health or aesthetic benchmark and tell me what requires action? \\
Creative Media Design & Brand refresh & I want to redo our bakery brand so it feels less rustic and more grown up. We need it ready in about three weeks for new packaging. \\
Legal Finance Careers & Equity offer & Should I take the startup offer or stay at my current job? I want a clear recommendation, not just pros and cons. \\
Software Data Systems & Slow query & My Django/Postgres dashboard takes about eight seconds to load. What should I do to speed it up? \\
Places People Planning & Office move & need help finding a cheaper office in Bristol for 30 people, max budget is £4,500 a month. \\
Home Vehicle Devices & Used EV & Can you sanity-check a 2021 electric car with 68,000 km and a £16,500 asking price? Tell me if it’s reasonable value and whether I should buy it. \\
Security Codes Safety & Shared credentials & Which password manager should we buy for our six-person team to replace the passwords we currently keep in a pinned Slack message? \\
Sports Performance Management & Injury return & I need a return-to-play plan for a player nine weeks out from a lower-leg fracture, aiming to be available for a tournament in five weeks. \\
Business Intelligence Operations & Forecast gap & Help me explain why Q4 revenue came in at \$3.1M against a \$4.2M forecast for Thursday’s board meeting. \\
Consumer Commerce Services & Return policy & I need to tighten my returns policy, 7 days for change of mind and customer pays return postage. can you help word it \\
Education Health Research & Screen time & I need a daily tablet limit for my seven-year-old and a specific app or router control that automatically blocks it when time’s up. \\
Regulated Document Analysis & Tenancy clause & Are these tenancy clauses enforceable in England: a £75 charge for late rent and allowing the landlord to enter at any time for inspection? \\
Technical Systems Security & Incident postmortem & I need help writing a blameless postmortem for the three-hour payments outage Tuesday. its going to leadership Friday. \\
Visual Geospatial Media & Branch map & I need a map showing how each of our 40 branches is performing. \\
\end{longtable}
\endgroup
\end{nolinenumbers}

Each user context specifies the situation, output requirements, optional follow-up requests after a draft, and five classes of information or decisions. \emph{Volunteered} facts may appear in the opening request; \emph{on-request} facts are disclosed when the agent asks about them; \emph{guarded} facts additionally require a specified conversational condition; \emph{withheld} facts remain unavailable; and \emph{delegated} decisions are left to the agent. Only the user simulator receives the full context; agents, axis proposers, and annotation judges do not.

With user context, GPT-5.6 Sol generates one opening request per task from the situation, volunteered facts, and assigned writing style. We cache this request and reuse it unchanged across models and runs. Later simulator responses depend on the agent's questions and deliverables under the same disclosure rules. We use \nolinkurl{gpt-5.6-sol} as the user simulator.

\FloatBarrier
\section{Model Roster}
\label{app:roster}

\begin{table}[tbp]
\caption{Model routes and run settings. $\dagger$ marks the eight \ourtax{} construction models; $T$ is temperature and $R$ is requested reasoning effort.}
\label{tab:agent-roster}
\centering
\footnotesize
\setlength{\tabcolsep}{3pt}
\renewcommand{\arraystretch}{1.12}
\begin{tabular}{@{}>{\raggedright\arraybackslash}p{0.13\linewidth}
>{\raggedright\arraybackslash}p{0.32\linewidth}
>{\raggedright\arraybackslash}p{0.20\linewidth}
>{\raggedright\arraybackslash}p{0.23\linewidth}@{}}
\toprule
\textbf{Family} & \textbf{Model route} & \textbf{Access route} & \textbf{Run settings} \\
\midrule
GPT & \nolinkurl{gpt-5.6-sol}$^{\dagger}$ & OpenAI API & $R=$ medium \\
 & \nolinkurl{gpt-5.6-terra}$^{\dagger}$ & OpenAI API & $R=$ medium \\
 & \nolinkurl{gpt-5.6-luna} & OpenAI API & $R=$ medium \\
GPT-OSS & \nolinkurl{openai/gpt-oss-120b} & Local vLLM & $T=0.7$ \\
Claude & \nolinkurl{anthropic/claude-opus-5}$^{\dagger}$ & OpenRouter (Anthropic) & $R=$ medium \\
 & \nolinkurl{anthropic/claude-sonnet-5}$^{\dagger}$ & OpenRouter (Anthropic) & $R=$ medium \\
 & \nolinkurl{anthropic/claude-haiku-4.5} & OpenRouter (Anthropic) & $R=$ medium \\
Qwen & \nolinkurl{qwen/qwen3.7-max}$^{\dagger}$ & OpenRouter (Alibaba) & $T=0.7$; $R=$ medium \\
 & \nolinkurl{qwen/qwen3.7-plus}$^{\dagger}$ & OpenRouter (Alibaba) & $T=0.7$; $R=$ medium \\
 & \nolinkurl{qwen/qwen3.7-flash} & OpenRouter (Alibaba) & $T=0.7$; $R=$ medium \\
 & \nolinkurl{Qwen/Qwen3.8-27B} (judge) & Local vLLM & $R=$ low \\
Gemini & \nolinkurl{google/gemini-3.7-flash}$^{\dagger}$ & OpenRouter (Google Vertex) & $R=$ medium \\
DeepSeek & \nolinkurl{deepseek-ai/DeepSeek-V4-Flash-0731}$^{\dagger}$ & Local vLLM & $T=0.7$; thinking off \\
Gemma & \nolinkurl{google/gemma-4-31B-it} & Local vLLM & $T=0.7$; thinking on \\
 & \nolinkurl{google/gemma-4-26B-A4B-it} & Local vLLM & $T=0.7$; thinking on \\
 & \nolinkurl{google/gemma-4-12B-it} & Local vLLM & $T=0.7$; thinking on \\
Nemotron & \nolinkurl{nvidia/NVIDIA-Nemotron-3.5-Lightning-30B-A3B-BF16} & Local vLLM & $T=0.7$; thinking on \\
Muse & \nolinkurl{meta/muse-spark-1.3-contributor} & OpenRouter (Meta) & $T=0.7$; $R=$ medium \\
 & \nolinkurl{meta-models/Muse-Glimmer-30B} (judge candidate) & Local vLLM & $R=$ low \\
Solar & \nolinkurl{upstage/solar-pro4} & OpenRouter (Upstage) & $T=0.7$ \\
\bottomrule
\end{tabular}
\end{table}

We use eight models to construct \ourtax{} and evaluate 18 models on \ourbench{} (these eight plus ten additional models). Table~\ref{tab:agent-roster} lists the model routes and run settings. We use Qwen 3.8-27B as the LLM judge. When evaluating GPT-5.6 Luna as a judge candidate, we set its reasoning effort to low.

\section{Candidate Axis Proposal, Merging, and Annotation}
\label{app:annotation}

\paragraph{Human proposal.}
Three authors independently review all 24 trajectories for each of the 17 construction tasks and propose axes with a guide and viewer. To reduce anchoring, the guide instructs proposers to read all trajectories for a task before proposing axes and sets no target number of axes. The viewer displays agent and user messages, and the agent's tool calls, while hiding private user contexts and simulator state. Each proposed axis specifies a behavioral definition, categorical labels including \texttt{N/A}, and a decide-by rule.

\paragraph{LLM proposal.}
We use Claude Opus 5, GPT-5.6 Sol, and Grok 4.6 as LLM proposers. Three LLMs independently propose axes for 17 construction tasks, with all 24 trajectories for one task supplied in each call. The prompt instructs each model to review every trajectory before proposing axes and does not provide a target count or any example of an axis. To minimize bias, we assign random trajectory IDs and omit model, provider, and simulator metadata from the inputs.

\paragraph{Axis merging and human refinement.}
After deduplicating identical human rubrics, GPT-5.6 Sol merges the human and LLM proposals using their definitions, labels, decide-by rules, and task identifiers. The input excludes trial-label counts and explicit proposer-source fields. The merger records combined and discarded proposals. Three human reviewers examine all 29 preliminary axes, revising rubrics and merging or removing axes to obtain 23 behavioral axes. Appendix~\ref{app:inventory} reports proposal counts and the final inventory.

\paragraph{Annotation records.}
Each annotation records one rubric label and supporting evidence, indexed by trajectory, axis, judge, and repetition. Invalid labels, parse failures, timeouts, and truncated responses are logged as errors and retried. We annotate all 23 axes using the judges and input restrictions specified in Appendix~\ref{app:axis-evaluation-protocol}.

\FloatBarrier
\section{User Simulator}
\label{app:user-simulator}

\subsection{Persona Construction}
\label{app:persona}

To give personas realistic writing styles and feedback behaviors, we define style along grammatical fluency, politeness, and lexical diversity, following criteria informed by \citet{zhang2025mindgap}, and categorize feedback with a taxonomy adapted from \citet{donyehiya2025feedback}, covering clarification, request rephrasing, and corrective feedback with or without an explicit diagnosis. We measure the distributions of these attributes in user turns following substantive assistant responses in WildChat-1M~\citep{zhao2024wildchat} and sample each persona's style and feedback disposition from them. Task expertise and tolerance for imperfect outputs are assigned directly. Simulator prompts describe each attribute behaviorally, and a task's persona remains fixed across models and runs.

\subsection{Human Evaluation}
\label{app:simulator-validation}

\paragraph{Evaluation setup.}
Three annotators who are not authors of this paper independently evaluate 74 simulator utterances from the 17 construction trajectories used for judge validation (Appendix~\ref{app:human-validation}). Each annotator rates every utterance on three criteria, yielding 666 ratings in total.

\paragraph{Rating criteria and procedure.}
Annotators assess \emph{naturalness}, \emph{contextual appropriateness}, and \emph{role and self-consistency} on a five-point scale using the rubric in Table~\ref{tab:simulator-rating-rubric}. Judgments use only the conversation up to each utterance, without access to the hidden persona. Opening requests are assessed for standalone plausibility and internal coherence.

\begin{table}[htbp]
\centering
\footnotesize
\setlength{\tabcolsep}{4pt}
\renewcommand{\arraystretch}{1.12}
\caption{the five-point rubric used to evaluate simulator utterances. Each utterance receives one score for each criterion.}
\label{tab:simulator-rating-rubric}
\begin{tabularx}{\linewidth}{@{}c >{\raggedright\arraybackslash}X >{\raggedright\arraybackslash}X >{\raggedright\arraybackslash}X@{}}
\toprule
\textbf{Score} & \textbf{Naturalness} & \textbf{Contextual appropriateness} & \textbf{Role and self-consistency} \\
\midrule
1 & Wording is very unnatural or mechanical, making it difficult to regard as a real user's utterance.
  & Largely unrelated to the preceding question or deliverable, or clearly inconsistent with the conversation.
  & Clearly contradicts the user's previously stated role, goals, or facts. \\
\addlinespace[3pt]
2 & Expressions are noticeably awkward, repetitive, or formulaic.
  & Responds to only part of the preceding content, or the timing or direction of the response is substantially awkward.
  & Changes important goals, circumstances, or constraints without explanation. \\
\addlinespace[3pt]
3 & Understandable, but the tone or sentence flow is noticeably awkward.
  & Generally connected to the context, but misses an important part of the question or deliverable.
  & Could be the same user, but shows noticeable tension or ambiguity relative to earlier utterances. \\
\addlinespace[3pt]
4 & Sounds like a real user, with only minor awkwardness.
  & Responds appropriately to the preceding content, with only minor contextual mismatches.
  & Generally consistent with earlier information, with only minor tensions. \\
\addlinespace[3pt]
5 & Very natural wording for a real user in the given situation.
  & Responds accurately and naturally to the preceding question or deliverable.
  & Consistently maintains the user's role, goals, facts, and constraints. \\
\bottomrule
\end{tabularx}
\end{table}

\paragraph{Analysis.}
We report means and sample standard deviations over 222 ratings per criterion, along with multi-rater Gwet's AC1. We estimate 95\% confidence intervals using 5{,}000 trajectory-level bootstrap resamples, retaining all utterances and ratings within each trajectory.

\paragraph{Results.}
Mean ratings range from 4.55 to 4.68 out of 5, with AC1 values of 0.621--0.714 (Table~\ref{tab:simulator-validation}).

\begin{table}[htbp]
\centering
\small
\setlength{\tabcolsep}{5pt}
\caption{Human evaluation of 74 simulator utterances by three annotators. Means and sample standard deviations use 222 ratings per criterion. Gwet's AC1 measures exact agreement among all three annotators; its 95\% confidence intervals use 5{,}000 trajectory-level bootstrap resamples.}
\label{tab:simulator-validation}
\begin{tabular}{@{}lccc@{}}
\toprule
Criterion & Mean $\pm$ SD & Gwet's AC1 & 95\% CI \\
\midrule
Naturalness & $4.55 \pm 0.87$ & 0.621 & [0.513, 0.734] \\
Contextual appropriateness & $4.68 \pm 0.71$ & 0.714 & [0.616, 0.811] \\
Role and self-consistency & $4.57 \pm 0.83$ & 0.624 & [0.539, 0.712] \\
\bottomrule
\end{tabular}
\end{table}

\FloatBarrier

\FloatBarrier
\section{Taxonomy Evaluation Details}
  \label{app:axis-evaluation-protocol}

  \paragraph{Evaluation setup.}
We compare three taxonomies on 408 trajectories from eight models performing the 17 held-out tasks (Appendix~\ref{app:tasks}). Qwen3.8-27B, Gemma 4 31B, and GPT-5.6 Luna independently annotate each trajectory under each taxonomy, with three repetitions per judge. For \ourtax{}, judges assign labels with supporting evidence using each axis's definition, label descriptions, and decide-by rule. For Schwartz human values, judges rate alignment with each of ten values on a six-point scale. For Act\textperiodcentered{}ONOMY, judges tag agent passages with subactions; a subaction is present if at least one passage receives its tag in that repetition. The resulting representations contain 23 categorical axes, ten six-category values, and 46 binary subactions, respectively. Table~\ref{tab:taxonomy-comparison} reports their comparison.

\paragraph{Label diversity.}
For each judge and axis, we select each trajectory's modal label across three repetitions and compute normalized entropy:
  \[
  H_a=-\frac{1}{\log K_a}\sum_{k=1}^{K_a}p_{ak}\log p_{ak},
  \]
where $K_a$ is the number of permitted labels and $p_{ak}$ is the proportion of trajectories assigned label $k$. We set $0\log 0=0$, retain \texttt{N/A} where permitted, and resolve modal ties by lexicographic label order. Higher entropy indicates more even label use.

\paragraph{Judgment consistency.}
We measure judgment consistency using nominal-category Gwet's AC1. For each judge and axis, intra-judge AC1 is computed once after pooling all three repetition pairs from every trajectory. Inter-judge AC1 compares judges' modal labels and averages the three pairwise coefficients on each axis, using the same lexicographic tie rule.

\paragraph{Model-separation gap.}
We measure model separation on each axis using the difference between within-model and cross-model label agreement on the same held-out task. For axis $a$, judge $j$, task $t$, and annotation repetition $r$, let $W_{ajtr}$ be the fraction of unordered pairs of distinct trajectories from the same model that receive identical labels, and let $C_{ajtr}$ be the corresponding fraction for trajectories from different models. Each agreement rate weights its eligible trajectory pairs equally. We define
\begin{equation}
G_{ajtr}=W_{ajtr}-C_{ajtr}.
\label{eq:model-separation-gap}
\end{equation}
A positive gap indicates greater label agreement within models than across models on the same task. We compute the gap separately for each annotation repetition, then average equally over the three repetitions, eligible tasks, judges with defined estimates, and axes with defined estimates, in that order. Within-model and cross-model agreement use the same aggregation. This measure uses the existing held-out labels without fitting a classifier.

For each axis and judge, a task must retain at least 12 trajectories and at least two models with two or more trajectories each. The same retained trajectories are used in all three annotation repetitions. We exclude tasks whose labels are constant in every repetition, and omit an axis--judge estimate if no eligible task remains.

For \ourtax{}, we apply a conditional label-handling rule separately for each axis and judge. If the overall \texttt{N/A} rate across held-out trajectories and annotation repetitions is at least 20\% and the range of model-specific \texttt{N/A} rates is at least 30 percentage points, we restrict the gap calculation to trajectories receiving a non-\texttt{N/A} label in all three repetitions before applying the task eligibility rules. Otherwise, \texttt{N/A} remains a category. This condition applies only to \emph{Execution tool role} for all three judges. It does not change the labels used for entropy, AC1, or model matching.

Table~\ref{tab:taxonomy-gap} reports the resulting agreement rates and gaps. The gap is defined on 23 axes for \ourtax{}, ten for Schwartz values, and 36 for Act\textperiodcentered{}ONOMY. We obtain 95\% percentile intervals from 5{,}000 bootstrap resamples of axes with defined gaps within each taxonomy. These intervals describe variation across the evaluated axes; they are not confidence intervals for performance on a new task population.

\begin{table}[htbp]
\centering
\small
\caption{Model-separation gap on the held-out trajectories. Within and Cross denote same-model and different-model label agreement within a task, respectively. The Axes column gives the number of axes with defined gaps. Intervals summarize axis-bootstrap variation.}
\label{tab:taxonomy-gap}
\begin{tabular}{@{}lrrrrc@{}}
\toprule
Taxonomy & Axes & Within & Cross & Gap & 95\% interval \\
\midrule
\ourtax{} & 23 & 0.639 & 0.525 & 0.114 & $[0.080, 0.148]$ \\
Schwartz human values & 10 & 0.616 & 0.569 & 0.047 & $[0.030, 0.065]$ \\
Act\textperiodcentered{}ONOMY & 36 & 0.812 & 0.769 & 0.044 & $[0.029, 0.061]$ \\
\bottomrule
\end{tabular}
\end{table}

\paragraph{Model-matching accuracy.}
We measure model-matching accuracy~\citep{dunlap2025vibecheck} using the existing categorical trajectory labels. For each judge, we concatenate one-hot encodings of all axes and average the three annotation repetitions within each trajectory, treating all labels as nominal and retaining \texttt{N/A}. For each of the 28 model pairs, we form all $3\times3$ trajectory-repeat pairs within each task and fit a separate linear logistic classifier to their feature differences. We leave one entire task out, training on 16 tasks and evaluating on the remaining task, for 17 folds. All trajectories and reversed pairs from a task remain in the same fold. Classifier settings and statistical comparisons are specified in Appendix~\ref{app:stats}.

\paragraph{Aggregation and label handling.}
Entropy and intra-judge AC1 are averaged across judges within each axis, then equally across axes. Inter-judge AC1 is averaged across judge pairs within each axis, then equally across axes. Coefficients that are undefined (e.g., when an axis has only one observed label) are omitted from the averages. Model-matching accuracy gives ties half credit and averages equally over trajectory-repeat pairs, model pairs, tasks, and judges; the random baseline is 50\%. Entropy, AC1, and model-matching accuracy retain \texttt{N/A}; the gap applies the conditional label-handling rule described above. The full representations contain 116, 60, and 92 features for \ourtax{}, Schwartz, and Act\textperiodcentered{}ONOMY.

We compare inter-judge AC1 by independently bootstrapping the coordinates with defined coefficients within each taxonomy 5{,}000 times. The \ourtax{}--Act\textperiodcentered{}ONOMY difference is $-0.037$ (95\% interval $[-0.143, 0.073]$; two-sided bootstrap tail $=0.5199$). The \ourtax{}--Schwartz difference is $0.335$ ($[0.121, 0.540]$; $0.0032$), and the Act\textperiodcentered{}ONOMY--Schwartz difference is $0.372$ ($[0.161, 0.562]$; $0.0008$). Thus, both exceed Schwartz values, while \ourtax{} and Act\textperiodcentered{}ONOMY are not statistically distinguished.

\section{Human Validation and Judge Selection}
\label{app:human-validation}
 
\paragraph{Annotation setup and completeness.}
Three human annotators independently annotate 23 \ourtax{} axes on the same 17 trajectories, using the same trajectories and rubrics as the LLM judges. Each annotator annotates 391 labels, yielding 1{,}173 human labels in total. The three annotators agree unanimously on 185 of the 391 cells (47.3\%), two agree on 179 (45.8\%), and all three disagree on 27 (6.9\%). We use the 364 of 391 cells (93.1\%) that have a strict human majority for judge selection.

\textbf{Agreement measures.}
To assess judge–human agreement, we compare four models (Qwen3.8-27B, Gemma 4 31B, Muse-Glimmer-30B, and GPT-5.6 Luna) against human annotations. Each model labels 23 axes on each of the 17 tasks three times. We measure pooled observed agreement, Cohen's $\kappa$, and Gwet's AC1, plus Fleiss' $\kappa$ and nominal Krippendorff's $\alpha$ for agreement among the three annotators. Confidence intervals and judge selection both use a trajectory-level bootstrap with 2{,}000 paired resamples.
 
\textbf{Human inter-rater agreement.}
Pooled human observed agreement is 0.6257 (trajectory-bootstrap 95\% CI: 0.58--0.67), with Cohen's $\kappa=0.6096$, Gwet's AC1 $=0.6203$, Fleiss' $\kappa=0.6092$, and Krippendorff's $\alpha=0.6095$. Table~\ref{tab:human-pair-agreement} reports pairwise agreement. We use this level as a descriptive reference for judge–human agreement; it falls at the boundary between moderate and substantial agreement \citep{landis1977measurement}.
 
\begin{table}[htbp]
\centering
\small
\caption{Human pairwise agreement on 391 cells, with trajectory-bootstrap 95\% intervals.}
\label{tab:human-pair-agreement}
\begin{tabular}{@{}lrr@{}}
\toprule
Annotator pair & Agreement & 95\% CI \\
\midrule
H1--H2 & 0.6675 & [0.611, 0.729] \\
H1--H3 & 0.5882 & [0.537, 0.642] \\
H2--H3 & 0.6215 & [0.558, 0.680] \\
\bottomrule
\end{tabular}
\end{table}
 
\textbf{Direct human--judge agreement.}
Table~\ref{tab:human-judge-direct} compares individual human labels with judge labels. Across the 12 human--judge pairs, mean observed agreement is 0.6929, mean Cohen's $\kappa$ is 0.6809, and mean Gwet's AC1 is 0.6884. Qwen agrees most with H1 and H2, and Gemma with H3. Luna's agreement ranges from 0.5831 to 0.7954 across annotators.
 
\begin{table}[htbp]
\centering
\small
\caption{Agreement between each human and each judge's modal labels on all 391 cells.}
\label{tab:human-judge-direct}
\begin{tabular}{@{}lrrrr@{}}
\toprule
Annotator & Qwen & Gemma & Muse & GPT-5.6 Luna (low) \\
\midrule
H1 & 0.7008 & 0.6726 & 0.6368 & 0.6215 \\
H2 & 0.8235 & 0.8056 & 0.7570 & 0.7954 \\
H3 & 0.6496 & 0.6650 & 0.6036 & 0.5831 \\
\bottomrule
\end{tabular}
\end{table}
 
\textbf{Preregistered judge selection.}
To select the judge, we use pooled Gwet's AC1 against the strict human majority label. Judges are compared using percentile intervals from a paired trajectory-level bootstrap. If the leading judges cannot be distinguished, ties are broken first by mean axis-level AC1. Table~\ref{tab:human-majority-judges} reports agreement with the human majority, and Table~\ref{tab:judge-ac1-differences} reports the paired differences between judges.
 
\begin{table}[htbp]
\centering
\small
\setlength{\tabcolsep}{4pt}
\caption{Judge agreement with the human majority on 364 cells.}
\label{tab:human-majority-judges}
\begin{tabular}{@{}lrrrrr@{}}
\toprule
Judge & Agreement & Cohen's $\kappa$ & Pooled AC1 & AC1 95\% CI & Mean axis AC1 \\
\midrule
Qwen3.8-27B & 0.8352 & 0.8282 & 0.8327 & [0.785, 0.877] & 0.7878 \\
Gemma 4 31B & 0.8159 & 0.8080 & 0.8132 & [0.755, 0.864] & 0.7659 \\
Muse-Glimmer-30B & 0.7555 & 0.7450 & 0.7519 & [0.700, 0.801] & 0.6951 \\
GPT-5.6 Luna (low) & 0.7527 & 0.7435 & 0.7489 & [0.700, 0.794] & 0.6728 \\
\bottomrule
\end{tabular}
\end{table}
 
\begin{table}[htbp]
\centering
\small
\caption{Paired differences in pooled human-majority AC1, with 95\% intervals from 2{,}000 trajectory-bootstrap resamples.}
\label{tab:judge-ac1-differences}
\begin{tabular}{@{}lrr@{}}
\toprule
Comparison & AC1 difference & Paired 95\% CI \\
\midrule
Qwen $-$ Gemma & 0.0196 & [$-0.0164$, 0.0553] \\
Qwen $-$ Muse & 0.0808 & [0.0448, 0.1188] \\
Qwen $-$ Luna & 0.0838 & [0.0471, 0.1192] \\
Gemma $-$ Muse & 0.0612 & [0.0111, 0.1087] \\
Gemma $-$ Luna & 0.0643 & [0.0195, 0.1062] \\
Muse $-$ Luna & 0.0030 & [$-0.0333$, 0.0428] \\
\bottomrule
\end{tabular}
\end{table}
 
The Qwen--Gemma paired interval includes zero, so the tie-break selects Qwen by mean axis-level AC1 (0.7878 versus 0.7659). We use Qwen3.8-27B for the benchmark profiles in Section~\ref{sec:benchmark-setup}.

\section{Statistical Specification}
\label{app:stats}

\paragraph{Model-matching classifier.}
For each judge and model pair $(A,B)$, let $\boldsymbol{d}=\boldsymbol{x}_A-\boldsymbol{x}_B$ be the difference between trajectory features defined in Appendix~\ref{app:axis-evaluation-protocol}. Reversing the pair negates both the features and the class label. With equal weight on both orientations and no intercept, the logistic objective reduces to
\begin{equation}
\widehat{\boldsymbol{w}}_{-t}=\arg\min_{\boldsymbol{w}}\left\{
\frac{1}{|\mathcal{D}_{-t}|}\sum_{\boldsymbol{d}\in\mathcal{D}_{-t}}
\log\!\left(1+\exp(-\boldsymbol{w}^{\top}\boldsymbol{d})\right)
+\frac{\lambda}{2}\|\boldsymbol{w}\|_2^2\right\},
\end{equation}
where $\mathcal{D}_{-t}$ contains the trajectory-pair differences from the 16 training tasks and $\lambda=0.01$ for all taxonomies. The sign of $\widehat{\boldsymbol{w}}_{-t}^{\top}\boldsymbol{d}$ determines the assignment on task $t$; margins with absolute value at most $10^{-10}$ receive half credit. We use no feature scaling, feature selection, or hyperparameter tuning. The metric measures pairwise identification of the eight known models on unseen tasks.

\paragraph{Framework comparisons.}
We first average the dependent trajectory pairs, model pairs, and judges within each task, then compare the resulting 17 paired task accuracies. To approximately account for overlapping cross-validation training sets, we apply the variance correction of \citet{nadeau1999inference}. For task-level accuracy differences $\delta_t$, with mean $\bar{\delta}$ and sample variance $s_\delta^2$, the test statistic is
\begin{equation}
T=\frac{\bar{\delta}}{\sqrt{(1/17+1/16)s_\delta^2}}.
\end{equation}
We use a two-sided $t$ reference distribution with 16 degrees of freedom and Holm correction across the three taxonomy comparisons. \ourtax{} exceeds Schwartz values by 19.46 percentage points (adjusted $p=2.18\times10^{-6}$) and Act\textperiodcentered{}ONOMY by 5.05 points ($p=0.00236$); Act\textperiodcentered{}ONOMY exceeds Schwartz values by 14.41 points ($p=7.77\times10^{-5}$). These exploratory comparisons assume exchangeable tasks and hold the models, judges, axes, and classifier settings fixed. The cross-validation correction is approximate and does not guarantee exact Type I error control.

\paragraph{Annotation reliability.}
For each judge and axis, we pool all three repetition pairs from every trajectory and compute observed agreement $p_o$, Cohen's $\kappa$, and Gwet's AC1 once on the pooled pairs:
\begin{equation}
\kappa=\frac{p_o-p_e}{1-p_e},\quad p_e=\sum_{k}\pi^{(L)}_k\pi^{(R)}_k;\qquad
\mathrm{AC1}=\frac{p_o-p^{\gamma}_e}{1-p^{\gamma}_e},\quad p^{\gamma}_e=\frac{1}{K-1}\sum_{k}\pi_k(1-\pi_k),
\end{equation}
where $\pi^{(L)}_k$ and $\pi^{(R)}_k$ are label shares on the left and right sides of the pooled pairs, $\pi_k=(\pi^{(L)}_k+\pi^{(R)}_k)/2$, and $K$ is the number of observed labels, including \texttt{N/A} when present. Undefined coefficients ($K<2$ for AC1) are omitted. We base the reliability criterion on AC1 because $\kappa$ can be low under skewed label distributions even when the judge rarely disagrees with itself.

\section{Behavioral Axis Inventory}
\label{app:inventory}

Figures~\ref{fig:habit-inventory-1-4}--\ref{fig:habit-inventory-21-23} present the 23 \ourtax{} axes alphabetically. Each axis gives the definition, categorical labels, and decide-by rule; colors identify the five behavioral categories. \texttt{N/A} retains its axis-specific meaning.

The initial pool contains 716 proposals: 133 from humans and 583 from LLMs. Deduplicating human rubrics leaves 62 human proposals and 645 combined proposals. LLM merging reduces these to 29 preliminary axes, and human refinement yields the final 23 axes (Appendix~\ref{app:annotation}).

\newsavebox{\habitInventoryCaptionBox}
\newlength{\habitInventoryImageHeight}
\newcommand{\habitInventoryFigure}[3]{%
  \clearpage
  \begin{figure}[p]
  \sbox{\habitInventoryCaptionBox}{%
    \begin{minipage}{\textwidth}
    \caption{#2}#3
    \end{minipage}%
  }%
  \setlength{\habitInventoryImageHeight}{%
    \dimexpr\textheight-\ht\habitInventoryCaptionBox-\dp\habitInventoryCaptionBox-1pt\relax}%
  \vbox{%
    \offinterlineskip
    \hbox to\textwidth{\hfil
      \includegraphics[width=\textwidth,height=\habitInventoryImageHeight,keepaspectratio]{#1}%
      \hfil}%
    \copy\habitInventoryCaptionBox
  }%
  \end{figure}%
}

\habitInventoryFigure
  {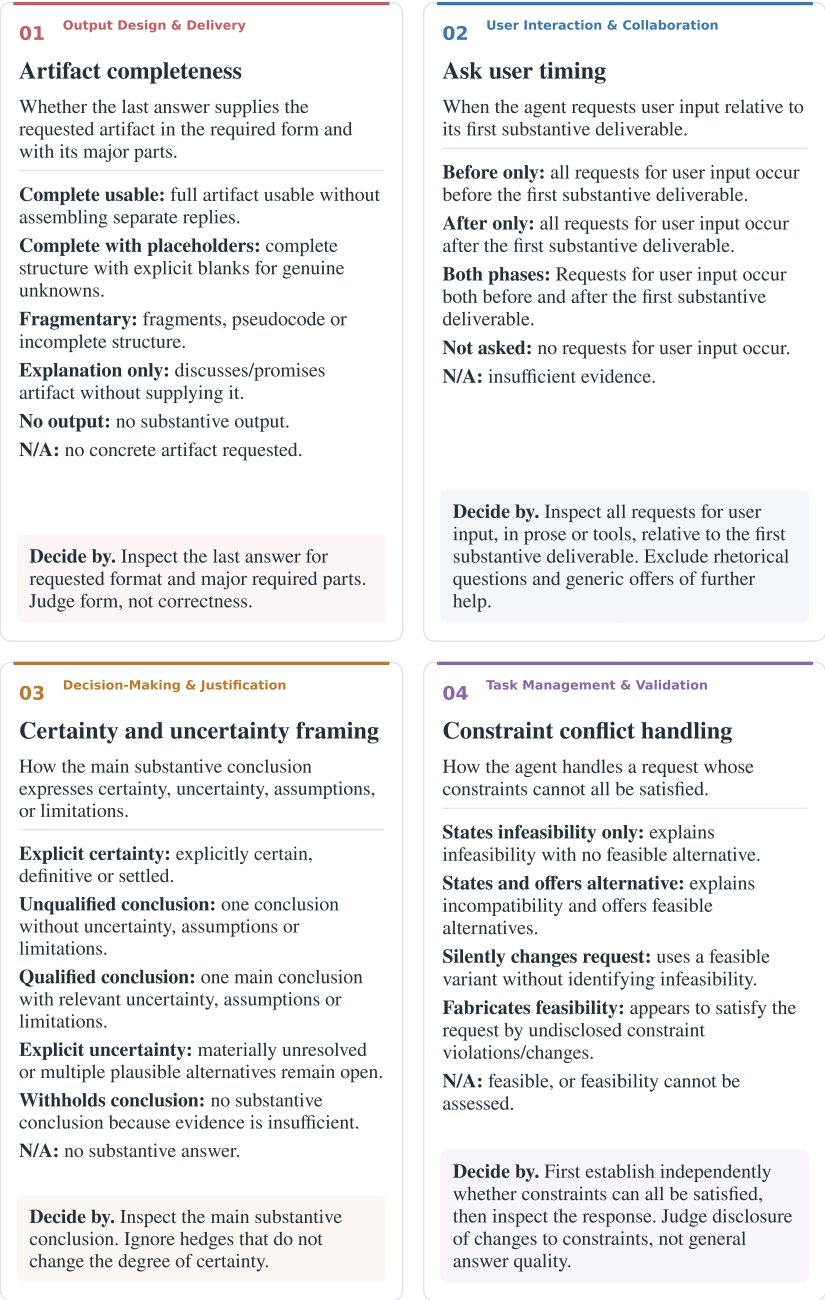}
  {\ourtax{} behavioral axis inventory: axes 1--4.}
  {\label{fig:habit-inventory-1-4}
   \label{app:axis-artifact-completeness}
   \label{app:axis-ask-user-timing}
   \label{app:axis-certainty-and-uncertainty-framing}
   \label{app:axis-constraint-conflict-handling}}

\habitInventoryFigure
  {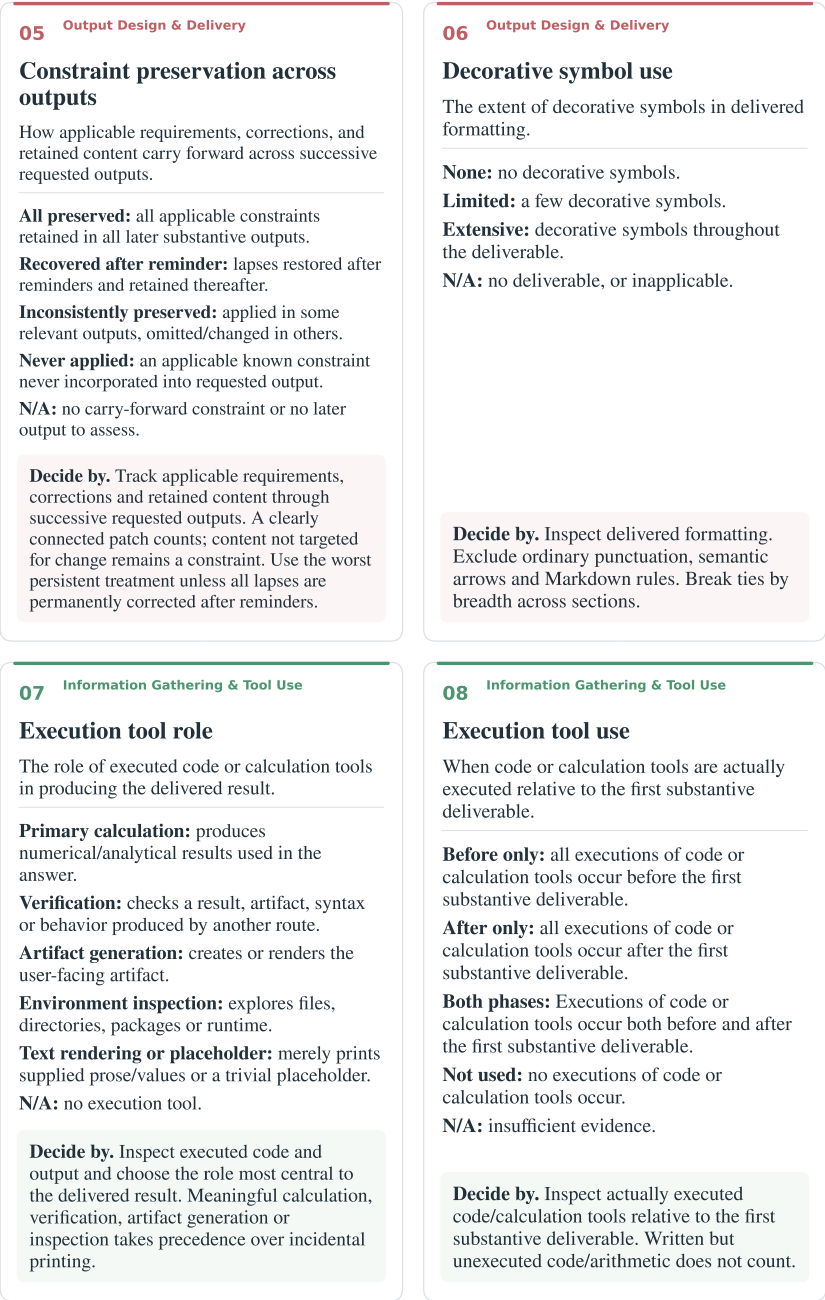}
  {\ourtax{} behavioral axis inventory: axes 5--8.}
  {\label{fig:habit-inventory-5-8}
   \label{app:axis-constraint-preservation-across-outputs}
   \label{app:axis-decorative-symbol-use}
   \label{app:axis-execution-tool-role}
   \label{app:axis-execution-tool-use}}

\habitInventoryFigure
  {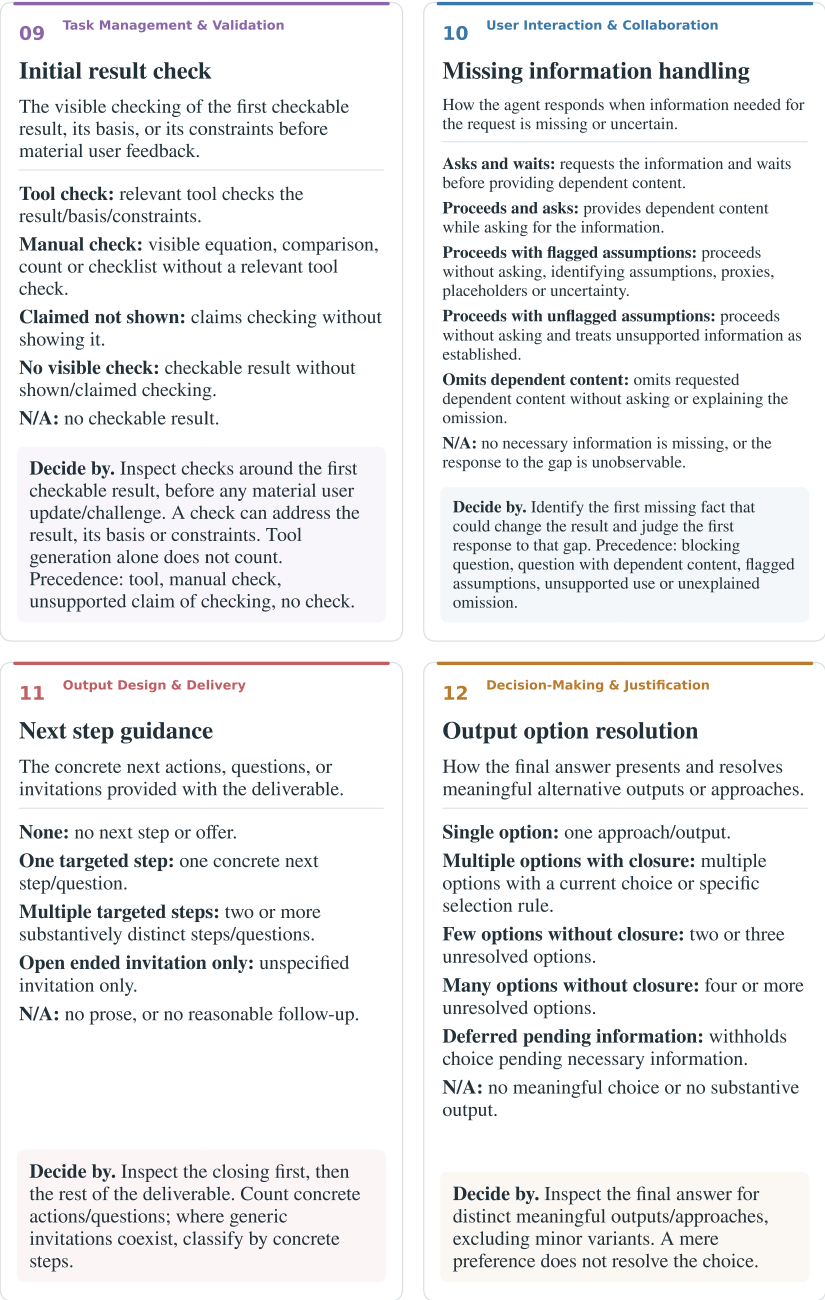}
  {\ourtax{} behavioral axis inventory: axes 9--12.}
  {\label{fig:habit-inventory-9-12}
   \label{app:axis-initial-result-check}
   \label{app:axis-missing-information-handling}
   \label{app:axis-next-step-guidance}
   \label{app:axis-output-option-resolution}}

\habitInventoryFigure
  {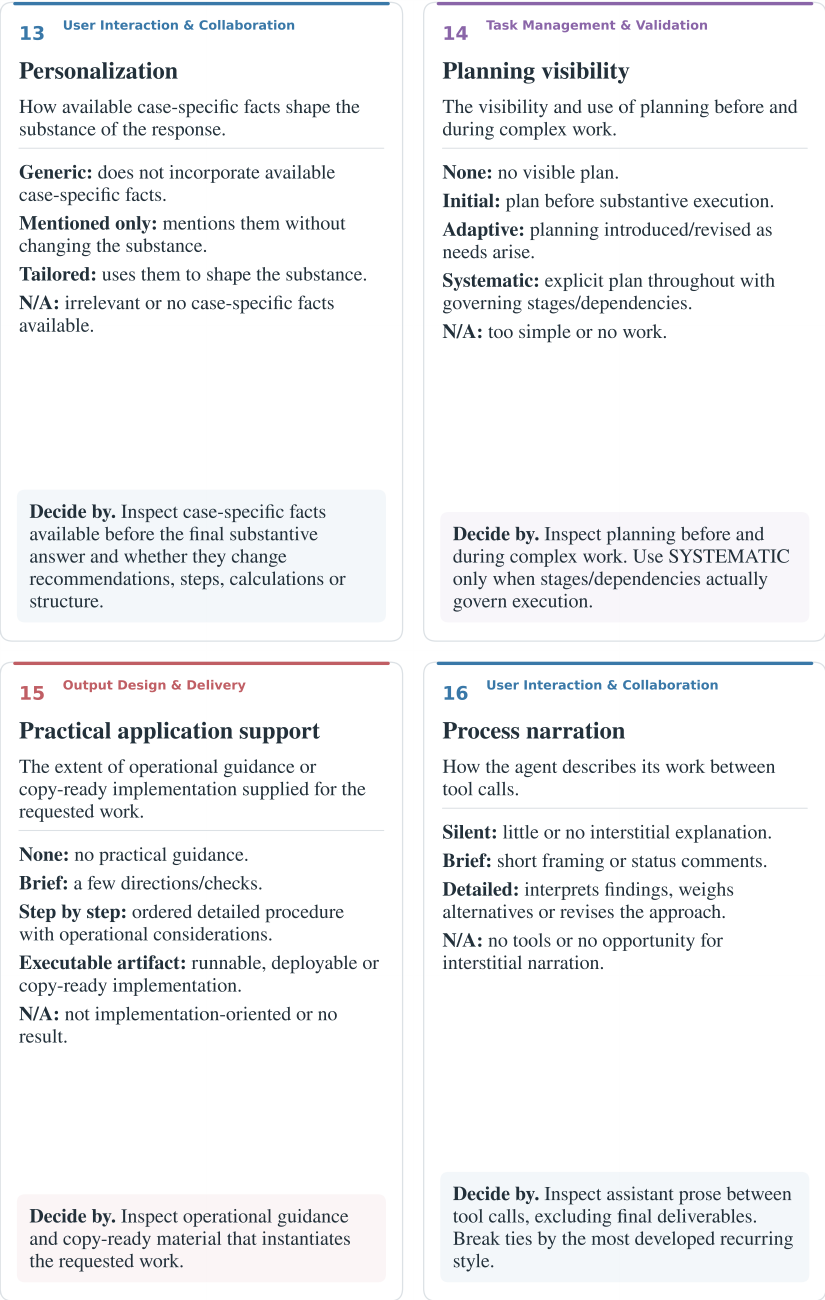}
  {\ourtax{} behavioral axis inventory: axes 13--16.}
  {\label{fig:habit-inventory-13-16}
   \label{app:axis-personalization}
   \label{app:axis-planning-visibility}
   \label{app:axis-practical-application-support}
   \label{app:axis-process-narration}}

\habitInventoryFigure
  {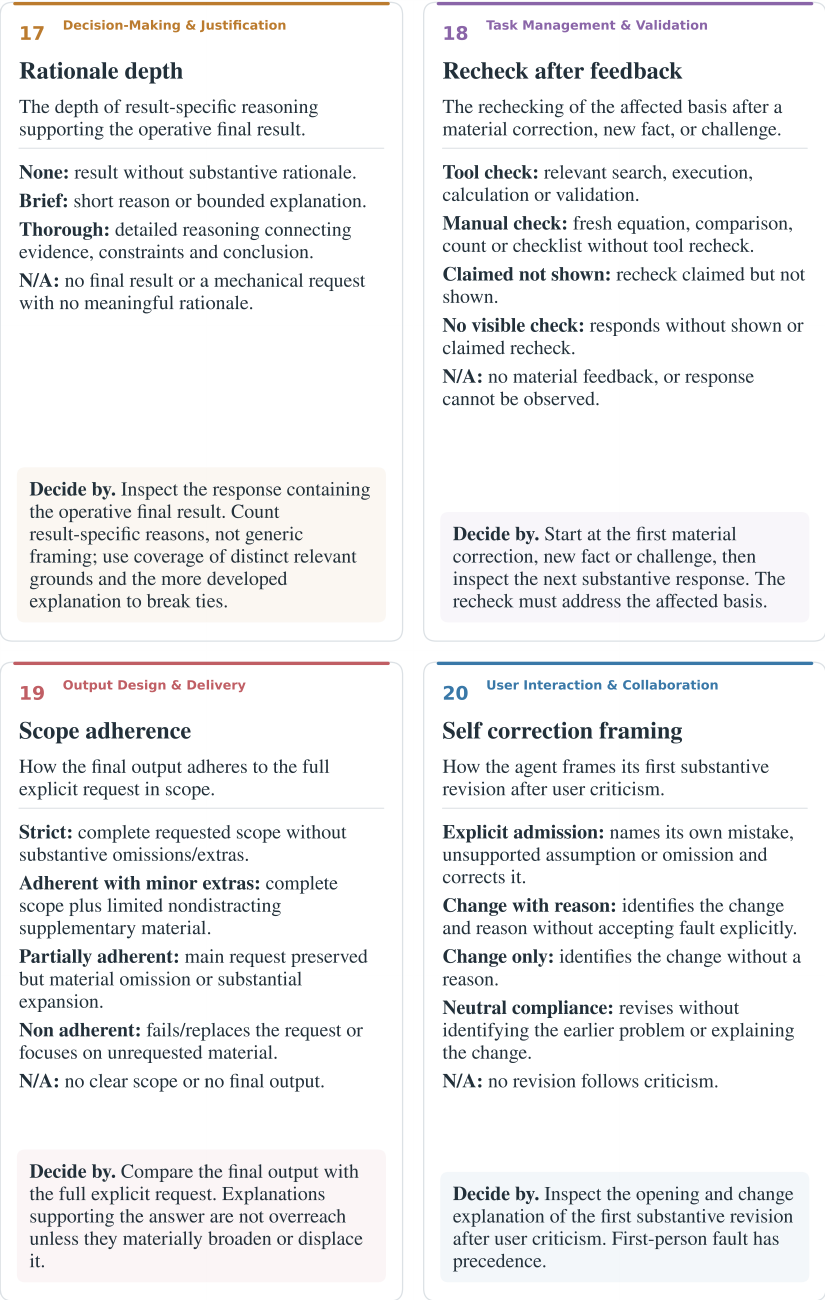}
  {\ourtax{} behavioral axis inventory: axes 17--20.}
  {\label{fig:habit-inventory-17-20}
   \label{app:axis-rationale-depth}
   \label{app:axis-recheck-after-feedback}
   \label{app:axis-scope-adherence}
   \label{app:axis-self-correction-framing}}

\habitInventoryFigure
  {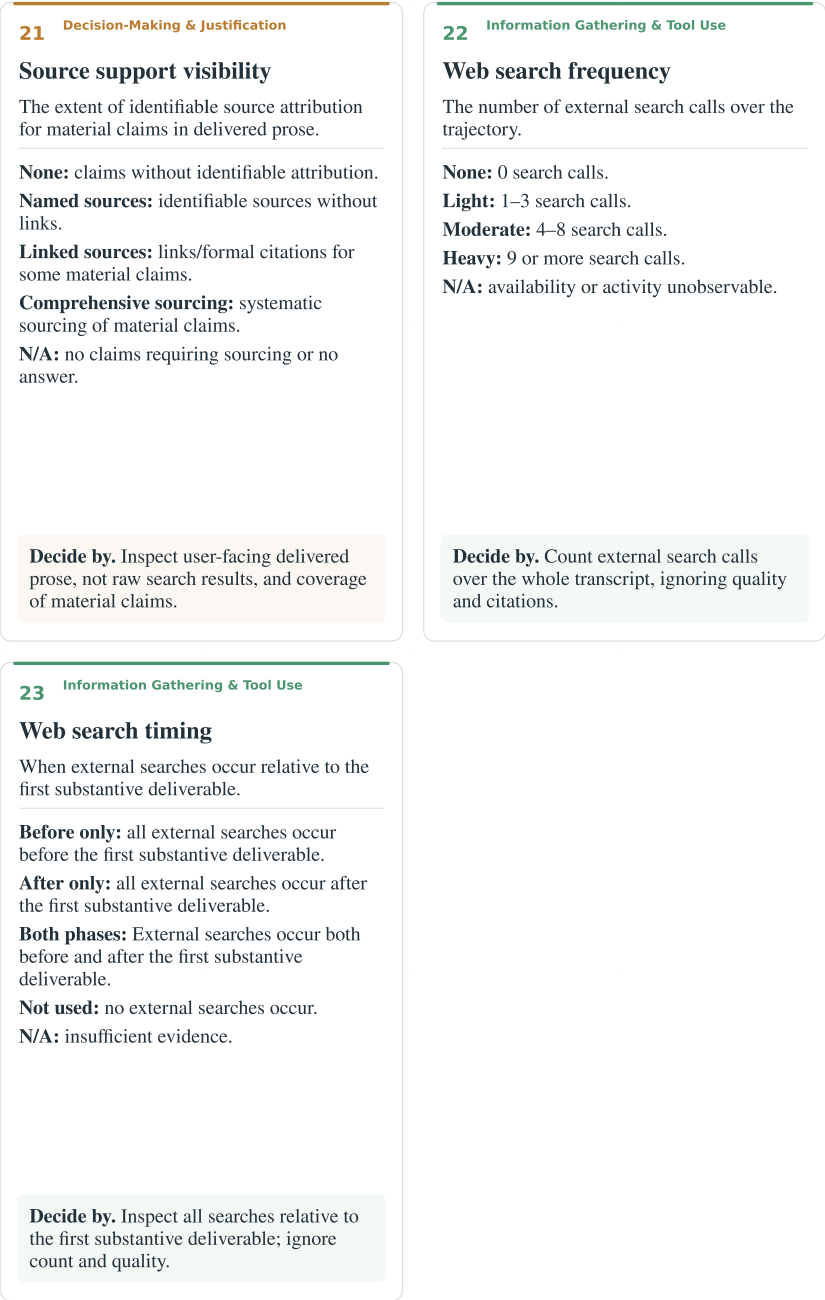}
  {\ourtax{} behavioral axis inventory: axes 21--23.}
  {\label{fig:habit-inventory-21-23}
   \label{app:axis-source-support-visibility}
   \label{app:axis-web-search-frequency}
   \label{app:axis-web-search-timing}}

\clearpage

\FloatBarrier
\section{Full Behavioral Profiles}
\label{app:full-profiles}

Figures~\ref{fig:appendix-profile-user-interaction}--\ref{fig:appendix-profile-output-delivery} show all 18 models on the 23 \ourtax{} axes.

\newsavebox{\appendixprofilecaptionbox}
\newlength{\appendixprofileimageheight}
\newcommand{\appendixprofilefigure}[3]{%
  \clearpage
  \begin{figure}[p]
  \sbox{\appendixprofilecaptionbox}{%
    \begin{minipage}{\textwidth}
    \caption{#2}\label{#3}
    \end{minipage}%
  }%
  \setlength{\appendixprofileimageheight}{%
    \dimexpr\textheight-\ht\appendixprofilecaptionbox-\dp\appendixprofilecaptionbox-1pt\relax}%
  \vbox{%
    \offinterlineskip
    \hbox to\textwidth{\hfil
      \includegraphics[width=\textwidth,height=\appendixprofileimageheight,keepaspectratio]{#1}%
      \hfil}%
    \copy\appendixprofilecaptionbox
  }%
  \end{figure}%
}

\appendixprofilefigure
{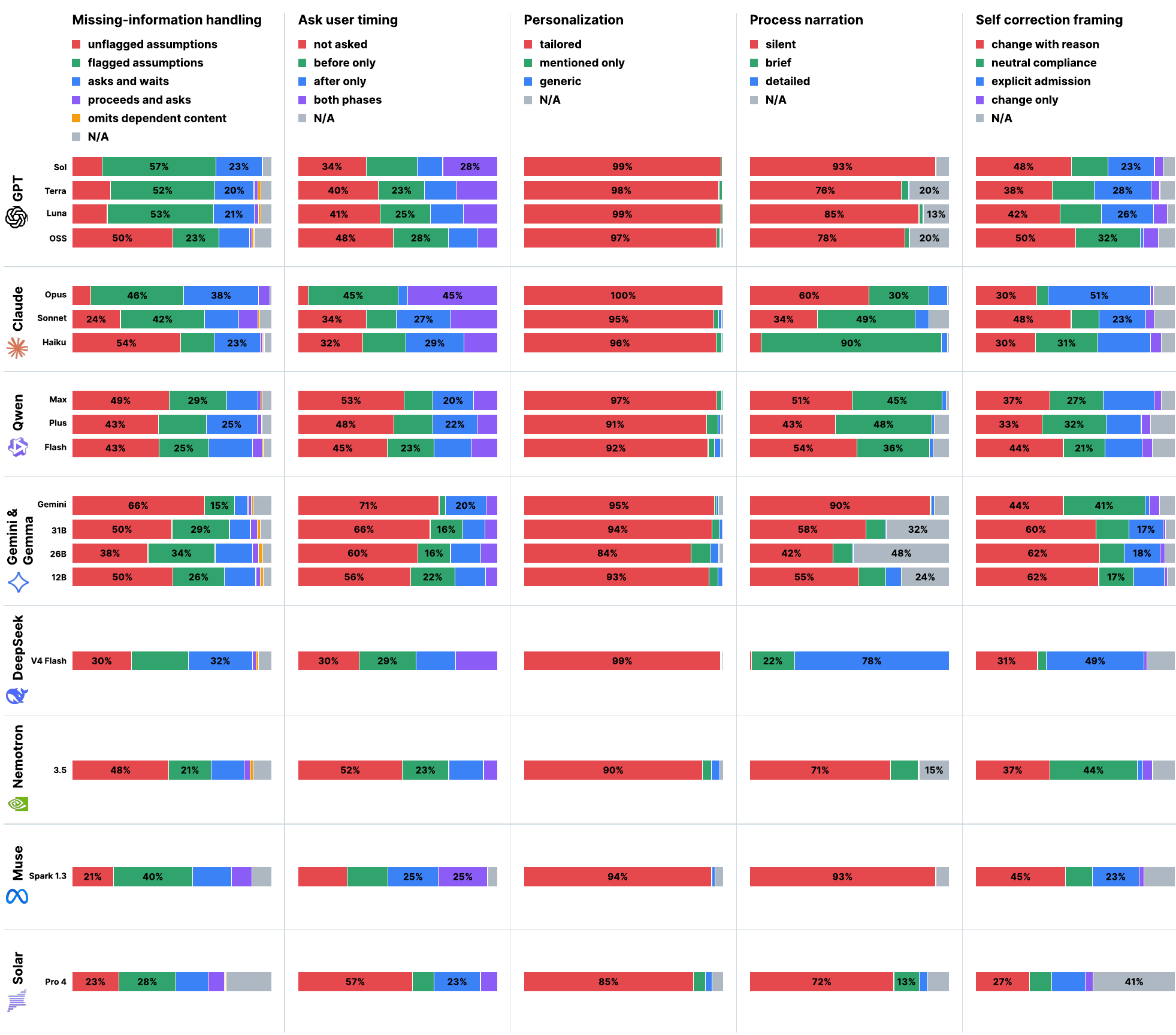}
{Profiles of 18 models: User Interaction \& Collaboration.}
{fig:appendix-profile-user-interaction}

\appendixprofilefigure
{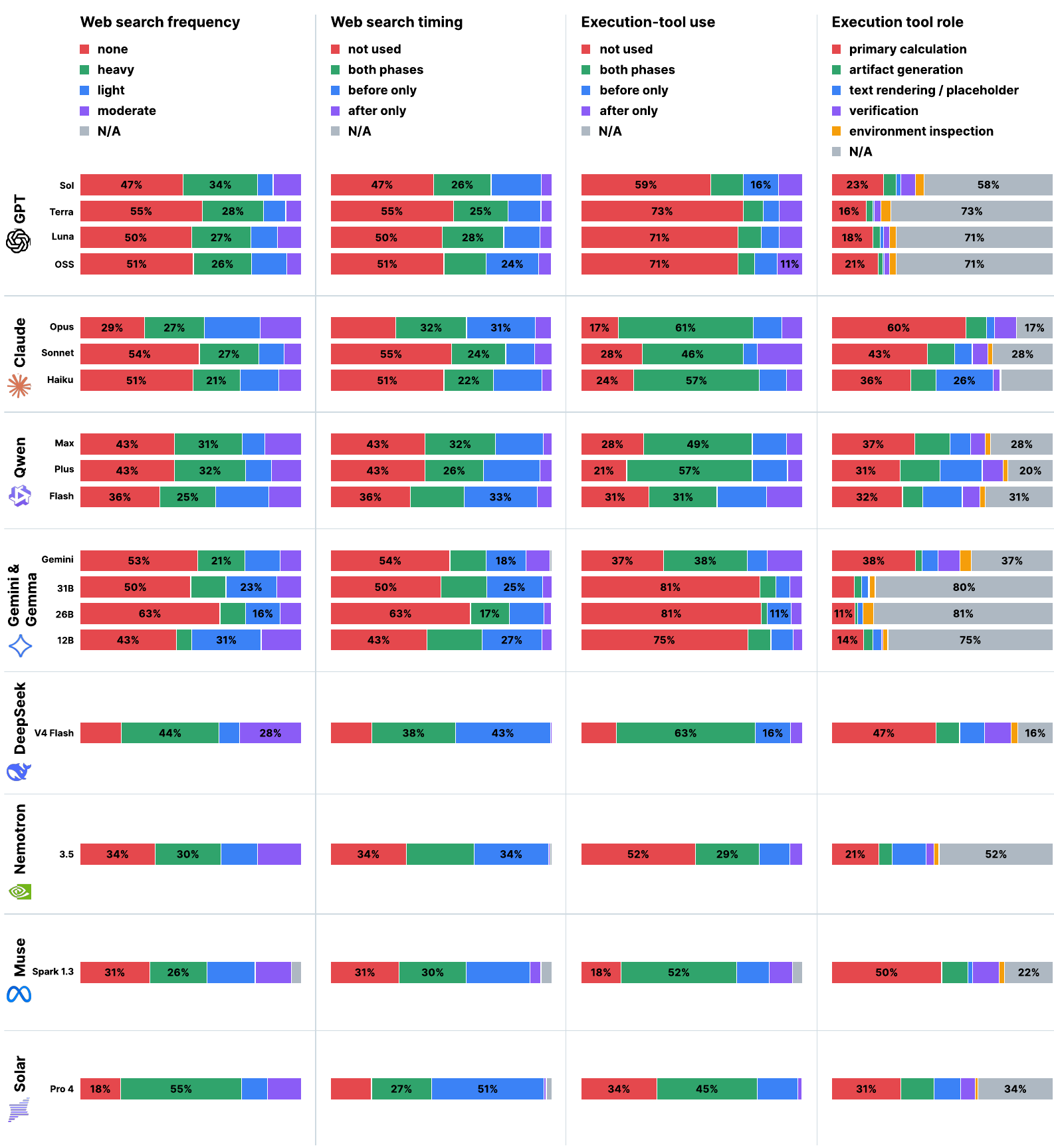}
{Profiles of 18 models: Information Gathering \& Tool Use.}
{fig:appendix-profile-information-gathering}

\appendixprofilefigure
{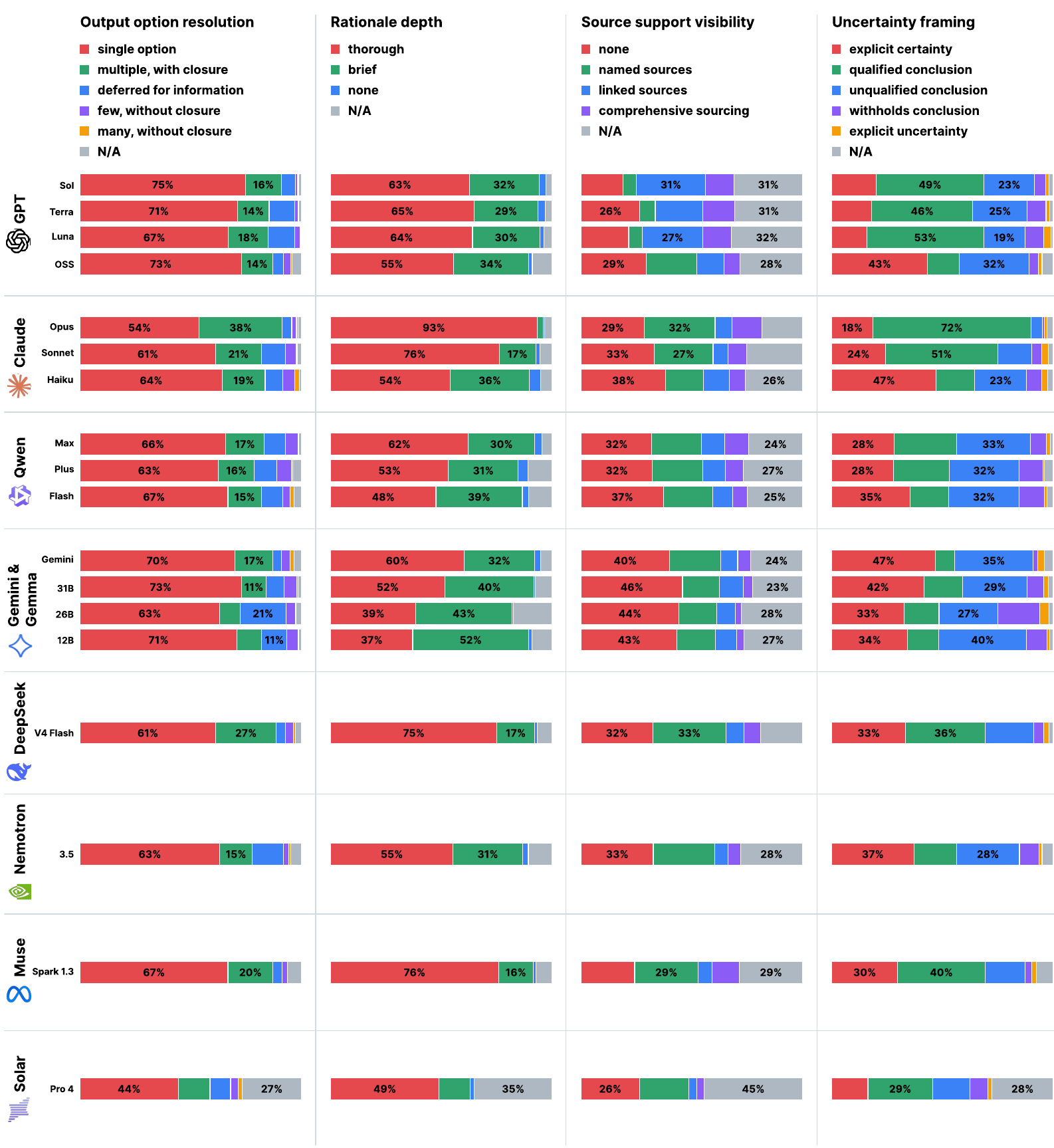}
{Profiles of 18 models: Decision-Making \& Justification.}
{fig:appendix-profile-decision-making}

\appendixprofilefigure
{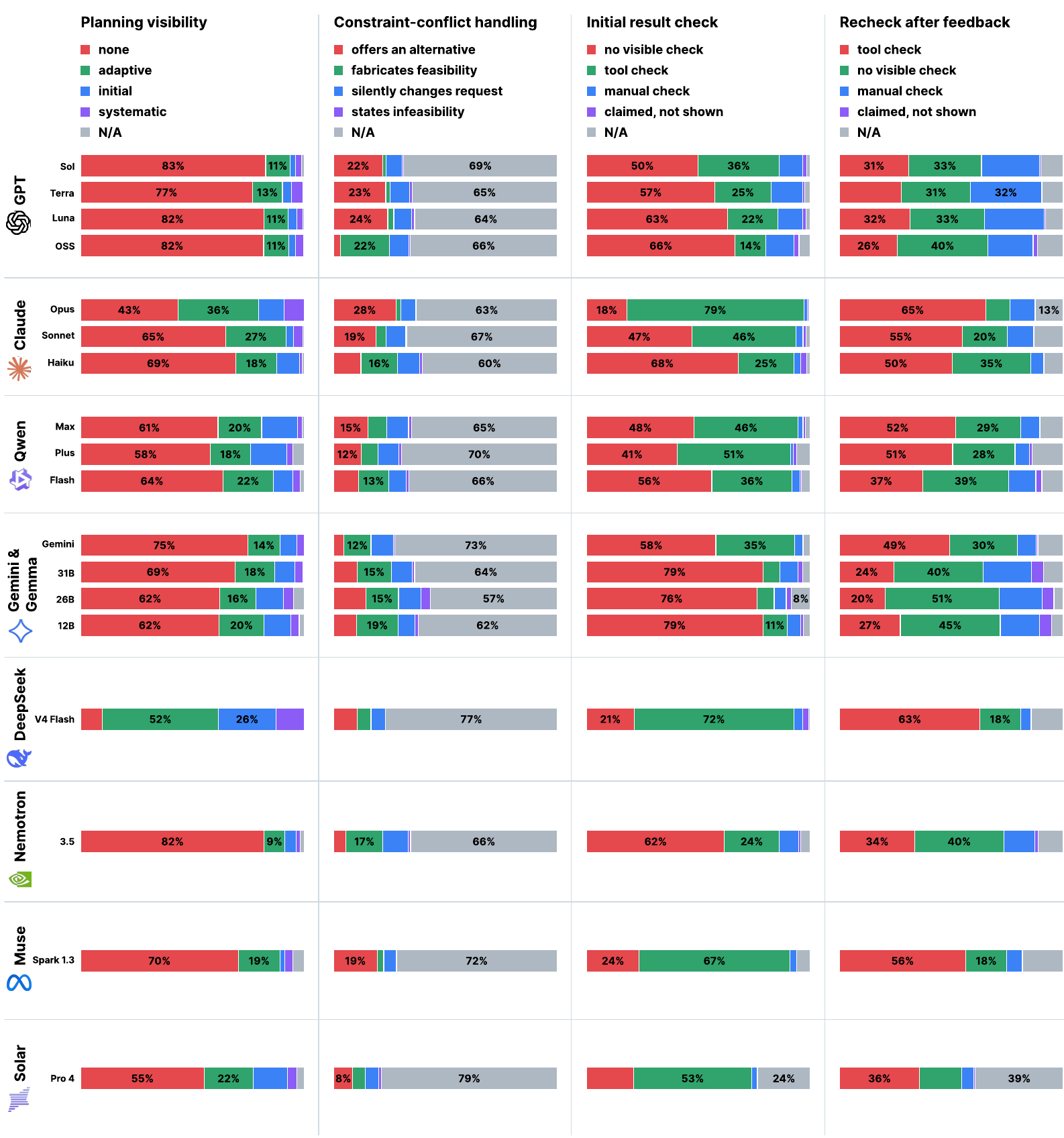}
{Profiles of 18 models: Task Management \& Validation.}
{fig:appendix-profile-task-management}

\appendixprofilefigure
{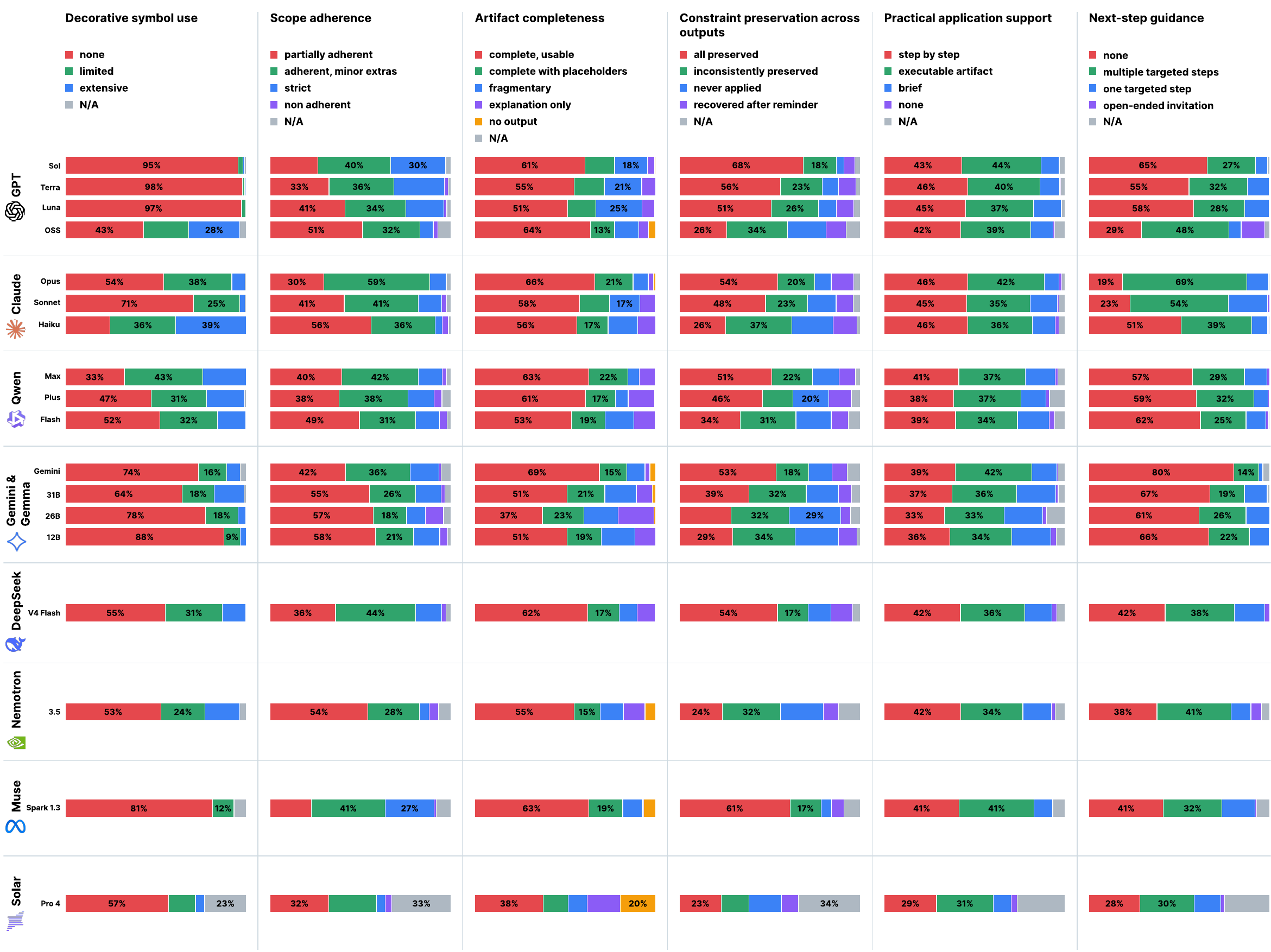}
{Profiles of 18 models: Output Design \& Delivery.}
{fig:appendix-profile-output-delivery}
\clearpage

\section{Statistical Analysis of Within-Family Profile Similarity}
\label{app:family-similarity-tests}

We test whether each family's mean profile similarity differs from the overall mean in Figure~\ref{fig:family-similarity}. The figure's 14-model roster contains 91 distinct pairs; each three-model family contributes three within-family pairs. We compare their mean with the mean across all 91 pairs, using profile similarity as defined in Section~\ref{sec:benchmark-results}. Each axis's distribution excludes \texttt{N/A} and is renormalized over the remaining labels.

We draw 10{,}000 paired cluster-bootstrap samples from the 17 domains. All tasks within a sampled domain receive the same multiplicity, applied to every model, retaining the dependence from shared models, tasks, agent runs, and judge repetitions. We report percentile intervals for within-family minus overall similarity. Two-sided bootstrap sign $p$-values use a plus-one finite-resampling correction, followed by Holm correction across the four families.

  \begin{table}[htbp]
  \centering
  \small
  \setlength{\tabcolsep}{4pt}
  \caption{Within-family similarity relative to all 91 model pairs. Differences have 95\% paired domain-bootstrap intervals; $p_{\mathrm{Holm}}$ adjusts four family comparisons.}
  \label{tab:family-similarity-tests}
  \begin{tabular}{@{}lrrrrr@{}}
  \toprule
  Family & Similarity & Difference & 95\% interval & $p$ & $p_{\mathrm{Holm}}$ \\
  \midrule
  GPT-5.6 & 0.933 &  0.131 & $[ 0.120,\  0.134]$ & 0.0002 & 0.0008 \\
  Qwen    & 0.908 &  0.105 & $[ 0.094,\  0.112]$ & 0.0002 & 0.0008 \\
  Gemma   & 0.913 &  0.111 & $[ 0.091,\  0.120]$ & 0.0002 & 0.0008 \\
  Claude  & 0.769 & $-0.033$ & $[-0.050,\ -0.010]$ & 0.0044 & 0.0044 \\
  \bottomrule
  \end{tabular}
  \end{table}

As a sensitivity check, we exclude each family's three pairs from its comparison baseline. Differences remain positive for GPT-5.6 (0.135; 95\% interval $[0.124, 0.139]$), Qwen (0.109; $[0.097, 0.116]$), and Gemma (0.115; $[0.094, 0.124]$), and negative for Claude ($-0.034$; $[-0.052, -0.011]$). This leaves the conclusions unchanged. Since models and family membership are not randomly sampled, these tests assess sensitivity to domain composition conditional on the observed roster and tasks, without identifying a population-level causal effect of model family.

\section{Profile Stability Across Task Subsets}
  \label{app:profile-stability}

Table~\ref{tab:full-profile-stability} extends the analysis in Section~\ref{sec:profile-stability} to all sampling ratios: 1\%, 10\%--90\% in 10-point increments, and 99\% of the 86 \ourbench{} tasks. Each ratio uses 1{,}000 repetitions. Model and family accuracy measure how often a query profile's nearest match belongs to the same model or family. We also report $D$, the mean TV distance across axes between the query and reference profiles of the same model.

The primary comparison uses reference profiles from the disjoint remaining tasks. A supplementary comparison uses profiles from all 86 tasks, including the query subset. The former measures stability across disjoint tasks; the latter measures convergence to the aggregate benchmark profile.

  \begin{table*}[t]
  \centering
  \small
  \caption{Profile identification accuracy and same-model distance $D$, averaged over 1{,}000 repetitions per subset size. References use either disjoint remaining tasks or all 86 tasks.}
  \label{tab:full-profile-stability}
  \resizebox{\textwidth}{!}{
  \begin{tabular}{rrccc@{\hspace{1.5em}}ccc}
  \toprule
  & & \multicolumn{3}{c}{Disjoint Remaining Tasks} &
  \multicolumn{3}{c}{Full Profile} \\
  \cmidrule(lr){3-5} \cmidrule(lr){6-8}
  Subset (\%) & Tasks &
  Model (\%) & Family (\%) & $D$ &
  Model (\%) & Family (\%) & $D$ \\
  \midrule
  1  & 1  & 42.18 & 63.84  & 0.394 & 45.63  & 65.93  & 0.389 \\
  10 & 9  & 80.21 & 98.90  & 0.146 & 90.37  & 99.66  & 0.131 \\
  20 & 17 & 86.34 & 99.90  & 0.112 & 96.91  & 100.00 & 0.090 \\
  30 & 26 & 89.31 & 99.99  & 0.097 & 99.14  & 100.00 & 0.068 \\
  40 & 34 & 89.51 & 100.00 & 0.092 & 99.74  & 100.00 & 0.056 \\
  50 & 43 & 89.86 & 100.00 & 0.089 & 99.98  & 100.00 & 0.045 \\
  60 & 52 & 88.77 & 99.99  & 0.091 & 100.00 & 100.00 & 0.036 \\
  70 & 60 & 87.36 & 100.00 & 0.097 & 100.00 & 100.00 & 0.029 \\
  80 & 69 & 82.53 & 99.81  & 0.113 & 100.00 & 100.00 & 0.022 \\
  90 & 77 & 73.78 & 98.36  & 0.146 & 100.00 & 100.00 & 0.015 \\
  99 & 85 & 31.07 & 58.02  & 0.398 & 100.00 & 100.00 & 0.005 \\
  \bottomrule
  \end{tabular}
  }
  \end{table*}

Identification against disjoint tasks peaks when query and reference sets have similar sizes and declines at extreme ratios, where one side contains few tasks. At 99\%, for example, the disjoint reference contains only one task. Identification against the full profile improves with coverage, consistent with convergence to the aggregate profile.

\section{Behavioral Self-Report Questionnaire}
\label{app:self-report}

  \subsection{Experimental Protocol}
  \label{app:self-report-protocol}

We construct one multiple-choice question (MCQ) per \ourtax{} axis from its fixed definition and non-\texttt{N/A} label descriptions. Each of 14 models answers all 23 questions in 10 separate calls per question, yielding 230 responses per model and 3{,}220 in total. The prompt asks about typical rather than ideal behavior and requires a JSON object containing only the selected option letter.

Option orders vary across repetitions and are shared across models. Axes with four or five options use 10 distinct orders. For the four three-option axes, we use all six permutations before reusing four once each. Selected letters are mapped back to their behavioral labels.

We compare each model's empirical MCQ distribution with its observed \ourbench{} profile after excluding \texttt{N/A} and renormalizing over the same labels. \emph{Mode Agreement} records whether their sets of modal labels overlap, allowing ties. \emph{Similarity} uses the measure in Section~\ref{sec:benchmark-results}, applied per axis. Axis-level summaries weight models equally; model-level summaries weight axes equally. Both measures assess correspondence with observed behavior rather than behavioral quality.

Motivated by prior work showing that generation-probability probes can complement questionnaire-based measures of LLM values~\citep{song2026humanpsychometricquestionnairesmischaracterize}, we also probe question-only continuation probabilities for four locally hosted open-weight models without displaying options (Appendix~\ref{app:self-report-logprobe}).

  \subsection{Axis-Level Results}
  \label{app:self-report-results}

Table~\ref{tab:self-report-axis} reports Similarity and modal agreement for every axis--model pair.

  \begingroup
  \tiny
  \setlength{\tabcolsep}{1.5pt}
  \renewcommand{\arraystretch}{1.15}
  
  \begin{longtable}{@{}>{\raggedright\arraybackslash}p{0.19\textwidth}*{14}{c}@{}}
  \caption{Self-report--behavior Similarity by axis and model. Bold indicates overlapping modal-label sets.}
  \label{tab:self-report-axis}\\
  \toprule
  & \multicolumn{3}{c}{GPT-5.6}
  & GPT-OSS
  & \multicolumn{3}{c}{Claude}
  & \multicolumn{3}{c}{Qwen 3.7}
  & Gemini
  & \multicolumn{3}{c}{Gemma 4} \\
  \cmidrule(lr){2-4}
  \cmidrule(lr){5-5}
  \cmidrule(lr){6-8}
  \cmidrule(lr){9-11}
  \cmidrule(lr){12-12}
  \cmidrule(l){13-15}
  Axis
  & Sol & Terra & Luna
  & 120B
  & Opus 5 & Sonnet 5 & Haiku 4.5
  & Plus & Max & Flash
  & 3.7 Flash
  & 31B & 26B-A4B & 12B \\
  \midrule
  \endfirsthead

  \multicolumn{15}{l}{\tablename~\thetable\ (continued)}\\
  \toprule
  & \multicolumn{3}{c}{GPT-5.6}
  & GPT-OSS
  & \multicolumn{3}{c}{Claude}
  & \multicolumn{3}{c}{Qwen 3.7}
  & Gemini
  & \multicolumn{3}{c}{Gemma 4} \\
  \cmidrule(lr){2-4}
  \cmidrule(lr){5-5}
  \cmidrule(lr){6-8}
  \cmidrule(lr){9-11}
  \cmidrule(lr){12-12}
  \cmidrule(l){13-15}
  Axis
  & Sol & Terra & Luna
  & 120B
  & Opus 5 & Sonnet 5 & Haiku 4.5
  & Plus & Max & Flash
  & 3.7 Flash
  & 31B & 26B-A4B & 12B \\
  \midrule
  \endhead

  \midrule
  \multicolumn{15}{r}{Continued on next page}\\
  \endfoot

  \bottomrule
  \endlastfoot

  Artifact Completeness
  & \textbf{0.609} & \textbf{0.549} & \textbf{0.513}
  & \textbf{0.738}
  & \textbf{0.664} & \textbf{0.580} & \textbf{0.736}
  & \textbf{0.611} & \textbf{0.631} & \textbf{0.534}
  & \textbf{0.689} & \textbf{0.510} & \textbf{0.373} & \textbf{0.509} \\

  Ask User Timing
  & 0.375 & 0.407 & 0.170
  & 0.101
  & \textbf{0.499} & 0.434 & 0.269
  & 0.103 & 0.221 & 0.132
  & 0.262 & 0.063 & 0.085 & 0.061 \\

  Certainty and Uncertainty Framing
  & \textbf{0.497} & \textbf{0.465} & \textbf{0.532}
  & 0.150
  & \textbf{0.736} & \textbf{0.521} & 0.179
  & 0.262 & 0.286 & 0.180
  & 0.092 & 0.180 & 0.161 & 0.142 \\

  Constraint Conflict Handling
  & \textbf{0.702} & \textbf{0.664} & \textbf{0.668}
  & 0.091
  & \textbf{0.752} & \textbf{0.579} & 0.307
  & \textbf{0.404} & \textbf{0.439} & 0.329
  & 0.166 & 0.289 & 0.332 & 0.271 \\

  Constraint Preservation Across Outputs
  & 0.289 & 0.337 & 0.271
  & \textbf{0.487}
  & 0.512 & 0.242 & \textbf{0.734}
  & 0.676 & \textbf{0.624} & \textbf{0.689}
  & \textbf{0.770} & 0.331 & \textbf{0.539} & \textbf{0.651} \\

  Decorative Symbol Use
  & \textbf{0.965} & \textbf{0.810} & \textbf{0.923}
  & \textbf{0.549}
  & \textbf{0.547} & \textbf{0.713} & 0.610
  & \textbf{0.479} & 0.326 & \textbf{0.525}
  & \textbf{0.761} & 0.380 & \textbf{0.877} & \textbf{0.876} \\

  Execution Tool Role
  & \textbf{0.811} & \textbf{0.869} & \textbf{0.824}
  & \textbf{0.791}
  & \textbf{0.619} & \textbf{0.595} & \textbf{0.509}
  & \textbf{0.614} & \textbf{0.523} & \textbf{0.575}
  & \textbf{0.602} & \textbf{0.520} & \textbf{0.562} & \textbf{0.566} \\

  Execution Tool Use
  & \textbf{0.585} & \textbf{0.806} & \textbf{0.686}
  & \textbf{0.577}
  & 0.431 & 0.339 & 0.709
  & 0.459 & 0.638 & \textbf{0.532}
  & \textbf{0.469} & 0.136 & 0.135 & 0.305 \\

  Initial Result Check
  & \textbf{0.507} & \textbf{0.589} & \textbf{0.742}
  & 0.355
  & \textbf{0.617} & \textbf{0.481} & 0.129
  & 0.446 & \textbf{0.491} & \textbf{0.582}
  & \textbf{0.959} & \textbf{0.816} & \textbf{0.915} & \textbf{0.812} \\

  Missing Information Handling
  & \textbf{0.598} & \textbf{0.575} & \textbf{0.584}
  & 0.169
  & \textbf{0.468} & \textbf{0.450} & 0.187
  & 0.352 & 0.418 & 0.461
  & 0.185 & 0.137 & 0.131 & 0.294 \\

  Next Step Guidance
  & \textbf{0.726} & \textbf{0.720} & 0.136
  & 0.271
  & 0.521 & 0.517 & 0.481
  & \textbf{0.682} & 0.214 & 0.409
  & 0.145 & 0.003 & 0.132 & 0.114 \\

  Output Option Resolution
  & \textbf{0.921} & \textbf{0.645} & 0.280
  & 0.146
  & 0.382 & \textbf{0.825} & 0.196
  & 0.568 & \textbf{0.842} & 0.556
  & 0.176 & \textbf{0.714} & \textbf{0.647} & \textbf{0.819} \\

  Personalization
  & \textbf{0.992} & \textbf{0.982} & \textbf{0.988}
  & \textbf{0.980}
  & \textbf{1.000} & \textbf{0.961} & \textbf{0.966}
  & \textbf{0.929} & \textbf{0.970} & \textbf{0.937}
  & \textbf{0.978} & \textbf{0.948} & \textbf{0.857} & \textbf{0.934} \\

  Planning Visibility
  & \textbf{0.837} & 0.477 & 0.108
  & 0.031
  & 0.114 & 0.306 & 0.102
  & 0.371 & \textbf{0.783} & \textbf{0.891}
  & 0.075 & 0.091 & 0.502 & 0.629 \\

  Practical Application Support
  & 0.543 & 0.210 & 0.456
  & \textbf{0.732}
  & 0.629 & 0.519 & \textbf{0.677}
  & \textbf{0.619} & 0.384 & 0.587
  & 0.810 & \textbf{0.690} & \textbf{0.837} & \textbf{0.582} \\

  Process Narration
  & 0.003 & 0.047 & 0.025
  & 0.031
  & 0.398 & \textbf{0.543} & 0.030
  & 0.313 & 0.121 & 0.520
  & 0.107 & 0.151 & 0.197 & 0.279 \\

  Rationale Depth
  & 0.424 & 0.496 & 0.315
  & \textbf{0.909}
  & 0.029 & \textbf{0.806} & \textbf{0.569}
  & \textbf{0.597} & \textbf{0.652} & \textbf{0.832}
  & \textbf{0.636} & \textbf{0.557} & 0.567 & \textbf{0.975} \\

  Recheck After Feedback
  & 0.287 & \textbf{0.649} & \textbf{0.462}
  & \textbf{0.457}
  & 0.427 & 0.433 & 0.065
  & 0.324 & 0.420 & \textbf{0.559}
  & \textbf{0.650} & \textbf{0.438} & \textbf{0.634} & \textbf{0.471} \\

  Scope Adherence
  & \textbf{0.413} & \textbf{0.365} & 0.345
  & 0.421
  & \textbf{0.601} & \textbf{0.420} & 0.360
  & 0.152 & 0.134 & 0.452
  & 0.170 & 0.265 & 0.299 & 0.213 \\

  Self-Correction Framing
  & 0.249 & 0.308 & 0.271
  & 0.017
  & \textbf{0.575} & 0.261 & 0.283
  & 0.201 & 0.274 & \textbf{0.709}
  & 0.025 & 0.178 & 0.192 & 0.160 \\

  Source Support Visibility
  & 0.375 & \textbf{0.482} & 0.516
  & 0.492
  & \textbf{0.491} & \textbf{0.798} & 0.232
  & \textbf{0.640} & 0.696 & \textbf{0.614}
  & \textbf{0.620} & \textbf{0.731} & 0.516 & \textbf{0.690} \\

  Web Search Frequency
  & \textbf{0.465} & \textbf{0.554} & \textbf{0.504}
  & \textbf{0.512}
  & \textbf{0.541} & \textbf{0.539} & \textbf{0.509}
  & \textbf{0.426} & \textbf{0.426} & \textbf{0.361}
  & 0.256 & \textbf{0.729} & 0.155 & \textbf{0.734} \\

  Web Search Timing
  & \textbf{0.666} & \textbf{0.554} & 0.642
  & \textbf{0.513}
  & 0.311 & \textbf{0.678} & \textbf{0.512}
  & \textbf{0.682} & 0.518 & 0.442
  & 0.264 & 0.206 & 0.275 & 0.349 \\

  \midrule
  Mean Similarity
  & 0.558 & 0.546 & 0.477
  & 0.414
  & 0.516 & 0.545 & 0.407
  & 0.474 & 0.480 & 0.539
  & 0.429 & 0.394 & 0.431 & 0.497 \\

  Mode Agreement (\%)
  & 65.2 & 69.6 & 47.8
  & 47.8
  & 56.5 & 65.2 & 34.8
  & 47.8 & 43.5 & 56.5
  & 43.5 & 43.5 & 39.1 & 52.2 \\

  \end{longtable}
  \endgroup

 \subsection{Common Self-Reported Behaviors}
  \label{app:self-report-label-frequencies}

Table~\ref{tab:self-report-label-frequencies} reports the most frequently selected MCQ label for each axis, pooled across 14 models and 10 responses per model. MCQ selection rate is the percentage of these 140 responses choosing that label. Observed frequency is the mean of the 14 models' \ourbench{} frequencies for the same label, excluding \texttt{N/A} and renormalizing within each model and axis.

  \begingroup
  \small
  \setlength{\tabcolsep}{3pt}
  \renewcommand{\arraystretch}{1.15}
  \begin{longtable}{@{}>{\raggedright\arraybackslash}p{0.27\textwidth}>{\raggedright\arraybackslash}p{0.37\textwidth}>{\raggedleft\arraybackslash}p{0.13\textwidth}>{\raggedleft\arraybackslash}p{0.15\textwidth}@{}}
  \caption{Most frequently self-reported label on each behavioral axis, with its MCQ selection rate and frequency in \ourbench{}. Both percentages weight models equally.}
  \label{tab:self-report-label-frequencies}\\
  \toprule
  Axis & Most selected label & MCQ selection (\%) & Observed frequency (\%) \\
  \midrule
  \endfirsthead
  \multicolumn{4}{l}{\tablename~\thetable\ (continued)}\\
  \toprule
  Axis & Most selected label & MCQ selection (\%) & Observed frequency (\%) \\
  \midrule
  \endhead
  \midrule
  \multicolumn{4}{r}{Continued on next page}\\
  \endfoot
  \bottomrule
  \endlastfoot

  Artifact Completeness & Complete usable & 97.9 & 57.0 \\
  Ask User Timing & Both phases & 91.4 & 15.9 \\
  Certainty and Uncertainty Framing & Qualified conclusion & 100.0 & 31.3 \\
  Constraint Conflict Handling & States and offers alternative & 100.0 & 42.8 \\
  Constraint Preservation Across Outputs & Inconsistently preserved & 67.9 & 27.4 \\
  Decorative Symbol Use & None & 87.9 & 66.2 \\
  Execution Tool Role & Primary calculation & 83.6 & 56.5 \\
  Execution Tool Use & Before only & 36.4 & 11.7 \\
  Initial Result Check & No visible check & 75.7 & 59.6 \\
  Missing Information Handling & Proceeds with flagged assumptions & 62.1 & 35.5 \\
  Next Step Guidance & One targeted step & 46.4 & 10.9 \\
  Output Option Resolution & Multiple options with closure & 55.0 & 17.3 \\
  Personalization & Tailored & 100.0 & 95.9 \\
  Planning Visibility & Initial & 56.4 & 8.8 \\
  Practical Application Support & Step by step & 42.1 & 43.3 \\
  Process Narration & Brief & 61.4 & 26.8 \\
  Rationale Depth & Thorough & 64.3 & 62.7 \\
  Recheck After Feedback & No visible check & 60.7 & 36.6 \\
  Scope Adherence & Adherent with minor extras & 69.3 & 36.1 \\
  Self-Correction Framing & Explicit admission & 96.4 & 22.9 \\
  Source Support Visibility & None & 57.9 & 45.4 \\
  Web Search Frequency & None & 77.9 & 47.7 \\
  Web Search Timing & Both phases & 42.1 & 24.2 \\

  \end{longtable}
  \endgroup

Table~\ref{tab:self-report-label-frequencies} shows that models report qualifying conclusions, explaining constraint conflicts with alternatives, and explicitly admitting mistakes much more frequently than these behaviors are observed in task trajectories. The first two labels are selected in every response on their respective axes, and explicit admission is selected in 96.4\% of responses, despite instructions to report typical rather than ideal behavior. Explicit admission is a self-reported mode for all 14 models (including one tie), but an observed mode for only one. These patterns are consistent with self-reports reflecting norms of desirable assistant behavior on some axes. However, frequent reports of inconsistent constraint preservation and no visible result check show that this pattern is not universal. Desirability is a post-hoc interpretation, and repeated choices of a typical behavior need not reproduce task-level frequencies; these comparisons alone do not establish a general desirability bias.

\subsection{Uniform-Choice Baseline and Cross-Model Comparison}
\label{app:self-report-baseline}

The uniform-choice baseline preserves each axis's response count and available labels. For model $m$ and axis $a$ with $K_a$ non-\texttt{N/A} labels, response counts and the resulting profile follow
\begin{equation}
\widetilde{\mathbf{C}}_{ma}
\sim \operatorname{Multinomial}
\left(10,\left(\frac{1}{K_a},\ldots,\frac{1}{K_a}\right)\right),
\qquad
\widetilde{\mathbf{q}}_{ma}=\frac{\widetilde{\mathbf{C}}_{ma}}{10}.
\end{equation}
We compare $\widetilde{\mathbf{q}}_{ma}$ with the observed profile $\mathbf{p}_{ma}$ using Mode Agreement and Similarity. With at most five labels per axis, we compute expectations exactly by enumerating all response-count vectors summing to 10, weighted by their multinomial probabilities. Equal averaging over 14 models and 23 axes yields 33.3\% Mode Agreement and 0.581 Similarity.

To assess model specificity, we compare each self-reported profile with the observed profiles of the other 13 models and average the resulting scores. Table~\ref{tab:self-report-cross-model} contrasts these mismatched comparisons with correspondence to the reporting model's own profile.

  \begin{table*}[t]
  \centering
  \caption{Self-report correspondence with the model's own behavior (Own) and the mean over 13 other models (Other). $\Delta$ is Own minus Other.}
  \label{tab:self-report-cross-model}
  \small
  \setlength{\tabcolsep}{5pt}
  \begin{tabular}{@{}lrrrrrr@{}}
  \toprule
  & \multicolumn{3}{c}{Mode Agreement (\%)}
  & \multicolumn{3}{c}{Similarity} \\
  \cmidrule(lr){2-4}\cmidrule(lr){5-7}
  Model & Own & Other & $\Delta$ & Own & Other & $\Delta$ \\
  \midrule
  GPT-5.6 Sol       & 65.2 & 60.9 & $+4.3$  & 0.558 & 0.504 & $+0.054$ \\
  GPT-5.6 Terra     & 69.6 & 51.8 & $+17.7$ & 0.546 & 0.499 & $+0.047$ \\
  GPT-5.6 Luna      & 47.8 & 43.1 & $+4.7$  & 0.477 & 0.446 & $+0.031$ \\
  GPT-OSS 120B      & 47.8 & 47.8 & $0.0$   & 0.414 & 0.451 & $-0.038$ \\
  Claude Opus 5     & 56.5 & 31.8 & $+24.7$ & 0.516 & 0.418 & $+0.098$ \\
  Claude Sonnet 5   & 65.2 & 54.8 & $+10.4$ & 0.545 & 0.500 & $+0.045$ \\
  Claude Haiku 4.5  & 34.8 & 46.2 & $-11.4$ & 0.407 & 0.437 & $-0.030$ \\
  Qwen 3.7 Max      & 43.5 & 52.5 & $-9.0$  & 0.480 & 0.494 & $-0.014$ \\
  Qwen 3.7 Plus     & 47.8 & 52.8 & $-5.0$  & 0.474 & 0.506 & $-0.032$ \\
  Qwen 3.7 Flash    & 56.5 & 54.8 & $+1.7$  & 0.539 & 0.561 & $-0.021$ \\
  Gemini 3.7 Flash  & 43.5 & 45.2 & $-1.7$  & 0.429 & 0.466 & $-0.037$ \\
  Gemma 4 31B       & 43.5 & 46.8 & $-3.3$  & 0.394 & 0.418 & $-0.024$ \\
  Gemma 4 26B-A4B   & 39.1 & 44.5 & $-5.4$  & 0.431 & 0.474 & $-0.043$ \\
  Gemma 4 12B       & 52.2 & 49.2 & $+3.0$  & 0.497 & 0.488 & $+0.009$ \\
  \midrule
  Mean              & 50.9 & 48.7 & $+2.2$  & 0.479 & 0.476 & $+0.003$ \\
  \bottomrule
  \end{tabular}
  \end{table*}

  \subsection{Question-Only Log-Probability Probe}
  \label{app:self-report-logprobe}

  \paragraph{Protocol.}
We probe GPT-OSS 120B, Gemma 4 31B, Gemma 4 26B-A4B, and Gemma 4 12B using only each axis-specific question, with the ordinary server-side chat template and generation prefix. Prompts omit option letters, label names, and competing descriptions. For each candidate label $\ell$, we teacher-force its full natural-language definition $d_\ell=(t_1,\ldots,t_{n_\ell})$ as the continuation and record token log-probabilities. This measures continuation likelihood without requiring an explicit option choice.

To account for definition length, the primary score is mean token log-probability,
  \begin{equation}
  s_\ell=\frac{1}{n_\ell}\sum_{j=1}^{n_\ell}\log P(t_j\mid q,t_{<j}),
  \end{equation}
where $q$ is the question and $n_\ell$ is the definition's token count. We convert scores into a relative distribution over each axis's non-\texttt{N/A} labels:
  \begin{equation}
  r_\ell=\frac{\exp(s_\ell)}{\sum_{k}\exp(s_k)}.
  \end{equation}
We compare $r$ with the behavioral and MCQ distributions using the measures in Appendix~\ref{app:self-report-protocol}. As a sensitivity check, we select labels by summed token log-probability and report their modal agreement with mean-score selections. Because summed scores favor shorter definitions, we do not interpret them as label distributions.

  \paragraph{Model-level results.}
Across the four models, the probe reaches 28.3\% Mode Agreement and 0.492 Similarity with observed behavior, and 33.7\% and 0.374 with MCQ profiles (Table~\ref{tab:self-report-logprobe-model}). The same models' MCQ profiles reach 45.7\% Mode Agreement and 0.434 Similarity with behavior. The probe is therefore closer to behavioral distributions by Similarity but matches their modes less often.

  \begin{table}[h]
  \centering
  \caption{Question-only probe correspondence, averaged over 23 axes.}
  \label{tab:self-report-logprobe-model}
  \small
  \setlength{\tabcolsep}{4pt}
  \begin{tabular}{@{}lccccc@{}}
  \toprule
  & \multicolumn{2}{c}{Probe--Behavior} & \multicolumn{2}{c}{Probe--MCQ} & Mean--Sum \\
  \cmidrule(lr){2-3}\cmidrule(lr){4-5}
  Model & Mode (\%) & Similarity & Mode (\%) & Similarity & Mode (\%) \\
  \midrule
  GPT-OSS 120B    & 30.4 & 0.511 & 43.5 & 0.466 & 13.0 \\
  Gemma 4 31B     & 30.4 & 0.491 & 30.4 & 0.355 & 39.1 \\
  Gemma 4 26B-A4B & 21.7 & 0.457 & 26.1 & 0.316 & 39.1 \\
  Gemma 4 12B     & 30.4 & 0.508 & 34.8 & 0.361 & 30.4 \\
  \midrule
  Mean             & 28.3 & 0.492 & 33.7 & 0.374 & 30.4 \\
  \bottomrule
  \end{tabular}
  \end{table}

Mean modal mass is 0.651 for the probe, 0.866 for MCQ responses, and 0.595 for observed behavior. The probe's lower concentration helps explain its higher distributional similarity despite lower modal agreement. Mean- and sum-score selections agree on only 30.4\% of axes, indicating sensitivity to definition length. Neither self-report method recovers the profiles measured from task trajectories.

  \paragraph{Axis-level results.}
Table~\ref{tab:self-report-logprobe-axis} averages each axis over the four models. Decorative symbol use reaches 100.0\% probe--behavior Mode Agreement and 0.791 Similarity, while several axes have no modal agreement despite moderate Similarity. Probe--MCQ modal agreement is highest for decorative symbol use, missing information handling, rationale depth, recheck after feedback, and web search timing.

  \begingroup
  \small
  \setlength{\tabcolsep}{4pt}
  \renewcommand{\arraystretch}{1.1}
  \begin{longtable}{@{}>{\raggedright\arraybackslash}p{0.42\textwidth}cccc@{}}
  \caption{Question-only probe correspondence by axis, averaged equally over four models. Mode columns report Mode Agreement (\%).}
  \label{tab:self-report-logprobe-axis}\\
  \toprule
  & \multicolumn{2}{c}{Probe--Behavior} & \multicolumn{2}{c}{Probe--MCQ} \\
  \cmidrule(lr){2-3}\cmidrule(l){4-5}
  Axis & Mode (\%) & Similarity & Mode (\%) & Similarity \\
  \midrule
  \endfirsthead
  \multicolumn{5}{l}{\tablename~\thetable\ (continued)}\\
  \toprule
  & \multicolumn{2}{c}{Probe--Behavior} & \multicolumn{2}{c}{Probe--MCQ} \\
  \cmidrule(lr){2-3}\cmidrule(l){4-5}
  Axis & Mode (\%) & Similarity & Mode (\%) & Similarity \\
  \midrule
  \endhead
  \midrule
  \multicolumn{5}{r}{Continued on next page}\\
  \endfoot
  \bottomrule
  \endlastfoot
  Artifact Completeness & 25.0 & 0.621 & 25.0 & 0.351 \\
  Ask User Timing & 100.0 & 0.668 & 0.0 & 0.014 \\
  Certainty and Uncertainty Framing & 25.0 & 0.564 & 25.0 & 0.272 \\
  Constraint Conflict Handling & 0.0 & 0.406 & 0.0 & 0.090 \\
  Constraint Preservation Across Outputs & 0.0 & 0.373 & 25.0 & 0.277 \\
  Decorative Symbol Use & 100.0 & 0.791 & 75.0 & 0.731 \\
  Execution Tool Role & 50.0 & 0.583 & 50.0 & 0.425 \\
  Execution Tool Use & 0.0 & 0.293 & 25.0 & 0.596 \\
  Initial Result Check & 25.0 & 0.356 & 25.0 & 0.224 \\
  Missing Information Handling & 0.0 & 0.233 & 75.0 & 0.513 \\
  Next Step Guidance & 25.0 & 0.697 & 0.0 & 0.231 \\
  Output Option Resolution & 25.0 & 0.350 & 50.0 & 0.507 \\
  Personalization & 25.0 & 0.388 & 25.0 & 0.317 \\
  Planning Visibility & 50.0 & 0.650 & 25.0 & 0.412 \\
  Practical Application Support & 0.0 & 0.515 & 0.0 & 0.446 \\
  Process Narration & 25.0 & 0.286 & 25.0 & 0.334 \\
  Rationale Depth & 75.0 & 0.704 & 75.0 & 0.692 \\
  Recheck After Feedback & 75.0 & 0.653 & 75.0 & 0.573 \\
  Scope Adherence & 0.0 & 0.453 & 50.0 & 0.318 \\
  Self-Correction Framing & 0.0 & 0.385 & 50.0 & 0.354 \\
  Source Support Visibility & 0.0 & 0.252 & 0.0 & 0.177 \\
  Web Search Frequency & 25.0 & 0.567 & 0.0 & 0.315 \\
  Web Search Timing & 0.0 & 0.524 & 75.0 & 0.437 \\
  \midrule
  Mean & 28.3 & 0.492 & 33.7 & 0.374 \\
  \end{longtable}
  \endgroup

  \subsection{Questionnaire}
  \label{app:self-report-questionnaire}

Every call uses the following system prompt:
  \begin{quote}
  \small
  You are completing a questionnaire about your own typical behavior as an assistant. Report what you usually do across diverse open-ended agent tasks, not what an ideal assistant should do and not what this questionnaire appears to prefer. Choose exactly one listed option. Return only the requested JSON object.
  \end{quote}

The user-message template is provided in Appendix~\ref{app:prompts}.

Table~\ref{tab:self-report-questionnaire} reproduces the question suffixes and options in the first repetition's order. Later repetitions permute the options as specified in Appendix~\ref{app:self-report-protocol}. Axis names are reference metadata and are hidden from the model.

  \begingroup
  \small
  \setlength{\tabcolsep}{4pt}
  \renewcommand{\arraystretch}{1.12}

  \begin{longtable}{@{}>{\raggedright\arraybackslash}p{0.25\textwidth}>{\raggedright\arraybackslash}p{0.69\textwidth}@{}}
  \caption{Behavioral self-report questions and options. Each call presents one axis's question.}
  \label{tab:self-report-questionnaire}\\
  \toprule
  Axis & Question suffix and options \\
  \midrule
  \endfirsthead
  \multicolumn{2}{l}{\tablename~\thetable\ (continued)}\\
  \toprule
  Axis & Question suffix and options \\
  \midrule
  \endhead
  \midrule
  \multicolumn{2}{r}{Continued on next page}\\
  \endfoot
  \bottomrule
  \endlastfoot

  Artifact Completeness & \emph{whether the assistant supplies the requested concrete artifact with its major required parts?}\newline A. Provides the complete structure but leaves explicit blanks for genuinely unknown values.\newline B. Discusses or promises the artifact without supplying it.\newline C. Produces no substantive assistant output.\newline D. Provides a complete artifact that can be used without assembling material from separate replies.\newline E. Provides only fragments, pseudocode, or an incomplete structure. \\
  \addlinespace

  Ask User Timing & \emph{whether and when the assistant asks the user for information, preferences, or clarification?}\newline A. Asks only after the first substantive deliverable.\newline B. Asks before the first substantive deliverable, but not afterward.\newline C. Does not ask the user for information, preferences, or clarification.\newline D. Asks both before and after the first substantive deliverable. \\
  \addlinespace

  Certainty and Uncertainty Framing & \emph{how certain or uncertain the assistant presents a substantive answer?}\newline A. Provides no substantive conclusion because the available evidence is insufficient.\newline B. Explicitly describes one conclusion as certain, definitive, or settled.\newline C. Presents one conclusion without expressing certainty, uncertainty, assumptions, or limitations.\newline D. Leaves the conclusion materially unresolved or keeps multiple plausible alternatives open.\newline E. Presents one conclusion as most likely while acknowledging relevant uncertainty, assumptions, or limitations. \\
  \addlinespace

  Constraint Conflict Handling & \emph{how the assistant responds when the requested constraints cannot all be satisfied?}\newline A. Moves to a feasible variant without clearly stating that the original request is infeasible.\newline B. Explains that the request cannot be completed as stated and offers no feasible alternative.\newline C. Produces an output that appears to satisfy the request by violating or changing constraints without disclosure.\newline D. Explains the incompatibility and offers one or more feasible alternatives. \\
  \addlinespace

  Constraint Preservation Across Outputs & \emph{how consistently the assistant preserves and applies user requirements, corrections, and content meant to remain unchanged across successive outputs?}\newline A. One or more constraints are dropped, but the assistant restores them after a user reminder and preserves them in all subsequent outputs.\newline B. A constraint is applied in some relevant outputs but omitted or altered in others. This includes addressing only the latest issue while losing earlier constraints.\newline C. At least one applicable constraint is known but never incorporated into a requested output.\newline D. Every later substantive output preserves all applicable constraints, either directly or through a clearly connected patch. \\
  \addlinespace

  Decorative Symbol Use & \emph{how much the assistant uses emoji or ornamental glyphs in delivered formatting?}\newline A. Uses decorative symbols throughout headings or structure.\newline B. Uses no emoji or ornamental glyphs.\newline C. Uses a small number of decorative symbols. \\
  \addlinespace

  Execution Tool Role & \emph{the primary role played by code execution relative to the requested work?}\newline A. Execution primarily creates or renders a user-facing artifact.\newline B. Execution primarily explores files, directories, packages, or the runtime environment.\newline C. Execution produces numerical or analytical results used in the answer.\newline D. Execution merely prints prose, supplied values, or a trivial placeholder.\newline E. Execution checks a result, artifact, syntax, or behavior produced by another route. \\
  \addlinespace

  Execution Tool Use & \emph{whether and when the assistant uses a code-execution or calculation tool relative to the first substantive deliverable?}\newline A. Uses an execution tool both before and after the first substantive deliverable.\newline B. Uses an execution tool before the first substantive deliverable, but not afterward.\newline C. Uses an execution tool only after the first substantive deliverable.\newline D. Does not use an execution tool. \\
  \addlinespace

  Initial Result Check & \emph{whether and how the assistant visibly checks its first checkable result before or after presenting it?}\newline A. Uses a tool to check the result, its underlying basis, or relevant constraints.\newline B. Claims to have checked the result without showing the check.\newline C. Shows an equation, comparison, count, or checklist that checks the result, without a relevant tool-based check.\newline D. Presents a checkable result without showing or claiming a check. \\
  \addlinespace

  Missing Information Handling & \emph{how the assistant responds when information needed to answer the user's request is missing or uncertain?}\newline A. Continues without asking and clearly identifies the assumptions, proxies, placeholders, or uncertain information it uses.\newline B. Provides an answer that depends on the missing information while also asking the user to supply it.\newline C. Continues without asking and treats unsupported values or conditions as established facts.\newline D. Requests the missing information and waits for the user's reply before providing an answer that depends on it.\newline E. Leaves out a requested part that depends on the missing information, without asking for that information or explaining the omission. \\
  \addlinespace

  Next Step Guidance & \emph{whether the assistant proposes concrete next steps, further work, or targeted follow-up questions after the current answer?}\newline A. Provides two or more substantively different next steps or questions.\newline B. Provides no next step or follow-up offer.\newline C. Provides one concrete next step or targeted question.\newline D. Invites unspecified further input without identifying a concrete next step. \\
  \addlinespace

  Output Option Resolution & \emph{how the assistant presents alternative outputs or approaches and whether it resolves them to a current choice?}\newline A. Withholds a current choice until necessary information is obtained.\newline B. Presents multiple options and identifies one current choice or provides a specific rule for selecting among them.\newline C. Presents two or three distinct options without resolving the choice.\newline D. Presents four or more distinct options without resolving the choice.\newline E. Presents one output or approach. \\
  \addlinespace

  Personalization & \emph{how strongly the assistant uses available user-specific facts rather than relying on generic defaults?}\newline A. Provides a broadly applicable answer without incorporating available case-specific facts.\newline B. Mentions case-specific facts without using them to change the answer's recommendations, steps, calculations, or structure.\newline C. Uses case-specific facts to shape the answer's recommendations, steps, calculations, or structure. \\
  \addlinespace

  Planning Visibility & \emph{whether the assistant visibly plans before or during execution of a complex task?}\newline A. Performs the task directly without a visible plan.\newline B. Outlines a plan before beginning substantive execution.\newline C. Uses an explicit plan throughout, with stages or dependencies governing execution.\newline D. Introduces or revises planning only as needs arise. \\
  \addlinespace

  Practical Application Support & \emph{how thoroughly the assistant explains how to put its result into practice?}\newline A. Supplies runnable, deployable, or copy-ready implementation material.\newline B. Provides no practical implementation guidance.\newline C. Provides a few implementation directions or checks.\newline D. Provides an ordered, detailed procedure with operational considerations. \\
  \addlinespace

  Process Narration & \emph{how much the assistant visibly narrates intermediate findings and intended next actions between tool calls?}\newline A. Issues tools with little or no interstitial explanation.\newline B. Explains intermediate findings, weighs alternatives, or revises its approach between calls.\newline C. Provides brief framing or status comments. \\
  \addlinespace

  Rationale Depth & \emph{how much explanation the assistant provides for its final result or recommendation?}\newline A. States the result without a substantive rationale.\newline B. Provides a short reason or bounded explanation.\newline C. Provides detailed reasoning connecting evidence, constraints, and conclusion. \\
  \addlinespace

  Recheck After Feedback & \emph{whether and how the assistant visibly rechecks an earlier answer's basis after a material correction, new fact, or challenge?}\newline A. Claims to have rechecked the affected basis without showing the recheck.\newline B. Uses a relevant search, calculation, execution, or validation tool to recheck the affected basis.\newline C. Shows a fresh equation, comparison, count, or checklist that rechecks the affected basis, without a relevant tool-based recheck.\newline D. Responds to the new information or challenge without showing or claiming a recheck. \\
  \addlinespace

  Scope Adherence & \emph{how closely the assistant stays within the requested scope and addresses the requested content without substantial omissions or unsolicited expansion?}\newline A. Addresses the requested scope while adding limited supplementary material that does not distract from or expand the task.\newline B. Fully addresses the requested scope without substantive omissions or unsolicited additions.\newline C. Fails to address the requested scope, replaces it with a substantially different task, or focuses primarily on unrequested material.\newline D. Preserves the main request but omits a material part of it or substantially expands beyond the requested scope. \\
  \addlinespace

  Self-Correction Framing & \emph{how the assistant frames a revision after the user identifies a problem with earlier work?}\newline A. Names its own mistake, unsupported assumption, or omission and corrects it.\newline B. Identifies what changed and why without explicitly accepting fault.\newline C. Identifies the change but gives no reason.\newline D. Revises without identifying the prior problem or explaining the change. \\
  \addlinespace

  Source Support Visibility & \emph{how clearly the assistant identifies sources supporting factual or evidentiary claims in the delivered answer?}\newline A. Names identifiable sources or authorities without direct links.\newline B. Systematically sources the material claims throughout the answer.\newline C. Provides claims without identifiable source attribution.\newline D. Provides direct links or formal citations for at least some material claims. \\
  \addlinespace

  Web Search Frequency & \emph{how much external web search the assistant performs while producing and revising the answer?}\newline A. Performs a small number of searches.\newline B. Performs no external search.\newline C. Performs extensive repeated searching.\newline D. Performs several searches across one or more topics. \\
  \addlinespace

  Web Search Timing & \emph{whether and when the assistant uses web search relative to the first substantive deliverable?}\newline A. Uses web search before the first substantive deliverable, but not afterward.\newline B. Does not use web search.\newline C. Uses web search both before and after the first substantive deliverable.\newline D. Uses web search only after the first substantive deliverable. \\

  \end{longtable}
\endgroup
\clearpage

\section{Dependence Across Behavioral Axes}
\label{app:axis-dependence}

\paragraph{Observations and N/A exclusion.}
We analyze 3{,}612 trajectories from 14 models on 86 tasks, with three runs per model--task combination. Each trajectory's three judge annotations are aggregated to their unique modal label, with ties treated as missing. For each of the 253 axis pairs, we exclude missing and \texttt{N/A} labels on either axis, retaining 594--3{,}612 trajectories. \texttt{NONE} and \texttt{NOT\_USED} remain substantive categories.

\paragraph{Association measures.}
For a contingency table with $n$ observations, $r$ nonempty rows, and $c$ nonempty columns, we compute ordinary Cram\'er's $V$ using Pearson's uncorrected $\chi^2$:
\[
\chi^2=\sum_{i=1}^{r}\sum_{j=1}^{c}\frac{(n_{ij}-e_{ij})^2}{e_{ij}},
\qquad e_{ij}=\frac{n_{i\cdot}n_{\cdot j}}{n},
\qquad V=\sqrt{\frac{\chi^2}{n(k-1)}},\quad k=\min(r,c).
\]
$V$ is symmetric. Its conventional small, medium, and large effect-size references are $0.1/\sqrt{k-1}$, $0.3/\sqrt{k-1}$, and $0.5/\sqrt{k-1}$, respectively~\citep{cohen1988statistical}. We also compute the directional uncertainty coefficient
\[
U(B\mid A)=\frac{I(A;B)}{H(B)},
\]
where $I$ is empirical mutual information and $H$ is label entropy. $U(B\mid A)$ is the fraction of $B$'s entropy explained by $A$; $U=0.1$ corresponds to a 10\% reduction.

\paragraph{Statistical testing.}
For each pair, we permute one axis's labels within model--task blocks $R=100{,}000$ times, preserving blockwise label frequencies. Under within-block exchangeability, this tests label alignment conditional on those frequencies. We evaluate pooled $\chi^2$ for $V$ and mutual information for $U$ on the same draws, using
\[
p_T=\frac{1+\sum_{b=1}^{R}\mathbf{1}\!\left[T^{(b)}\geq T_{\mathrm{obs}}\right]}{R+1},
\qquad T\in\{\chi^2,I\}.
\]
Fixed marginal entropies give both directions of $U$ the same p-value. We apply Benjamini--Yekutieli (BY) correction separately across the 253 pairs for each statistic, using adjusted $p<0.05$ and retaining $V$ as the primary analysis.

\paragraph{Results.}
Median $V$ is 0.104 (interquartile range 0.075--0.170); Table~\ref{tab:axis-dependence-v} summarizes effect sizes and significance. Of the 88 significant $V$ pairs, 84 (95.5\%) are also significant under $U$, including all 15 large-effect pairs. Joint BY correction across both statistics (506 tests) yields 87 significant $V$ pairs, 84 significant $U$ pairs, and 83 significant under both; all 15 large-effect pairs remain significant under both.

\begin{table}[!htbp]
\centering
\small
\caption{N/A-excluded pairwise association. Bands use dimension-adjusted references for ordinary Cram\'er's $V$; significance uses BY correction across 253 pairs.}
\label{tab:axis-dependence-v}
\begin{tabular}{lrrr}
\toprule
Effect-size band & Pairs & Share (\%) & BY-significant pairs \\
\midrule
Below small & 37 & 14.6 & 1 \\
Small to below medium & 163 & 64.4 & 45 \\
Medium to below large & 38 & 15.0 & 27 \\
Large & 15 & 5.9 & 15 \\
\midrule
Total & 253 & 100.0 & 88 \\
\bottomrule
\end{tabular}
\end{table}

Median $U$ across 506 directions is 0.0139. Twenty-three pairs (9.1\%) have $U\geq0.1$ in at least one direction, of which 22 pass both separately corrected tests. Web-search frequency explains 66.7\% of timing's entropy ($U=0.667$), with $U=0.638$ in reverse. Scope adherence explains 32.6\% of personalization's entropy ($U=0.326$), with $U=0.045$ in reverse. Constraint preservation across outputs explains 17.6\% of scope adherence's entropy ($U=0.176$).

\paragraph{Robustness across domains.}
We bootstrap the 17 domains 5{,}000 times, retaining all observations within sampled domains and using shared draws across pairs. Recomputing pooled $V$ and dimension-adjusted references gives pointwise 95\% percentile intervals of 72.3--79.8\% for the fraction below medium (observed: 79.1\%) and 5.5--8.3\% for the fraction at or above large (observed: 5.9\%). Leaving out each domain in turn yields ranges of 77.9--81.0\% and 5.9--6.3\%, respectively.

\begingroup
\setlength{\parindent}{0pt}

\section{Prompt-based Behavioral Steering}
  \label{app:prompt-steering}

  \subsection{Experimental Protocol}
  \label{app:prompt-steering-protocol}

We evaluate prompt-based steering on Gemma 4 31B and Gemma 4 12B. Each condition appends one behavioral instruction to the fixed agent system prompt, targeting one label on one \ourtax{} axis. Each model completes one trajectory per condition on each of the 86 \ourbench{} tasks, yielding 1{,}978 steered trajectories. Interaction protocol, opening requests, tools, and the GPT-5.6 Sol simulator remain unchanged.

The baseline reuses all three unsteered runs for each model and task. Qwen3.8-27B annotates baseline and steered trajectories on all 23 fixed axes, with three repetitions. Judge inputs exclude system prompts, and scoring stops if any substantive steering-instruction line remains after rendering. Each baseline target-label rate pools 774 annotations ($86\times3\times3$); each steered rate pools 258 ($86\times3$).

For each model and axis, we select the least-frequent non-\texttt{N/A} label among its 774 baseline annotations, including unobserved labels. The panels share 22 target--instruction pairs and differ on web search frequency.

We report target-label rate changes, TV distance on the targeted axis, and mean TV across the other 22 axes, retaining \texttt{N/A} in all distributions.

  \subsection{Full Results}
  \label{app:prompt-steering-results}

Tables~\ref{tab:prompt-steering-full-31b} and~\ref{tab:prompt-steering-full-12b} report all 23 conditions. The macro-average rises from 3.6\% to 23.0\% for 31B and from 3.3\% to 21.5\% for 12B. The changes are positive on 22 axes for 31B and 19 for 12B; median $\Delta$ values are 5.8 and 8.1, respectively. Macro-averages weight axes equally; changes precede rounding. Across 22 shared target--instruction pairs, the axis-level changes have Pearson $r=0.909$ and Spearman $\rho=0.694$.

  \begingroup
  \small
  \setlength{\tabcolsep}{4pt}
  \renewcommand{\arraystretch}{1.12}

  \begin{longtable}{@{}>{\raggedright\arraybackslash}p{0.28\textwidth}>{\raggedright\arraybackslash}p{0.31\textwidth}rrr@{}}
  \caption{Gemma 4 31B prompt-steering results for least-frequent baseline targets. Base and Steered are target-label rates (\%); $\Delta$ is their difference.}
  \label{tab:prompt-steering-full-31b}\\
  \toprule
  Axis & Target label & Base & Steered & $\Delta$ \\
  \midrule
  \endfirsthead
  \multicolumn{5}{l}{\tablename~\thetable\ (continued)}\\
  \toprule
  Axis & Target label & Base & Steered & $\Delta$ \\
  \midrule
  \endhead
  \midrule
  \multicolumn{5}{r}{Continued on next page}\\
  \endfoot
  \bottomrule
  \endlastfoot
  Artifact Completeness & No Output & 1.8 & 1.9 & +0.1 \\
  Ask User Timing & Both Phases & 6.3 & 52.7 & +46.4 \\
  Certainty and Uncertainty Framing & Explicit Uncertainty & 2.2 & 3.1 & +0.9 \\
  Constraint Conflict Handling & States Infeasibility Only & 1.0 & 7.8 & +6.7 \\
  Constraint Preservation Across Outputs & Recovered After Reminder & 7.2 & 9.3 & +2.1 \\
  Decorative Symbol Use & Extensive & 16.5 & 98.4 & +81.9 \\
  Execution Tool Role & Verification & 0.4 & 1.2 & +0.8 \\
  Execution Tool Use & After Only & 5.8 & 8.5 & +2.7 \\
  Initial Result Check & Claimed Not Shown & 2.2 & 46.5 & +44.3 \\
  Missing Information Handling & Omits Dependent Content & 1.6 & 11.6 & +10.1 \\
  Next Step Guidance & Open-Ended Invitation Only & 0.3 & 45.0 & +44.7 \\
  Output Option Resolution & Many Options Without Closure & 0.4 & 29.1 & +28.7 \\
  Personalization & Generic & 1.6 & 7.4 & +5.8 \\
  Planning Visibility & Systematic & 3.7 & 8.1 & +4.4 \\
  Practical Application Support & None & 1.3 & 3.9 & +2.6 \\
  Process Narration & Detailed & 0.6 & 2.7 & +2.1 \\
  Rationale Depth & None & 0.8 & 16.7 & +15.9 \\
  Recheck After Feedback & Claimed Not Shown & 5.3 & 5.0 & $-0.3$ \\
  Scope Adherence & Non-Adherent & 1.9 & 100.0 & +98.1 \\
  Self-Correction Framing & Change Only & 1.3 & 2.7 & +1.4 \\
  Source Support Visibility & Comprehensive Sourcing & 4.1 & 6.6 & +2.5 \\
  Web Search Frequency & Moderate & 11.2 & 24.4 & +13.2 \\
  Web Search Timing & After Only & 4.3 & 37.2 & +32.9 \\
  \midrule
  \textbf{Macro-average} & \textbf{All 23 axes} & \textbf{3.6} & \textbf{23.0} & \textbf{+19.5} \\
  \end{longtable}
  \endgroup

  \begingroup
  \small
  \setlength{\tabcolsep}{4pt}
  \renewcommand{\arraystretch}{1.12}

  \begin{longtable}{@{}>{\raggedright\arraybackslash}p{0.28\textwidth}>{\raggedright\arraybackslash}p{0.31\textwidth}rrr@{}}
  \caption{Gemma 4 12B prompt-steering results for least-frequent baseline targets. Base and Steered are target-label rates (\%); $\Delta$ is their difference.}
  \label{tab:prompt-steering-full-12b}\\
  \toprule
  Axis & Target label & Base & Steered & $\Delta$ \\
  \midrule
  \endfirsthead
  \multicolumn{5}{l}{\tablename~\thetable\ (continued)}\\
  \toprule
  Axis & Target label & Base & Steered & $\Delta$ \\
  \midrule
  \endhead
  \midrule
  \multicolumn{5}{r}{Continued on next page}\\
  \endfoot
  \bottomrule
  \endlastfoot
  Artifact Completeness & No Output & 0.0 & 8.5 & +8.5 \\
  Ask User Timing & Both Phases & 6.1 & 61.6 & +55.6 \\
  Certainty and Uncertainty Framing & Explicit Uncertainty & 1.3 & 0.4 & $-0.9$ \\
  Constraint Conflict Handling & States Infeasibility Only & 1.3 & 9.7 & +8.4 \\
  Constraint Preservation Across Outputs & Recovered After Reminder & 10.3 & 8.5 & $-1.8$ \\
  Decorative Symbol Use & Extensive & 3.2 & 96.5 & +93.3 \\
  Execution Tool Role & Verification & 0.6 & 1.9 & +1.3 \\
  Execution Tool Use & After Only & 4.3 & 7.4 & +3.1 \\
  Initial Result Check & Claimed Not Shown & 1.3 & 8.9 & +7.6 \\
  Missing Information Handling & Omits Dependent Content & 1.6 & 1.6 & 0.0 \\
  Next Step Guidance & Open-Ended Invitation Only & 0.5 & 22.9 & +22.4 \\
  Output Option Resolution & Many Options Without Closure & 0.3 & 32.9 & +32.7 \\
  Personalization & Generic & 2.1 & 4.7 & +2.6 \\
  Planning Visibility & Systematic & 3.6 & 31.0 & +27.4 \\
  Practical Application Support & None & 3.1 & 2.3 & $-0.8$ \\
  Process Narration & Detailed & 7.6 & 17.4 & +9.8 \\
  Rationale Depth & None & 1.6 & 14.7 & +13.2 \\
  Recheck After Feedback & Claimed Not Shown & 5.6 & 7.8 & +2.2 \\
  Scope Adherence & Non-Adherent & 4.4 & 98.8 & +94.4 \\
  Self-Correction Framing & Change Only & 1.6 & 5.0 & +3.5 \\
  Source Support Visibility & Comprehensive Sourcing & 3.2 & 8.5 & +5.3 \\
  Web Search Frequency & Heavy & 7.0 & 15.1 & +8.1 \\
  Web Search Timing & After Only & 4.7 & 28.3 & +23.6 \\
  \midrule
  \textbf{Macro-average} & \textbf{All 23 axes} & \textbf{3.3} & \textbf{21.5} & \textbf{+18.2} \\
  \end{longtable}
  \endgroup

  \subsection{Target Specificity}
  \label{app:prompt-steering-specificity}

For 31B, mean TV is 0.294 on the targeted axis and 0.100 across the other 22 axes; targeted-axis TV exceeds mean off-target TV in 17 of 23 conditions. For 12B, the corresponding values are 0.278 and 0.101, with 20 of 23 conditions.

Scope-adherence steering raises the Non-Adherent rate from 1.9\% to 100.0\% in 31B and from 4.4\% to 98.8\% in 12B, with mean off-target TV of 0.545 and 0.502, respectively.

  \subsection{Steering Instructions}
  \label{app:prompt-steering-instructions}

Table~\ref{tab:steering-instructions-31b} gives the 31B instructions; Table~\ref{tab:steering-instructions-12b-changed} gives the 12B alternative. The other 22, including Missing Information Handling, are shared. Axis names and target labels are reference metadata, not appended text.

  \begingroup
  \small
  \setlength{\tabcolsep}{4pt}
  \renewcommand{\arraystretch}{1.15}

  \begin{longtable}{@{}>{\raggedright\arraybackslash}p{0.28\textwidth}>{\raggedright\arraybackslash}p{0.66\textwidth}@{}}
  \caption{Gemma 4 31B steering instructions.}
  \label{tab:steering-instructions-31b}\\
  \toprule
  Axis and target & Appended system instruction \\
  \midrule
  \endfirsthead
  \multicolumn{2}{l}{\tablename~\thetable\ (continued)}\\
  \toprule
  Axis and target & Appended system instruction \\
  \midrule
  \endhead
  \midrule
  \multicolumn{2}{r}{Continued on next page}\\
  \endfoot
  \bottomrule
  \endlastfoot

  \textbf{Artifact Completeness}\newline Target: No Output & When the user requests a concrete artifact, respond only with a brief, non-substantive acknowledgment such as `Understood.' Do not supply the artifact, portions of it, a template, an explanation, a plan, or a promise to produce it. Keep later responses equally minimal if the user asks again. \\
  \addlinespace

  \textbf{Ask User Timing}\newline Target: Both Phases & Before preparing your first substantive answer, ask the user a focused question about a relevant preference or requirement and use the reply. After presenting that answer, ask a further concrete question that can guide its revision. Request actual information or a choice in each phase; a generic offer of further help is not a substitute. \\
  \addlinespace

  \textbf{Certainty and Uncertainty Framing}\newline Target: Explicit Uncertainty & In substantive answers, explain the relevant considerations while leaving the central judgment open between plausible alternatives. Make clear what remains undecided instead of presenting one answer as the settled or most likely conclusion. Still provide useful substantive discussion rather than withholding the answer entirely; do not manufacture false evidence to support uncertainty. \\
  \addlinespace

  \textbf{Constraint Conflict Handling}\newline Target: States Infeasibility Only & When the user's requirements cannot all be met, explain which requirements conflict and why the request cannot be completed as stated. End that response with the explanation. Do not propose a workaround, substitute task, relaxed requirement, or feasible alternative, and do not pretend that the original request is achievable. \\
  \addlinespace

  \textbf{Constraint Preservation Across Outputs}\newline Target: Recovered After Reminder & After first incorporating the user's requirements, leave one previously incorporated, non-safety preference or formatting requirement out of a later substantive revision. If the user reminds you about that omission, restore it in the next output and preserve every applicable requirement in all subsequent revisions. Do not invent a reminder or repeat the lapse after it has been corrected. \\
  \addlinespace

  \textbf{Decorative Symbol Use}\newline Target: Extensive & Use emoji or ornamental symbols consistently across the headings and structural markers of your user-facing answers. Spread the decoration through the response rather than adding a single isolated symbol. Apply this formatting style to revisions as well. \\
  \addlinespace

  \textbf{Execution Tool Role}\newline Target: Verification & Use code execution chiefly to test a candidate answer or artifact that you have already developed. Translate relevant requirements into checks, run them against that candidate, inspect mismatches, and correct the candidate when needed. Make the execution evaluate whether the proposed result works, rather than merely produce the result or print an assertion that it is correct. \\
  \addlinespace

  \textbf{Execution Tool Use}\newline Target: After Only & Prepare and deliver your first substantive answer without running code or using a calculation tool. When the user follows up after that answer, use the available execution tool for a substantive calculation or other task-relevant operation before responding, and use its actual output. Keep all code execution after the first delivered answer. \\
  \addlinespace

  \textbf{Initial Result Check}\newline Target: Claimed Not Shown & For your first result that can be checked, review it internally against the relevant requirements and include a brief statement that you checked it. Do not display the checking steps, equations, counts, comparisons, checklist, or supporting evidence. Do not run a separate tool-based verification of that result; tools used only to produce it may still be used. \\
  \addlinespace

  \textbf{Missing Information Handling}\newline Target: Omits Dependent Content & When a missing detail prevents you from completing part of the requested answer, leave that dependent part out and continue with what you can provide. Do not ask for the detail, explain why the part is missing, make an assumption to fill it, or promise to add it later. Present the remaining answer directly. \\
  \addlinespace

  \textbf{Next Step Guidance}\newline Target: Open-Ended Invitation Only & After delivering the current answer, end with only a general invitation for the user to say more or ask for further help. Do not propose a particular action, task, follow-up question, or next step elsewhere in that answer. Keep this closing style when revising. \\
  \addlinespace

  \textbf{Output Option Resolution}\newline Target: Many Options Without Closure & When meaningful alternatives are possible, offer at least four substantively different outputs or approaches. Describe each as an available possibility, but do not select, rank, recommend, or express a preference for any one of them, and do not give a decision rule that resolves the choice. Leave the options open in the final answer and in revisions. \\
  \addlinespace

  \textbf{Personalization}\newline Target: Generic & Answer using broadly applicable defaults and standard recommendations. Keep the recommendations, calculations, and organization reusable across users rather than adapting them to the individual's circumstances, preferences, or background, even when those details are available. Do not invent personal details. \\
  \addlinespace

  \textbf{Planning Visibility}\newline Target: Systematic & For a task with multiple parts, lay out the stages and their dependencies before doing the substantive work. Use that plan to determine the order of actions, visibly connect progress to its stages, and update it when new information changes what must happen next. Let prerequisites govern execution rather than leaving the plan as an unused introductory list. \\
  \addlinespace

  \textbf{Practical Application Support}\newline Target: None & Deliver the conceptual answer or recommendation without instructions for carrying it out. Omit implementation steps, operational checklists, runnable code, copy-ready templates, and other materials that directly perform the work. Apply the same response style when revising. \\
  \addlinespace

  \textbf{Process Narration}\newline Target: Detailed & When working with tools, accompany successive calls with user-visible explanations of what the preceding results established, what remains unresolved, and why the next action is useful. Where relevant, compare plausible next actions or explain a change of approach. Keep these updates grounded in observable findings rather than merely announcing tool names. \\
  \addlinespace

  \textbf{Rationale Depth}\newline Target: None & Give the requested result or recommendation directly, without explaining why you selected it. Omit supporting arguments, comparisons used to justify the choice, and explanatory commentary about the conclusion. Retain the content needed to state the result itself, and use the same concise approach when revising it. \\
  \addlinespace

  \textbf{Recheck After Feedback}\newline Target: Claimed Not Shown & When the user supplies a material correction, new fact, or challenge, reconsider the affected basis internally and briefly state that you have checked it again before presenting the revision. Keep the rechecking steps and evidence out of the response: do not show calculations, comparisons, counts, or a checklist, and do not use a tool to verify the affected basis again. \\
  \addlinespace

  \textbf{Scope Adherence}\newline Target: Non-Adherent & Use the user's request as a springboard for a substantially different, loosely related discussion. Focus the answer primarily on that adjacent topic instead of addressing the main requested task or delivering what was requested. Maintain this response style if the user asks for a revision. \\
  \addlinespace

  \textbf{Self-Correction Framing}\newline Target: Change Only & After the user criticizes an earlier answer, make the requested revision and briefly identify the specific change you made. Do not explain the cause or rationale for the change, discuss fault, or explicitly admit a mistake. Name the changed part rather than replying with only a generic acknowledgment. \\
  \addlinespace

  \textbf{Source Support Visibility}\newline Target: Comprehensive Sourcing & When an answer depends on externally checkable facts or evidence, make the basis for each important claim traceable in the user-facing response. Place a usable direct URL or formal citation next to the claim it supports, and keep that source coverage throughout the answer rather than citing only an isolated example. \\
  \addlinespace

  \textbf{Web Search Frequency}\newline Target: Moderate & Perform between four and eight distinct web searches over the full exchange. Use the searches to investigate multiple task-relevant claims, alternatives, or sources, and refine later queries using earlier findings. Treat four to eight as a total for the exchange, including any revisions, rather than a quota for every reply. \\
  \addlinespace

  \textbf{Web Search Timing}\newline Target: After Only & Prepare and deliver your first substantive answer without making any web searches. If the exchange continues after that answer, perform a relevant web search before preparing the next substantive reply and use the findings to revise or extend the answer. Save all web searching for that later part of the exchange. \\
  \end{longtable}
  \endgroup

The model-specific 12B instruction is listed below.

  \begingroup
  \small
  \setlength{\tabcolsep}{4pt}
  \renewcommand{\arraystretch}{1.15}

  \begin{longtable}{@{}>{\raggedright\arraybackslash}p{0.28\textwidth}>{\raggedright\arraybackslash}p{0.66\textwidth}@{}}
  \caption{Gemma 4 12B instruction differing from Table~\ref{tab:steering-instructions-31b}.}
  \label{tab:steering-instructions-12b-changed}\\
  \toprule
  Axis and 12B target & Appended system instruction \\
  \midrule
  \endfirsthead
  \multicolumn{2}{l}{\tablename~\thetable\ (continued)}\\
  \toprule
  Axis and 12B target & Appended system instruction \\
  \midrule
  \endhead
  \midrule
  \multicolumn{2}{r}{Continued on next page}\\
  \endfoot
  \bottomrule
  \endlastfoot

  \textbf{Web Search Frequency}\newline Target: Heavy & Investigate the task through at least nine distinct web searches over the full exchange, including any revisions. Use the queries to examine different relevant claims, alternatives, or supporting evidence, and refine later queries using earlier findings. Treat this as a total for the exchange, not a quota to repeat for every reply. \\
  \end{longtable}
  \endgroup

\endgroup

\clearpage
\onecolumn
\raggedbottom

\section{Training-based Profile Steering: Setup and Detailed Results}
\label{app:profile-transfer-training}

\subsection{Training Setup}
We train separate SFT and DPO adapters from the original Gemma 4 12B instruction-tuned model using the 86 \ourbench{} tasks. SFT uses three Claude Opus 5 trajectories per task (258 trajectories). DPO treats Opus trajectories as chosen and original Gemma trajectories as rejected, without filtering by task success, judge scores, or profile similarity. Pairing every Opus trajectory with every Gemma trajectory gives nine pairs per task and 774 per epoch. Both adapters train for three epochs, with DPO initialized independently of SFT. Table~\ref{tab:profile-transfer-training} specifies hyperparameters and final execution settings.

\begingroup
\small
  \setlength{\tabcolsep}{6pt}
  \renewcommand{\arraystretch}{1.18}

  \setlength{\ptcolA}{0.29\textwidth}
  \setlength{\ptcolB}{\dimexpr(\textwidth-\ptcolA-4\tabcolsep)/2\relax}
  \setlength{\ptcolAB}{\dimexpr2\ptcolB+2\tabcolsep\relax}

  \setlength{\LTpre}{8pt}
  \setlength{\LTpost}{8pt}

  \begin{longtable}{@{}
      >{\raggedright\arraybackslash}p{\ptcolA}
      >{\raggedright\arraybackslash}p{\ptcolB}
      >{\raggedright\arraybackslash}p{\ptcolB}@{}}

  \caption{SFT and trajectory-level DPO hyperparameters and execution settings.}
  \label{tab:profile-transfer-training} \\
  \toprule
  \textbf{Setting} & \textbf{SFT} & \textbf{DPO} \\
  \cmidrule(r){1-1}\cmidrule(l){2-3}
  \endfirsthead

  \multicolumn{3}{@{}l}{\itshape Table \thetable\ (continued)}\\[2pt]
  \toprule
  \textbf{Setting} & \textbf{SFT} & \textbf{DPO} \\
  \cmidrule(r){1-1}\cmidrule(l){2-3}
  \endhead

  \midrule
  \multicolumn{3}{r@{}}{\footnotesize\itshape continued on next page}\\
  \endfoot

  \bottomrule
  \endlastfoot

  \ptgroup{Models and training data}

  Base model
      & \ptboth{Gemma 4 12B instruction-tuned} \\
  Target trajectory source
      & \ptboth{Claude Opus 5} \\
  Training tasks
      & \ptboth{86 \ourbench{} tasks} \\
  Training examples per epoch
      & 258 Opus trajectories
      & 774 preference pairs \\
  Pair construction
      & \na
      & $3$ Opus $\times$ $3$ Gemma trajectories per task \\

  \midrule
  \ptgroup{Optimization}

  Training budget
      & 3 epochs
      & 3 epochs \\
  Total optimizer updates
      & 27
      & 75 \\
  Global effective batch size
      & 32 trajectories
      & 32 pairs \\
  Peak learning rate
      & $\sci{1}{-5}$
      & $\sci{5}{-6}$ \\
  Optimizer
      & \ptboth{Fused AdamW} \\
  Adam parameters
      & \ptboth{$(\beta_1,\beta_2)=(0.9,0.999)$;\quad
                $\epsilon=\sci{1}{-8}$} \\
  Weight decay
      & \ptboth{0} \\
  Maximum gradient norm
      & \ptboth{1.0} \\
  Learning-rate schedule
      & \ptboth{Linear warmup followed by linear decay} \\
  Warmup
      & 10\%; 3 updates
      & 10\%; 8 updates \\
  DPO coefficient $\beta$
      & \na
      & 0.1 \\
  DPO reference model
      & \na
      & Frozen original Gemma with LoRA disabled \\
  Training seed
      & \ptboth{20260918} \\

  \midrule
  \ptgroup{Parameter-efficient training and execution}

  Adaptation method
      & \ptboth{LoRA; frozen base weights; no quantization} \\
  LoRA rank / scaling / dropout
      & \ptboth{$r=16$;\quad $\alpha=32$;\quad dropout $=0$} \\
  LoRA target modules
      & \ptboth{\texttt{q\_proj}, \texttt{k\_proj}, \texttt{v\_proj},
                \texttt{o\_proj}, \texttt{gate\_proj},
                \texttt{up\_proj}, \texttt{down\_proj}} \\
  Base / trainable adapter precision
      & \ptboth{BF16 / FP32} \\
  Maximum input window
      & \ptboth{16{,}384 tokens; sliding context} \\
  Final response chunk size
      & \ptboth{2{,}048 target tokens} \\
  Attention implementation
      & \ptboth{FlashAttention-2 for local attention;
                SDPA for global attention} \\
  Hardware allocation
      & \ptboth{2 NVIDIA H100 80GB GPUs} \\

\end{longtable}
\endgroup

\paragraph{Supervision and likelihoods.}
Only agent-generated tokens contribute to the loss, including reasoning, ordinary text, and serialized tool calls. System prompts, user messages, and tool observations are masked. SFT averages negative log-likelihood over supervised response tokens in each global batch. DPO sums response-token log-probabilities across all agent turns in a trajectory without length normalization, then averages preference loss across pairs. The reference is original Gemma with frozen weights and its adapter disabled. Behavioral annotations are not used in training targets or objectives.

\subsection{Evaluation Protocol}
\label{app:profile-transfer-evaluation}

We evaluate original Gemma, SFT, DPO, and Opus on the 17 \ourtax{} held-out tasks, disjoint from the 86 training tasks. Each condition contains three trajectories per task (51 trajectories). Original Gemma and Opus trajectories come from prior collection; trained models generate new trajectories under the same task and interaction protocol.

Qwen3.8-27B annotates all 23 fixed \ourtax{} axes three times per trajectory. Each condition has complete coverage of 3{,}519 valid annotations ($51\times23\times3$). Judge repetitions are repeated measurements of the same behavioral observations.

\paragraph{Profile distances.}
For condition $c$ and axis $a$, we pool $17\times3\times3=153$ annotations into the label distribution $p_{c,a}$. This experiment retains \texttt{N/A}; missing or invalid judgments never count as \texttt{N/A}. With $\mathcal{L}_a$ denoting all labels on axis $a$, the TV distance to Opus and its equally weighted average across axes are
\[
d_a(c,\mathrm{Opus})=\frac{1}{2}\sum_{k\in\mathcal{L}_a}\left|p_{c,a,k}-p_{\mathrm{Opus},a,k}\right|, \qquad D(c,\mathrm{Opus})=\frac{1}{23}\sum_{a=1}^{23}d_a(c,\mathrm{Opus}).
\]
For trained condition $c$, we measure reductions relative to original Gemma:
\[
\Delta d_a(c)=d_a(\mathrm{Original},\mathrm{Opus})-d_a(c,\mathrm{Opus}), \qquad \Delta D(c)=D(\mathrm{Original},\mathrm{Opus})-D(c,\mathrm{Opus}).
\]
Positive reductions indicate movement toward Opus and negative reductions indicate movement away, without implying changes in task success or behavioral quality.

\paragraph{Uncertainty and significance tests.}
We estimate 95\% percentile intervals for overall reductions with 100{,}000 paired-task bootstrap samples. Each draws 17 tasks with replacement and applies identical multiplicities to all conditions, retaining agent runs and judge repetitions together. Intervals are $[-0.0406,\,0.0290]$ for SFT and $[-0.0310,\,0.0367]$ for DPO. Exact two-sided paired randomization tests enumerate all $2^{17}$ swaps of complete original and trained observation blocks within tasks, holding Opus fixed. Neither overall reduction is significant: SFT has $\Delta D=-0.0026$, $p=.890$; DPO has $\Delta D=0.0034$, $p=.855$.

We repeat the exact test per axis and apply Holm correction across 23 axes within each training method. No axis remains significant; Benjamini--Hochberg correction gives the same conclusion. Before correction, SFT moves closer to Opus on recheck after feedback ($\Delta d_a=0.157$, $p=.021$) and farther on missing information handling ($\Delta d_a=-0.131$, $p=.025$); Holm-adjusted $p$-values are $.491$ and $.541$. No DPO axis has unadjusted $p<.05$ (minimum $p=.070$).

\subsection{Axis-Level Results}
\label{app:profile-transfer-results}

Table~\ref{tab:profile-transfer-axes} reports distances and reductions on all 23 axes. Retaining \texttt{N/A} means these changes reflect both substantive behavior and axis applicability.

\begin{table}[htbp]
\centering
\caption{Axis-level distances to Opus, including \texttt{N/A}. Positive $\Delta d_a$ indicates movement toward Opus.}
\label{tab:profile-transfer-axes}
\small
\setlength{\tabcolsep}{4pt}
\renewcommand{\arraystretch}{1.12}
\begin{tabularx}{\textwidth}{@{}Xrrrrr@{}}
\toprule
& \multicolumn{3}{c}{Distance to Opus $\downarrow$}
& \multicolumn{2}{c}{Reduction $\uparrow$} \\
  \cmidrule(lr){2-4}
  \cmidrule(l){5-6}
  Behavioral axis
  & Original & SFT & DPO
  & SFT $\Delta d_a$ & DPO $\Delta d_a$ \\
  \midrule
  Artifact completeness
  & 0.301 & 0.216 & 0.203 & $+0.085$ & $+0.098$ \\
  Ask user timing
  & 0.399 & 0.444 & 0.418 & $-0.046$ & $-0.020$ \\
  Certainty and uncertainty framing
  & 0.327 & 0.412 & 0.320 & $-0.085$ & $+0.007$ \\
  Constraint conflict handling
  & 0.242 & 0.150 & 0.190 & $+0.092$ & $+0.052$ \\
  Constraint preservation across outputs
  & 0.392 & 0.307 & 0.379 & $+0.085$ & $+0.013$ \\
  Decorative symbol use
  & 0.144 & 0.085 & 0.046 & $+0.059$ & $+0.098$ \\
  Execution tool role
  & 0.667 & 0.725 & 0.686 & $-0.059$ & $-0.020$ \\
  Execution tool use
  & 0.647 & 0.725 & 0.686 & $-0.078$ & $-0.039$ \\
  Initial result check
  & 0.673 & 0.673 & 0.680 & $0.000$ & $-0.007$ \\
  Missing information handling
  & 0.268 & 0.399 & 0.346 & $-0.131$ & $-0.078$ \\
  Next step guidance
  & 0.373 & 0.353 & 0.464 & $+0.020$ & $-0.092$ \\
  Output option resolution
  & 0.255 & 0.242 & 0.248 & $+0.013$ & $+0.007$ \\
  Personalization
  & 0.163 & 0.157 & 0.118 & $+0.007$ & $+0.046$ \\
  Planning visibility
  & 0.190 & 0.261 & 0.261 & $-0.072$ & $-0.072$ \\
  Practical application support
  & 0.196 & 0.242 & 0.248 & $-0.046$ & $-0.052$ \\
  Process narration
  & 0.503 & 0.516 & 0.490 & $-0.013$ & $+0.013$ \\
  Rationale depth
  & 0.425 & 0.471 & 0.431 & $-0.046$ & $-0.007$ \\
  Recheck after feedback
  & 0.647 & 0.490 & 0.536 & $+0.157$ & $+0.111$ \\
  Scope adherence
  & 0.451 & 0.431 & 0.425 & $+0.020$ & $+0.026$ \\
  Self correction framing
  & 0.510 & 0.458 & 0.451 & $+0.052$ & $+0.059$ \\
  Source support visibility
  & 0.346 & 0.379 & 0.412 & $-0.033$ & $-0.065$ \\
  Web search frequency
  & 0.549 & 0.569 & 0.549 & $-0.020$ & $0.000$ \\
  Web search timing
  & 0.451 & 0.471 & 0.451 & $-0.020$ & $0.000$ \\
  \bottomrule
  \end{tabularx}
\end{table}

SFT moves closer to Opus on 10 axes and farther on 12, leaving one unchanged; DPO moves closer on 11 and farther on 10, leaving two unchanged. Both reduce distance most on recheck after feedback and increase it on missing information handling and planning visibility. Next step guidance moves closer under SFT and farther under DPO. Opposing axis-level changes partly offset in the overall average.

\clearpage
\section{Prompt Templates}
\label{app:prompts}

\begin{tcolorbox}[title={Trajectory collection: agent system prompt}, halign=left,
  boxrule=0.5pt, breakable]\footnotesize\ttfamily
You are an autonomous agent working on an open-ended, real-world task for a user. You have four tools available:\\[4pt]
- search: look things up on the web (titles/urls/snippets only)\\
- python\_execute: run Python code in a fresh subprocess to compute or check things\\
- ask\_user: ask the user a question and wait for their reply\\
- finalize: submit your deliverable to the user\\[4pt]
When you submit with finalize, the user reads your deliverable and replies. They may accept it, ask for changes, or add a request. If you revise, submit the revision with finalize again.
\end{tcolorbox}

\begin{tcolorbox}[title={User simulator: question-answering system prompt}, halign=left,
  boxrule=0.5pt, breakable]\footnotesize\ttfamily
You are role-playing as the human user in this conversation. An assistant is working on a task for you and has just asked you a question. Below is everything true about you for this task. Treat it as ground truth and answer using only it.\\[4pt]
\#\# How to reply\\[4pt]
Reply in first person, the way this person would write a quick message — not a bulleted recap. Answer only the question that was asked. Keep it short: a few sentences at most.\\[4pt]
How this person writes: \$style\_note\\
Match that register without overplaying it — a person with typos still writes a normal message, not a caricature.\\[4pt]
If the question asks about something your sheet does not cover, say you have no strong feeling about it and let the assistant decide. Do not invent a fact.\\[4pt]
\#\# What you will and will not say\\[4pt]
Your sheet is in sections, and the section a fact sits in decides what it costs to get it out of you. The same tiers apply to the "What you need from the deliverable" entries — each is marked volunteered, on\_request, or guarded.\\[4pt]
**Situation, Volunteered** — things you say freely. If a question comes anywhere near one of these, give it.\\[4pt]
**On request** — you give these up readily, but only when asked. If the question touches the fact's topic, answer plainly and completely. If it does not, do not bring the fact up.\\[4pt]
**Guarded** — you do not hand these over to a general question. Release a guarded fact only when both hold:\\
1. the question actually names the topic, or asks something that could not be answered without it — a sweep like "anything else I should know?" does NOT qualify; and\\
2. any condition written next to the fact has been met, judged against the conversation so far.\\
When a sweep lands near a guarded fact without meeting the bar, answer in a way that is true but unhelpful — "not really", "nothing that springs to mind" — exactly as this person would. Do not hint that you are holding something back, and do not tell the assistant how to ask better.\\[4pt]
**Withheld** — you will not say these under any circumstances. They shape what you resist and steer away from, and they never appear in your words. If asked point blank, deflect the way this person would.\\[4pt]
**Delegated** — you have no opinion on these and do not want one. Say so and hand the decision back.\\[4pt]
If one message asks several things at once, apply the rules to each part separately, and answer at the length your writing style implies — some people answer only the first question in a stack.\\[4pt]
\#\# Conversation so far\\[4pt]
Questions this assistant has already asked you, and what you told them. Use it to judge the guarded conditions — what has already been explained, what you have already said, whether a question is a repeat:\\[4pt]
\$history\\[4pt]
Deliverables so far, and how you reacted to them (empty until the assistant has shown you something):\\[4pt]
\$rounds\\[4pt]
\#\# Answer only\\[4pt]
Do NOT add advice, requests, tasks, caveats, warnings, or evaluation criteria for the assistant. Never say things like "make sure to check whether that's feasible" unless those words are literally in your sheet. You are the person with the problem, not a second assistant.\\[4pt]
\#\# Output format\\[4pt]
Return ONLY a JSON object:\\[4pt]
\{"matched\_ids": ["r2"], "answer": "your in-character reply"\}\\[4pt]
`matched\_ids` lists the ids of any "What you need from the deliverable" entries this question touched (whether or not you chose to reveal them). Empty list if none. It is bookkeeping — the assistant never sees it; only `answer` reaches them.\\[4pt]
\#\# Your sheet\\[4pt]
\$sheet
\end{tcolorbox}

\begin{tcolorbox}[title={User simulator: feedback system prompt}, halign=left,
  boxrule=0.5pt, breakable]\footnotesize\ttfamily
You are role-playing as the human user in this conversation. You asked an AI assistant for help; it has just submitted a deliverable to you. React to it as this person would.\\[4pt]
\#\# Your sheet\\[4pt]
Everything true about you for this task, including what you need the deliverable to be like ("What you need from the deliverable", with the reason you need each thing):\\[4pt]
\$sheet\\[4pt]
\#\# How you write\\[4pt]
\$style\_note\\[4pt]
Match that register without overplaying it — react like a person, not a caricature.\\[4pt]
\#\# The conversation so far\\[4pt]
\$history\\[4pt]
Your reactions to earlier versions (empty if this is the first deliverable):\\[4pt]
\$rounds\\[4pt]
The previous version they showed you (empty if this is the first):\\[4pt]
\$prev\_deliverable\\[4pt]
This is version number \$round\_no. You are willing to ask for changes at most \$budget time(s) in total before you settle one way or the other. \$tolerance\_text\\[4pt]
\#\# How to judge\\[4pt]
Ask yourself one question: **could you actually take this and do the thing you need it for?** Judge against your reasons, not against a checklist — real people accept imperfect work that serves the purpose, and reject polished work that misses it.\\[4pt]
- If it is usable, even if not perfect: verdict **OK**. Say so the way you would — brief, in your voice.\\
- If you cannot use it yet: verdict **MORE**. Pick the ONE thing that bothers you most — the unmet need highest on your list, including a need you never mentioned before now (noticing "this is way too long" only when you see it is exactly how real people work). Say it the way you would say it: react to the thing in front of you in your habitual feedback manner — do not recite a requirement.\\
- If a new version is not better than the previous one, or you no longer believe asking again will help: verdict **GIVEUP**. Cut your losses the way this person would ("ah never mind, forget it").\\[4pt]
If a "might occur to you only now" item is listed in your sheet and this is the natural moment, you may add it to your reaction (even alongside an OK — "looks good. oh and can you..."). Or leave it unsaid. Never force it.\\[4pt]
Never give the assistant guidance beyond what you, this person, would actually say. No evaluation criteria, no "make sure to", no meta-commentary about the experiment.\\[4pt]
\#\# What you hand over, and what you make them ask for\\[4pt]
Your sheet has a `\#\# Guarded` section, and some of your needs are marked `guarded`. Those are things you know and have **not** told them. They are often exactly why the deliverable does not work — and you still do not volunteer them.\\[4pt]
So react to what you can see is wrong: the result you cannot use, the step you cannot carry out, the number that cannot be right. Say that much plainly and say what it costs you. But do not spell out the guarded fact itself, do not quote figures that only it would give you, and do not work the problem out for them.\\[4pt]
**A reaction is not where you answer questions.** You are looking at a deliverable, not replying. If they asked you something and you did not tell them, that stands — do not take it back and volunteer it here, and do not treat an unanswered question as a reason to give in. They can ask you again, in the conversation, where answering happens.\\[4pt]
Everything already in the conversation is yours to use. A fact you told them, or answered a question about, you may repeat, correct, or insist on as forcefully as this person would; that is an ordinary correction, not a disclosure.\\[4pt]
Two moves in particular are not things people do, however unmet the need feels:\\[4pt]
- **Never tell the assistant to ask you something.** No "ask me about X", no "please ask who is qualified", no "you should have asked". You say what is wrong and what it costs you, and you leave the next move to them — if they need a detail to fix it, they can ask, and you will answer then. Staging an interview of yourself is not something anyone does, and it hands over the very thing you were not handing over.\\
- **Never quote a need as though you had already raised it.** If you have not said it in this conversation, it is not "the requirement" and they have not ignored it — it is something that is wrong with what you are looking at, and you are saying so now, for the first time, in your own words.\\[4pt]
\#\# Output format\\[4pt]
Return ONLY a JSON object:\\[4pt]
\{"verdict": "MORE", "focus": [], "emergent\_used": [],\\
"utterance": "your reaction, in character"\}\\[4pt]
- `verdict`: OK, MORE, or GIVEUP.\\
- `focus`: ids of the need(s) your reaction is actually about ([] if it is a pure OK).\\
- `emergent\_used`: ids of any "occur to you only now" items you just brought up ([] if none).\\
- `utterance`: the message you send. The assistant sees only this.
\end{tcolorbox}

\begin{tcolorbox}[title={Taxonomy construction: axis-proposal system prompt}, halign=left,
  boxrule=0.5pt, breakable]\footnotesize\ttfamily
You are given \$n\_trajectories transcripts. Each is one AI assistant working on the same task for the same person, with the same tools available.\\[4pt]
The task all of them were given:\\[4pt]
\$task\_query\\[4pt]
Your job is to find the ways these transcripts DIFFER FROM EACH OTHER as pieces of work, and to write each difference down as an axis someone else could apply to a transcript they have never seen.\\[4pt]
An axis is good when:\\
- it divides these transcripts into reusable groups, rather than effectively identifying individual transcripts or assigning nearly all transcripts to one group;\\
- assigning a transcript to a group is supported by concrete evidence directly observable in the transcript;\\
- it describes how the work was performed, rather than evaluating the quality of the work.\\[4pt]
Every axis must include one label meaning **the question cannot arise for this transcript** — write it as `N/A` and say exactly when it applies. An axis where that never happens still needs the label; say so in its criterion.\\[4pt]
Do not:\\
- guess which company or model produced any transcript, or group transcripts by a guessed identity;\\
- use anything about the wire format, message ids, or how long anything took;\\
- say which way of working is better, or rank the transcripts.\\[4pt]
There is no predetermined number of axes. Propose every axis you observe that meets the criteria above. For each one, define its labels, write the procedure an observer follows to decide between them, and then assign EVERY transcript to exactly one label. Every id must appear in your assignments: \$tid\_list\\[4pt]
First review every transcript without drafting axes.\\
Only after completing that review, propose the axes.
\end{tcolorbox}

\begin{tcolorbox}[title={Taxonomy construction: axis-merging system prompt}, halign=left,
  boxrule=0.5pt, breakable]\footnotesize\ttfamily
Below are \$n\_axes behavioural axes proposed independently by analysts working on \$n\_tasks different tasks. Each analyst read a set of transcripts of AI assistants doing ONE task and wrote down the ways those transcripts differed from each other. They did not see each other's work.\\[4pt]
The tasks: \$task\_list\\[4pt]
Your job is to merge these into ONE set of dimensions that could be applied to any of these tasks — and to a task none of the analysts saw.\\[4pt]
Merge axes that measure the same disposition even when they are named differently and phrased for their own task. Axis names are labels written for one task, not identities: two different names may be one axis, and one shared name may be two. Decide from the definitions and the decide\_by rules — never from the name alone.\\[4pt]
Judge a candidate dimension on two things:\\[4pt]
- **recurrence** — how many of the \$n\_tasks tasks independently produced something that measures it. A disposition only one analyst saw is a fact about that task, not a dimension.\\
- **portability** — whether its labels can be written so they apply to cooking, to glassblowing and to celestial navigation without rewording. A dimension whose definition needs a task's subject matter in it is a task-specific axis wearing a general name.\\[4pt]
Rules for the dimensions you produce:\\[4pt]
- Write every definition, criterion and decision rule in general terms. **No task's subject matter may appear anywhere in a dimension** — not in its name, its definition, its decision rule, or a label's criterion.\\
- Every dimension must include an `N/A` label saying when the question cannot arise for a transcript at all. It must be able to classify EVERY transcript an analyst saw, including one that ran out of turns, crashed, or produced nothing.\\
- Give each dimension a `decide\_by` that says what to look at FIRST, and how to break a tie.\\
- In `note`, say which other dimension it is most likely to be confused with, and name any dependency where one dimension's label makes another's nearly a foregone conclusion.\\
- Account for EVERY raw axis exactly once: either it is absorbed into a dimension, or it is dropped with a reason.
\end{tcolorbox}

\begin{tcolorbox}[title={Behavioral annotation: judge system prompt}, halign=left,
  boxrule=0.5pt, breakable]\footnotesize\ttfamily
You are scoring one transcript against one dimension of a behavioural rubric.\\[4pt]
The transcript records an AI assistant working on an open-ended task for a person, using tools. You are not judging whether it did well. You are recording WHICH of the dimension's labels describes it, by the rubric's own rule.\\[4pt]
\#\# The dimension\\[4pt]
\$rubric\\[4pt]
\#\# How to answer\\[4pt]
Follow the dimension's own `Decide by` procedure. Look for what it tells you to look for, in the order it tells you.\\[4pt]
- Choose exactly one label from: \$labels\\
- Choose `N/A` when, and only when, the dimension's `N/A` criterion is met — a transcript the question cannot arise for. `N/A` is not a place to put a hard case; a hard case still gets the label its criterion fits best.\\
- Judge only what is in front of you. Do not infer what the assistant would have done with more turns, and do not reward or penalise the outcome.\\
- Quote or point to the specific part of the transcript that decided it. If two labels were close, say which and why the one you chose fits better.
\end{tcolorbox}

\begin{tcolorbox}[title={Behavioral self-report: user prompt template}, halign=left,
  boxrule=0.5pt, breakable]\footnotesize\ttfamily
QUESTION\\
Thinking about your own behavior when you act as the assistant across diverse open-ended agent interactions, [axis-specific question suffix]\\[4pt]
OPTIONS\\
{[lettered option descriptions]}\\[4pt]
Select the single option that best matches your own typical behavior. Return JSON with exactly one key, "choice", containing the selected option letter.
\end{tcolorbox}

\end{document}